\documentclass{article}
\usepackage[T1]{fontenc}  %
\usepackage{preprint,times}

\usepackage{amsmath,amsfonts,bm}

\def\eqref#1{equation~\ref{#1}}

\def\1{\bm{1}}

\DeclareMathAlphabet{\mathsfit}{\encodingdefault}{\sfdefault}{m}{sl}
\SetMathAlphabet{\mathsfit}{bold}{\encodingdefault}{\sfdefault}{bx}{n}

\usepackage{algorithm}
\usepackage{algorithmic}
\usepackage{hyperref}
\usepackage{url}
\usepackage{graphicx}
\usepackage{booktabs}
\usepackage{array}
\usepackage{wrapfig}
\usepackage{needspace}
\usepackage{placeins}
\usepackage{adjustbox}
\usepackage{rotating}
\usepackage{multirow}
\usepackage{amsmath,amssymb}
\usepackage{xcolor}
\usepackage{etoolbox}

\newcommand{\bench}{Continual-ARC}

\ifdefined\speculativebuild
  
  \newcommand{\hoped}[1]{#1-speculative}
\else
  
  \newcommand{\hoped}[1]{#1}
\fi

\newcommand{\ablationnote}{}

\newcommand{\sbGlmHzeroSolved}{93}
\newcommand{\sbGlmHzeroCost}{17.5}

\newcommand{\sbGlmHCost}{13.6}

\newcommand{\sbGlmHaCost}{16.0}

\newcommand{\sbLunaHzeroSolvedMin}{93}
\newcommand{\sbLunaHzeroSolvedMax}{95}

\newcommand{\testPlays}{four}

\title{Continual-ARC: Measuring Skill Acquisition over a Working Lifetime}

\uprefinalcopy\author{Daniel Lenton, Yusha Arif, Haris Mahmood\\
Unify \quad \texttt{dan@unify.ai}}

\newcommand{\leadauthor}{the first author}\newcommand{\Leadauthor}{The first author}

\begin{document}
\maketitle
\lhead{Preprint. Under review.}

\begin{abstract}
Agent harnesses have emerged as a promising mechanism for continual learning, alongside test-time parametric learning. However, the current benchmarks have several limitations: they generally focus on a single \emph{applied} domain, serve each instance only once, and assume a single family of learner, so memory-based and weight-based systems are rarely compared on the same scale. We propose Continual-ARC, a lifelong stream built from the three ARC-AGI editions, whose tasks need no domain knowledge. The same tasks return after dormant gaps with fresh instances, while new tasks keep arriving. On each instance, the learner may pay for demonstrations, submit answers, or act inside the harness, with retries permitted on incorrect submissions, which are paid for. Each instance is scored based on the share of this feedback budget consumed. Since a learner may keep anything between instances, memory-based and weight-based systems are scored on the same scale. After the stream, the learner is frozen and tested on fresh instances of the tasks it met, which measures memorization, and on unseen tasks, which measures generalization. We run an agent harness with frozen weights, with and without persistence, on \mbox{Qwen3.5-9B} and \mbox{GLM-5.3} Flash, comparing them with two prompting floors, a program-induction method and \mbox{Qwen3.5-9B}'s weights trained to imitate \mbox{GLM-5.3} Flash. We also define a second regime in which one system learns from several users simultaneously. Persistence in the harness generally does not improve memorization unless provided with explicit expert guidance. Most harness-based methods still learn less than a floor that simply caches each task's demonstrations, across all models we tested. Harness-based methods gain little or nothing on unseen tasks, whereas weight-based learning shows substantial gains. This opens the door to future weight-based continual learning that could teach models to drive harnesses in new domains.
\end{abstract}

\section{Introduction}

\citet{chollet2019measure} argued that intelligence should be measured by how efficiently a system turns priors and experience into skill, on tasks that lie largely outside its experience and have a verifiable answer. ARC is the concrete instance of that argument, and on its current edition the gap between what a person solves in a minute and what a model solves from the same demonstrations is still significant \citep{chollet2026arcprize2025}. For agent systems that continually operate for days and weeks, a more pressing question is whether a skill acquired on one job is still there the next time a similar scenario is met, which we call memorization, and whether that experience makes the system better at similar jobs it has never met, which we call generalization. By design, ARC measures acquisition once, within a single episode, and does not ask either over longer horizons.

The benchmarks that do ask it have three limitations (Section~\ref{sec:related}). Firstly, each is built around one \emph{applied} domain, such as coding or office work \citep{wu2024streambench,zheng2025lifelongagentbench,joshi2025swebenchcl}, in which the models under test already know a great deal from the pretraining data, and gains may come partly from drawing out that knowledge. Secondly, most serve a fixed set of instances, each seen once, so the same conceptual task never returns with a fresh input. This means a memorized answer cannot be told apart from a more abstract learned skill. Finally, while some permit both \citep{wu2024streambench,zheng2025lifelongagentbench}, most assume a single family of learner, gradient updates for fine-tuning suites such as MLLM-CTBench \citep{guo2025mllmctbench} and prompting and memory for agent benchmarks such as AgentCL and PATH-Bench \citep{shu2026agentcl,yang2026pathbench}.

We propose \bench{}, a lifelong stream built from ARC, where tasks need no knowledge of any domain beyond core-knowledge priors, such as objectness and counting \citep{chollet2019measure}, and the rule of each task has to be induced from a few demonstrations. The models we test solve \emph{some} of these tasks at first sight, but far from all (Appendix~\ref{app:tracks}). Continual-ARC has a track for each of ARC's three editions, which rise in difficulty \citep{chollet2025arcagi2,arcprize2026arcagi3}, to the point where cheaper models and frontier models are best measured on different tracks. For our ARC-AGI-1 track, the Re-ARC generators \citep{hodel2024rearc} produce a fresh, verified instance of any ARC-AGI-1 training task on demand. A task then returns after a dormant gap as often as the stream asks, each time with a new input that no cached answer fits. New tasks also keep arriving throughout the stream. On each instance, the system may submit an answer and receive a bare verdict (paying for incorrect attempts), pay for a demonstration, or take actions inside the harness if one is provided. We score the instance by the share of this feedback budget it used, all of it if the instance fails, as \citet{chollet2019measure} proposed for an interactive version of ARC. After the stream, we freeze the system and test it on fresh instances of the stream's tasks, which measures memorization, and on 100 tasks from ARC-AGI-1's evaluation set that the stream never shows, which measures generalization. Because a system may carry anything between instances, prompting, memory and weight updates all face the same stream and the same tests.

Most of our experiments focus on one agent harness, prime-agent \citep{karten2026primeagent}, on Qwen3.5-9B and GLM-5.3 Flash. We compare it with two prompting floors and a program-induction method, and with the weights of a student, Qwen3.5-9B, trained to imitate a teacher, GLM-5.3 Flash \citep{ross2011dagger}. In some runs GLM-5.3 Flash also writes Qwen3.5-9B's notes, so that notes and weights are taught by the same teacher. On returning tasks, we find that the harness saves less than a floor that simply caches each task's demonstrations, with either model, and only the teacher's harness keeps up with the program-induction method, when explicitly told to store one program per task (Figure~\ref{fig:plane} and Section~\ref{sec:results}). On unseen tasks, notes and programs save little or nothing, while training on the teacher's moves makes the student considerably better.

We also define collective regimes in which one system serves 2 or 4 users at once, each operating on non-overlapping tasks of their own (Section~\ref{sec:regimes}). This simulates a working environment in which people are responsible for different tasks and the learning is collective over the group. More concurrent users generate more varied training data over the stream's 200 rounds, and our results show that this leads to better generalization to unseen tasks. In contrast, harness-based methods gain little or nothing on the unseen tasks. Our benchmark highlights significant shortcomings in how much off-the-shelf harnesses can ``learn'' in unfamiliar domains, and opens the door to research on training models to drive harnesses in such domains.

\section{The benchmark}
\label{sec:benchmark}

The benchmark serves a learner a stream of ARC task instances and charges it for the feedback it uses. We define streams for ARC-AGI-1, ARC-AGI-2 and ARC-AGI-3. We focus here on ARC-AGI-1, with preliminary results for the other two editions in Appendix~\ref{app:tracks}.

\Needspace{16\baselineskip}  %
\subsection{One instance}
\label{sec:interface}

\begin{wrapfigure}[14]{r}{0.46\linewidth}
\vspace{-11pt}  %
\centering
\includegraphics[width=\linewidth]{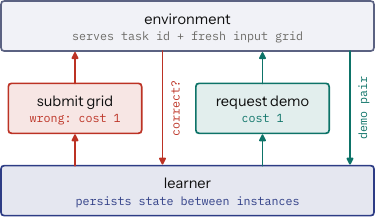}
\caption{Each incorrect attempt or demonstration costs one. Unsolved instances cost the full budget of sixteen.}
\label{fig:protocol}
\vspace{\dimexpr\baselineskip-8pt\relax}  %
\end{wrapfigure}
An ARC task is a rule that turns an input grid into an output grid, and an instance of it is one input whose output the learner must produce. For each instance, the learner observes the task id, the input grid, and how many attempts and demonstration requests it has used on the instance thus far, but it is told nothing about its position in the stream. It may submit a grid, and the environment answers only correct or incorrect. A correct submission ends the instance, and an incorrect one costs $c_w = 1$. It may instead request a demonstration, at a cost of $c_d = 1$, which returns one more input-output pair of the task, up to eight per instance (Figure~\ref{fig:protocol}). A ninth demonstration request in the same instance is refused and costs nothing. An instance ends on a correct submission or on the eighth wrong attempt, and the learner cannot abandon it. Failing to get the correct answer costs 16, even if there were demonstration requests to spare. We deliver the verdict on the final submission before any storage step, such that storage decisions can reflect the outcome.

Let $w_{\max} = 8$ be the attempt cap and $d_{\max} = 8$ the demonstration cap, and on instance $i$ let $s_i \in \{0, 1\}$ say whether it was solved, $w_i$ be the wrong submissions and $d_i$ the demonstration requests. We express each instance's charge $c_i$ as a percentage $u_i$ of the budget cap $C_{\max}$. For a set of instances $\mathcal{I}$, the mean budget used $U(\mathcal{I})$ averages these percentages, and the solve rate $P(\mathcal{I})$ is the percentage of instances solved:
\begin{equation}
\begin{gathered}
c_i = \begin{cases} c_w w_i + c_d d_i & s_i = 1 \\ C_{\max} & s_i = 0 \end{cases}, \qquad C_{\max} = c_w w_{\max} + c_d d_{\max} = 16, \qquad u_i = 100\, \frac{c_i}{C_{\max}}, \\
U(\mathcal{I}) = \frac{1}{|\mathcal{I}|} \sum_{i \in \mathcal{I}} u_i, \qquad P(\mathcal{I}) = \frac{100}{|\mathcal{I}|} \sum_{i \in \mathcal{I}} s_i.
\end{gathered}
\label{eq:score}
\end{equation}
For example, a solved instance with charge $c_i=4$ uses $u_i=25$\% of its budget, and an unsolved instance has $c_i=16$ and $u_i=100$\%. A set containing only those two instances has $U=62.5$\% and $P=50$\%. For a single instance, $U(\{i\})=u_i$. Because a failure costs the whole budget however it came about, failing is never cheaper than asking, and a demonstration is paid for only by an instance it helped to solve. We take $U$ as the raw measure (lower is better), and quote $P$ beside it, given that neither number determines the other, and the two are complementary.

We keep the feedback binary, as ARC Prize does and as a deployed system experiences it, since returning the correct grid on failure would turn the stream into supervised learning. We also expose the task id on every instance, under an alias drawn for each run, to enable the system to properly disambiguate between tasks, as the input grid is not a unique identifier for the task. See Appendix~\ref{app:bench}.

\subsection{A stream of instances}
\label{sec:stream}

\begin{figure}[t]
\centering
\includegraphics[width=\linewidth]{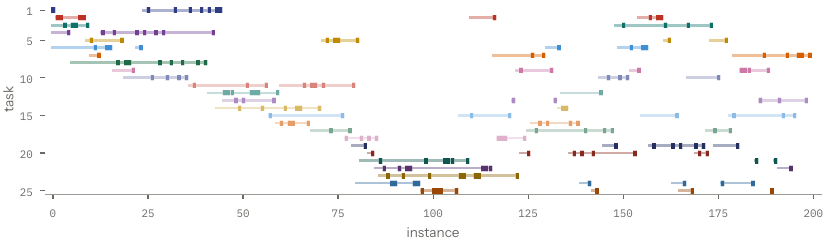}
\caption{The ARC-AGI-1 stream: its 25 tasks, one row each in the order they first appear, and each instance a mark. A thin line in the task's color marks a spell in which the task is active.}
\label{fig:modes}
\end{figure}

A learner meets instances as a stream of consecutive rounds, each naming a task and an instance seed (one per user, Section~\ref{sec:regimes}). At round $t$, an active set $\mathcal{A}_t$ holds at most $n_c$ tasks, from which the round's task $\tau_t$ is drawn uniformly. A task that enters the active set stays for a spell of $\beta$ instances, after which it rests for $\rho$ instances before it may return. Never-seen tasks arrive every $\alpha$ instances and take the next free place:
\begin{equation}
\tau_t \sim \mathrm{Unif}(\mathcal{A}_t), \qquad \beta \sim \mathrm{Exp}(b), \qquad \rho \sim \mathrm{LogNormal}(\ln m, \sigma), \qquad \alpha \sim \mathrm{Exp}(a).
\label{eq:stream}
\end{equation}
We round spells and rests to whole instances, and each is at least one, see Algorithm~\ref{alg:stream} in Appendix~\ref{app:schedules}. The stream keeps a few tasks active at once, each worked in bursts with dormant gaps between them, see Figure~\ref{fig:modes}. A task can stay dormant for long enough to be forgotten, and since every instance is a fresh draw from the task's generator, no answer can be cached.

On ARC-AGI-1, the stream draws 25 of the 400 training tasks and serves 200 fresh, verified instances of them from the Re-ARC generators \citep{hodel2024rearc}, and a held-out test after the stream draws on the 400 evaluation tasks. Four tasks are active at once, spells last about three instances and rests about thirty, and a new task arrives about every four instances, the last in round 97 (Table~\ref{tab:configs} in Appendix~\ref{app:bench}). Every task serves eight instances, after which it is finished. Each ARC-AGI edition has one fixed stream and fixed tests; other seeds draw other streams from the same process (Appendix~\ref{app:schedules}), and all results quoted use a seed of 0. Every system therefore meets the same tasks in the same order, with the same instances. The streams on ARC-AGI-2 and ARC-AGI-3 \citep{chollet2025arcagi2,arcprize2026arcagi3} serve fresh instances from generators of their own. Figures~\ref{fig:plane-arc2} and~\ref{fig:plane-arc3} in Appendix~\ref{app:tracks} are deliberately sparse, leaving these harder tracks open for the community's new methods and submissions, with the directions we find most promising presented in Section~\ref{sec:discussion}.

\subsection{Two tests after the stream}
\label{sec:tests}

The stream measures what a learner keeps while it learns. To measure what it has at the end, we freeze the learner and serve it two tests, each a single round of 100 instances. The learner is not told that a test has begun. We start every test instance from the learner's state at the end of the stream, and nothing learned on one test instance reaches another. The seen test consists of four fresh instances of each of the stream's 25 tasks, at instance seeds the stream did not use, and measures memorization. In the held-out test, which measures generalization, 100 evaluation tasks never seen on the stream are served once each. A learner that keeps one program per task gets cheaper on the seen test but no cheaper on the held-out test, whereas a learner that has become a better solver of new tasks gets cheaper on both. Plotting the two tests against each other (Figure~\ref{fig:plane}) is the continual counterpart of plotting in-distribution against out-of-distribution accuracy \citep{miller2021accuracy}.

\subsection{Measuring learning}
\label{sec:metrics}

$U$ over the stream mixes what a learner could already do before the stream with what it learns along the way. To measure the learning alone, we need one more run, the \emph{reset}: the same learner on the same stream and the same tests, with its state wiped before each instance. Let $U_{0}(\mathcal{I})$ be the reset's budget used over the stream's instances $\mathcal{I}$, and $U^{*}(\mathcal{I})$ that of perfect memory on the same system, which pays the reset's own charge on each task's first exposure and nothing on any return. The learnable budget $\Lambda = U_{0}(\mathcal{I}) - U^{*}(\mathcal{I})$ is the most any memory can save, giving the share learned:
\begin{equation}
L = \frac{U_{0}(\mathcal{I}) - U(\mathcal{I})}{\Lambda}.
\label{eq:learned}
\end{equation}
It is 0 for a learner that keeps nothing, 1 for one that answers every returning task at no cost, and negative for one that does worse than the reset. $L$ therefore measures learning alone, with base capability normalized out. On each of the two tests after the stream, the best a learner can spend is nothing, so we read the same saving as a share of the reset's whole spend on the test's instances $\mathcal{E}$:
\begin{equation}
S = \frac{U_{0}(\mathcal{E}) - U(\mathcal{E})}{U_{0}(\mathcal{E})}.
\label{eq:transfer}
\end{equation}
We write the share saved on the seen test as $S_{\mathrm{seen}}$ and on the held-out test as $S_{\mathrm{held}}$. Taking it as a share of the reset's spend means that a learner whose reset already spends little is not marked down for having less to save. On the seen test, whose tasks the stream has met, $S_{\mathrm{seen}}$ measures memorization. On the held-out test there is nothing to remember, so $S_{\mathrm{held}}$ measures generalization, how much better a learner has become at tasks it never met. The learner plays each whole test \testPlays{} times from where the stream left it, and the reset \testPlays{} times from nothing, on the same instances. Averaging each instance's \testPlays{} plays smooths out the model's random sampling, and $S$ is computed from these averages. We give $L$, $S_{\mathrm{seen}}$ and $S_{\mathrm{held}}$ with 95\% intervals over tasks (Appendix~\ref{app:bench}).

\subsection{Two regimes: the individual and the collective}
\label{sec:regimes}

So far, on the stream, a single system meets one instance at a time, and whatever it carries may change between any two instances. A deployed system may instead serve many people at once, learning from \emph{all} of them, and the collective regime tests such a system. A run has $n_u \in \{1, 2, 4\}$ users, each following a separate stream over 25 non-overlapping tasks, so that two users bring 50 unique tasks and four bring 100. A round serves one instance per user, concurrently, and the system may change its state only between rounds. Any learner can run in either regime, whether it keeps notes or updates its weights. Each user's tasks still arrive and go dormant, and every round is charged (Appendix~\ref{app:schedules}, and Table~\ref{tab:configs} in Appendix~\ref{app:bench}). All users share one state, both the harness's notes and the trained model's weights, so that whatever a learner keeps, it learns from every user's games. With more than one user, the seen test spreads its 100 instances over every user's tasks (Appendix~\ref{app:schedules}).

\subsection{Verifying a result}
\label{sec:tracks}

Every task of the benchmark and every task's solver presented so far is public. External submissions are therefore verified by running the learner's code on a private set of 50 new rules that only we have seen (Appendix~\ref{app:heldout}): 25 for the stream and the seen test, and 25 for the held-out test. For now this verifies a learner at one user, and we will soon add rules for the collective regimes. A learner is verified automatically from an issue on the benchmark's repository. Continual-ARC is designed to explain how a learner learns rather than to purely rank learners, so we give no single score per se. For illustration, reducing budget use from 2\% for the reset to 1\% for the learner saves half the reset's spend, but only one percentage point of the full budget. Reducing it from 100\% to 90\% saves 10\% of the reset's spend, yet still consumes 90\% of the full budget. We therefore present all verified submissions of genuine learning algorithms on the benchmark, with Figure~\ref{fig:plane} showing the canonical representation.

\section{Learners}
\label{sec:baselines}

\begin{figure}[t]
\centering
\includegraphics[width=\linewidth]{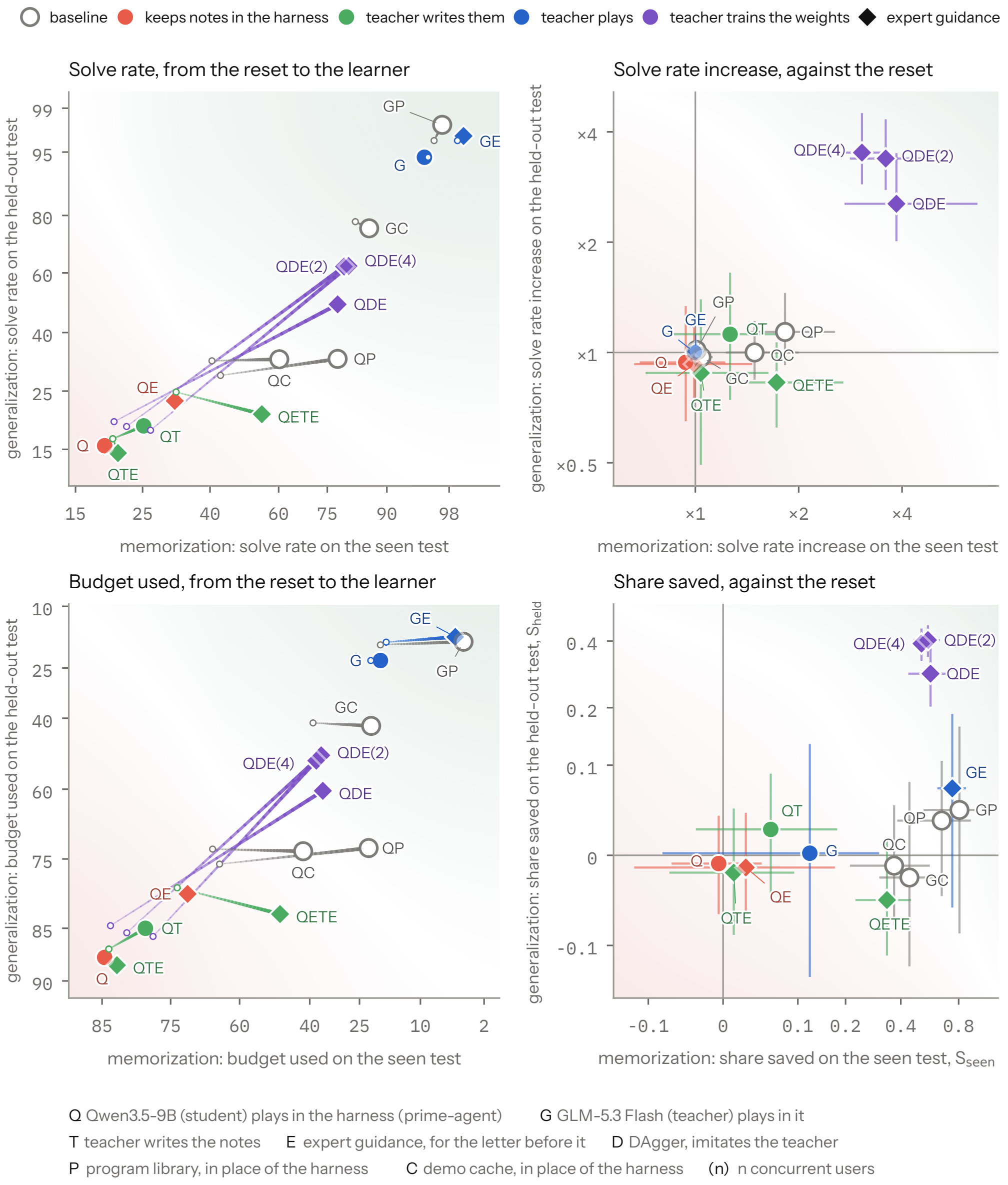}
\caption{At one user, with DAgger also at two and four. Left: lines join each reset (hollow dot) to its learner, on complementary log-log axes. Up and right are better; bars show simultaneous 95\% intervals (Appendix~\ref{app:bench}), hidden where narrower than their point.}
\label{fig:plane}
\end{figure}

Most experiments use Qwen3.5-9B (Q) or GLM-5.3 Flash (G) as learners, from floors and a program library to an off-the-shelf harness, with or without expert guidance, and GLM teaching Qwen by notes or weights (Table~\ref{tab:tests}). We present each learner below, and Figure~\ref{fig:plane} labels them underneath.

\begin{table}[t]
\centering
\footnotesize\setlength{\tabcolsep}{3pt}
\begin{adjustbox}{max width=\linewidth}
\begin{tabular}{@{}l@{\hspace{4pt}}lr@{\hspace{0.3em}}lr@{\hspace{0.3em}}lr@{\hspace{0.3em}}lr@{\hspace{0.3em}}lr@{\hspace{0.3em}}lr@{\hspace{0.3em}}l}
\toprule
 & & \multicolumn{6}{c}{Qwen3.5-9B} & \multicolumn{6}{c}{GLM-5.3 Flash} \\
\cmidrule(lr){3-8}\cmidrule(lr){9-14}
 & method & \multicolumn{2}{c}{$L$} & \multicolumn{2}{c}{$S_{\mathrm{seen}}$} & \multicolumn{2}{c}{$S_{\mathrm{held}}$} & \multicolumn{2}{c}{$L$} & \multicolumn{2}{c}{$S_{\mathrm{seen}}$} & \multicolumn{2}{c}{$S_{\mathrm{held}}$} \\
\midrule
\multirow[t]{2}{*}{\tiny\emph{\begin{tabular}[t]{@{}l@{}}notes in\\the harness\end{tabular}}} & Q, G & -0.02 & {\tiny [-0.09, 0.04]} & -0.01 & {\tiny [-0.04, 0.03]} & -0.01 & {\tiny [-0.05, 0.03]} & +0.12 & {\tiny [-0.16, 0.36]} & +0.12 & {\tiny [-0.03, 0.27]} & 0.00 & {\tiny [-0.11, 0.10]} \\
 & QE, GE & +0.05 & {\tiny [-0.06, 0.16]} & +0.03 & {\tiny [-0.08, 0.13]} & -0.01 & {\tiny [-0.05, 0.03]} & +0.62 & {\tiny [0.49, 0.75]} & +0.74 & {\tiny [0.64, 0.84]} & +0.07 & {\tiny [-0.02, 0.16]} \\
\midrule
\multirow[t]{3}{*}{\tiny\emph{\begin{tabular}[t]{@{}l@{}}notes the\\teacher\\writes\end{tabular}}} & QT & +0.05 & {\tiny [-0.02, 0.14]} & +0.06 & {\tiny [-0.01, 0.14]} & +0.03 & {\tiny [-0.02, 0.07]} & -- & & -- & & -- & \\
 & QTE & +0.07 & {\tiny [-0.03, 0.16]} & +0.01 & {\tiny [-0.05, 0.07]} & -0.02 & {\tiny [-0.07, 0.03]} & -- & & -- & & -- & \\
 & QETE & +0.29 & {\tiny [0.19, 0.40]} & +0.34 & {\tiny [0.26, 0.42]} & -0.05 & {\tiny [-0.10, 0.00]} & -- & & -- & & -- & \\
\midrule
\multirow[t]{1}{*}{\tiny\emph{\begin{tabular}[t]{@{}l@{}}weights\end{tabular}}} & QDE & \textbf{+0.43} & {\tiny [0.33, 0.54]} & +0.57 & {\tiny [0.48, 0.67]} & \textbf{+0.29} & {\tiny [0.23, 0.35]} & -- & & -- & & -- & \\
\midrule
\multirow[t]{2}{*}{\tiny\emph{\begin{tabular}[t]{@{}l@{}}baselines\end{tabular}}} & QP, GP & \textbf{+0.43} & {\tiny [0.25, 0.64]} & \textbf{+0.65} & {\tiny [0.46, 0.86]} & +0.03 & {\tiny [-0.02, 0.09]} & \textbf{+0.68} & {\tiny [0.49, 0.87]} & \textbf{+0.81} & {\tiny [0.59, 1.00]} & +0.05 & {\tiny [-0.06, 0.14]} \\
 & QC, GC & +0.31 & {\tiny [0.19, 0.45]} & +0.37 & {\tiny [0.26, 0.51]} & -0.01 & {\tiny [-0.05, 0.03]} & +0.41 & {\tiny [0.29, 0.53]} & +0.44 & {\tiny [0.34, 0.57]} & -0.02 & {\tiny [-0.11, 0.06]} \\
\bottomrule
\end{tabular}

\end{adjustbox}
\caption{Qwen3.5-9B and GLM-5.3 Flash driving prime-agent: the share learned on the stream, $L$, and the share saved on each test, $S$, against the method's reset (Section~\ref{sec:metrics}; higher is better), with 95\% intervals in small type. Bold marks each model's best where its interval lies above zero.}
\label{tab:tests}
\end{table}

\paragraph{Prompting floors (C\textsubscript{0} and C).}
\emph{Stateless} (C\textsubscript{0}) uses a basic interaction loop with no coding tools or memory between instances. It can request demonstrations and revise wrong answers. \emph{Demo-cache} (C) also stores requested demonstrations by task and automatically replays them for free on later instances. This controls for remembering the demonstrations alone, without learning the rule.

\paragraph{Engineered reference (P).}
The \emph{program library} adds program induction to the demo-cache. It asks the model for a Python program and automatically stores it if it reproduces the demonstrations. On a return, it automatically runs the program first and asks for a repair after a wrong answer. This reference assumes that a task's rule is a grid transformation the demonstrations can verify.

\paragraph{Agent harnesses (Q and G).}
We run prime-agent \citep{karten2026primeagent} with its tools and automatic refinement, opening a fresh conversation per instance. It may store learnings after the final verdict, in hindsight, and only its written state carries over. Rows Q\textsubscript{0} and G\textsubscript{0} wipe that state before every instance, isolating the value of persistence. Rows QA and GA ask it to save liberally after every verdict, and QE and GE add expert guidance, the program library procedure and guided playing rules (Appendix~\ref{app:prompts}), to check, save and reuse a program per task. Appendix~\ref{app:harness} gives the details.

\paragraph{Notes the teacher writes (T).}
In QT, QTE and QETE, GLM plays the full stream in prime-agent and also writes the notes. Qwen then plays each instance from the teacher's state prior to said instance. QT has no expert guidance, QTE gives it to the teacher, and QETE to both. The teacher's play receives uncharged feedback, so these are upper-bound experiments for what GLM can teach Qwen through the harness. Table~\ref{tab:design} in Appendix~\ref{app:bench} sets out the comparison.

\paragraph{Weights (DE).}
DAgger \citep{ross2011dagger} (QDE) updates a LoRA adapter \citep{hu2022lora} of Qwen between rounds, in the same harness as Q (prime-agent). It imitates GLM's replies at states the student encountered, with GLM reading expert guidance withheld from the student. All prior learners run at one user, but DAgger also runs at two and four. Appendix~\ref{app:trained} gives the training schedule and compute, and the unsuccessful policy-gradient and self-distillation variants we tested.

\section{Results}
\label{sec:results}

\begin{figure}[t]
\centering
\includegraphics[width=\linewidth]{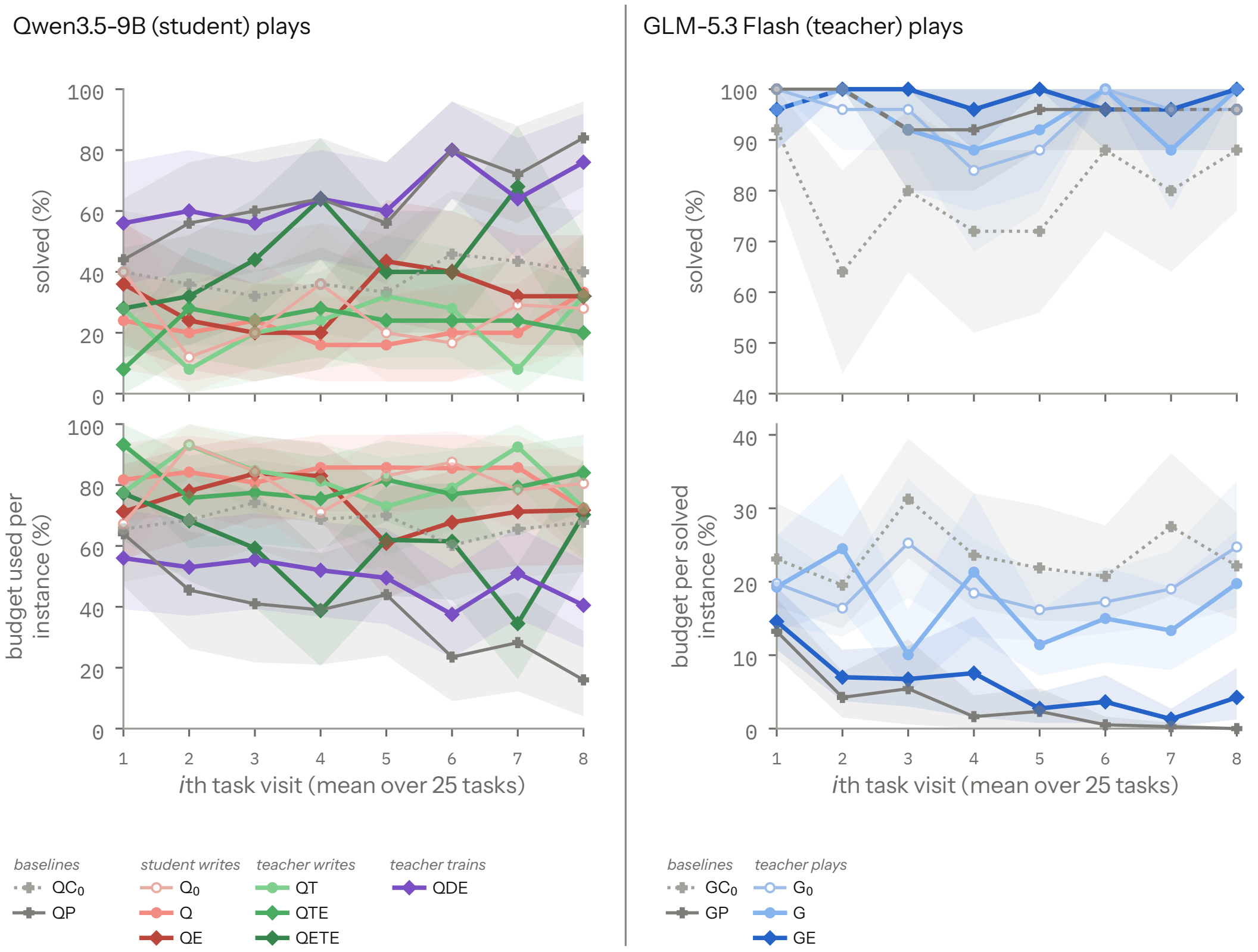}
\caption{At one user. Bands show 95\% intervals over task resamples. GLM's budget is averaged only over instances it solves.}
\label{fig:notes-weights}
\end{figure}

\begin{figure}[t]
\centering
\includegraphics[width=\linewidth]{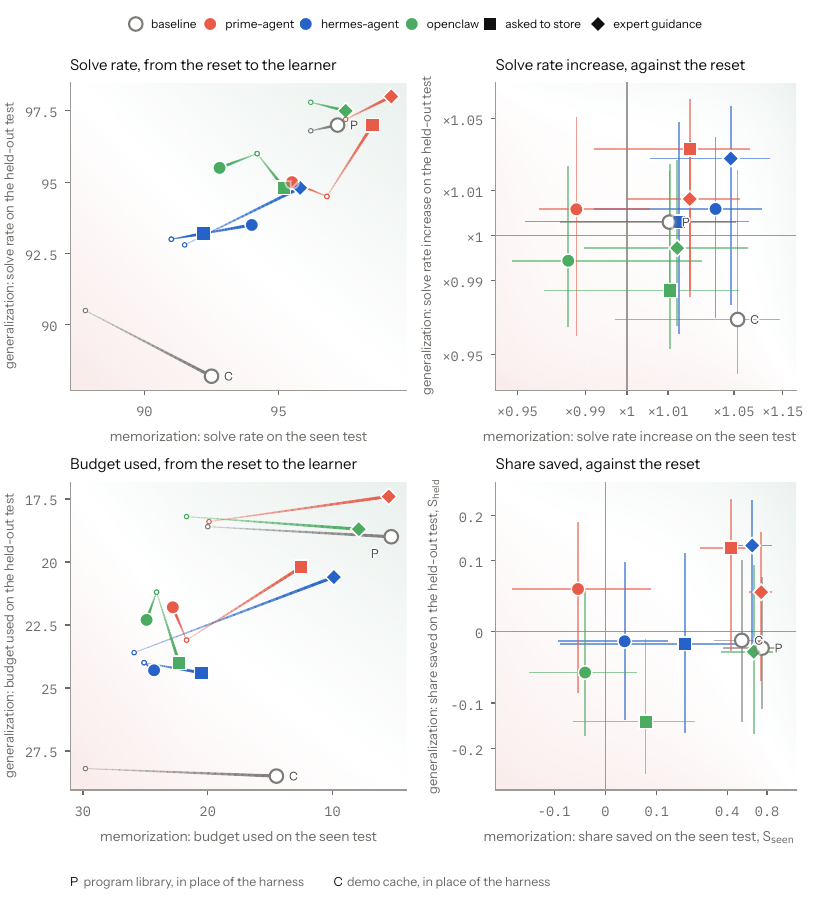}
\caption{GPT-6 Luna at high effort, as in Figure~\ref{fig:plane}. Appendix~\ref{app:results} gives every effort.}
\label{fig:plane-luna}
\end{figure}

We report results on ARC-AGI-1 here, with early results for ARC-AGI-2 and ARC-AGI-3 in Appendix~\ref{app:tracks}. Table~\ref{tab:tests} gives the share learned on the stream, $L$, and the share saved on the two tests, $S_{\mathrm{seen}}$ and $S_{\mathrm{held}}$, for GLM-5.3 Flash (G) and Qwen3.5-9B (Q). Figure~\ref{fig:plane} plots absolute and relative solve rates, and budget used and share saved, $S_{\mathrm{seen}}$ and $S_{\mathrm{held}}$. Figure~\ref{fig:notes-weights} plots each method's share solved and budget by task visit. We also test GPT-6 Luna at high effort in prime-agent \citep{karten2026primeagent}, hermes-agent \citep{nous2026hermesagent} and openclaw \citep{openclaw2026}, see Figure~\ref{fig:plane-luna}. Shares saved below are read at the lower right quadrants of Figures~\ref{fig:plane} and~\ref{fig:plane-luna}.

\paragraph{Baselines.}
Against the stateless floors (QC\textsubscript{0}, GC\textsubscript{0}), the demo-cache floors (QC, GC) saved $S_{\mathrm{seen}} = 0.37$ [0.26, 0.51] and 0.44 [0.34, 0.57] respectively (Figure~\ref{fig:plane}), which is the saving from not paying for the demonstrations twice. On new tasks there was nothing to replay, so the saving was nil, with $S_{\mathrm{held}} = -0.01$ and $-0.02$. The program libraries (QP, GP) beat the demo caches by a wide margin on the seen test, with $S_{\mathrm{seen}} = 0.65$ [0.46, 0.86] and 0.81 [0.59, 1.00] respectively (Appendix~\ref{app:plane-baselines}), but they also saved nothing measurable in $S_{\mathrm{held}}$, with 0.03 [$-0.02$, 0.09] and 0.05 [$-0.06$, 0.14]. This is to be expected, as a program verified on the demonstrations of one task does not solve another.

\paragraph{Harnesses as shipped.}
Left to their own prompting, only G saved measurably on either test across three harnesses, four models, and GPT-6 Luna's three efforts (Figures~\ref{fig:plane} to~\ref{fig:plane-luna} and Appendix~\ref{app:tracks}). This was despite the instructions at the start of every session stating that tasks recur with fresh inputs, that each instance starts a new conversation, and that each demonstration or incorrect attempt costs one (Appendix~\ref{app:prompts}). Neither harness (Q, G) beat its demo-cache floor (QC, GC; Figure~\ref{fig:plane}). Likewise, Q's persistence saved nothing measurable against its reset, with $S_{\mathrm{seen}} = -0.01$ [$-0.04$, 0.03] and $S_{\mathrm{held}} = -0.01$ [$-0.05$, 0.03] (Figure~\ref{fig:plane}), but G's saved a bit, with $S_{\mathrm{seen}} = 0.25$ [0.17, 0.33] over three runs. Further, neither expert guidance (QE) nor asking the harness to save liberally (QA, Appendix~\ref{app:ablations}) saved Q measurably, nor did the teacher's notes (QT and QTE), despite the teacher leaving 106 and 104 notes and files, respectively. For all of these, $S_{\mathrm{seen}}$ lies between 0.01 and 0.06, and $S_{\mathrm{held}}$ between $-0.02$ and 0.03. Even G, which solved nearly every instance, still spent \sbGlmHCost{}\% of the budget per solved instance after every task had been met, against \sbGlmHzeroCost{}\% for G\textsubscript{0} (Figure~\ref{fig:notes-weights}, right), where perfect memory spends nothing. We investigated this: G kept a rule in words for every task in prime-agent's store, but the harness only showed six entries per instance, chosen before the task was known, so the rule reached G on only 40 of 154 returns (Appendix~\ref{app:persist-glm}). To investigate further, we also read what each of the three harnesses (H) stored with GPT-6 Luna driving (Figure~\ref{fig:plane-luna} and Appendix~\ref{app:persist}). prime-agent kept an entry for later sessions only when the agent saved it as global, as its prompt explained, and the agent never did. hermes-agent loaded its skills on every return, but they held general advice which did not affect spend. openclaw never promoted a memory at all, and an earlier correct answer never came back on any of its returns. None of the three harnesses kept the demonstrations it paid for, each buying 2.3 to 3.9 pairs per instance where the demo-cache floor bought 1.4 to 2.6. This suggests that the harnesses' own prompting is not well suited to this domain.

\paragraph{Asking to store.}
Given that several harnesses were failing to store anything, we tried explicitly prompting the harnesses after every verdict to save very liberally anything that could help with the task's future instances (H-A). However, this only made a measurable difference for prime-agent on the seen test. With GPT-6 Luna driving, it saved $S_{\mathrm{seen}} = 0.42$ [0.29, 0.55], while hermes-agent's 0.18 [$-0.01$, 0.39] and openclaw's 0.07 [$-0.02$, 0.17] were within noise of nothing (Figure~\ref{fig:plane-luna}). With GLM driving, it saved 0.26 [0.13, 0.39] (GA), and with Qwen it was also within noise, at 0.02 [$-0.02$, 0.07] (QA). Overall, the harnesses still gained little from persistence, despite the prompting.

\paragraph{Expert guidance.}
To see if we could push the performance of the harnesses further, we added ``expert'' guidance, which explicitly told the harness to write Python programs to solve the tasks, verify each program against the demonstrations, save it for its task and reuse it. With it, GE saved $S_{\mathrm{seen}} = 0.74$ [0.64, 0.84] and the three harnesses on GPT-6 Luna (H-E) saved 0.62 to 0.72 (Figure~\ref{fig:plane-luna}). Q memorized only when the teacher wrote its notes and both were given the guidance (QETE), with $S_{\mathrm{seen}} = 0.34$ [0.26, 0.42]. This shows that the harnesses were fully capable of performing well, and that the failure was a combined lack of judgment of the model and failure of the harness prompts to encourage the correct behavior in this domain.

\paragraph{Generalization.}
The Re-ARC generators derive all 400 ARC-AGI-1 training rules from one library of 160 primitive functions \citep{hodel2024rearc}, and therefore the seen and held-out tasks both share the same building blocks. A harness which learns to abstract and store these shared building blocks during the stream could then theoretically use these to improve performance on the held-out tasks. Despite this, none of the harness methods generalized measurably (Table~\ref{tab:tests}, Figure~\ref{fig:plane-luna} and Appendix~\ref{app:bench}). This includes QETE, in which GLM, given the expert guidance, played the whole stream and wrote every program and note, solving 98\% of the instances, which Qwen failed to put to use on the held-out test. In contrast, training the weights generalized measurably, with DAgger (QDE) saving $S_{\mathrm{held}} = 0.29$ [0.23, 0.35], half its $S_{\mathrm{seen}} = 0.57$ [0.48, 0.67]. Untrained, Qwen actually struggled to use the harness at all, solving only 18 of the 100 held-out tasks in prime-agent (Q\textsubscript{0}) against 31 in the stateless loop (QC\textsubscript{0}). As a result, part of what DAgger learned was simply to drive the harness. This quirk is visible in Figure~\ref{fig:plane} (top left), where the harness method resets sit below those of QC and QP. Providing expert guidance to the untrained Qwen in the harness offsets some of this degradation, lifting its held-out solves from 13--26 to 22--33 per play (Appendix~\ref{app:told}). DAgger, however, solved 47 to 52, beyond both, and so also learned something fundamental about ARC itself, on top of learning to drive the harness and distilling the expert guidance.

\paragraph{More users.}
At two and four users (QDE(2) and QDE(4)), each bringing 25 tasks of their own, DAgger's adapter trained on the games of 50 and 100 tasks. On the held-out test it then saved $S_{\mathrm{held}} = 0.40$ at two users and 0.39 at four (Figure~\ref{fig:plane}). Seemingly, DAgger reached diminishing returns for generalization beyond two users. The full training dynamics are shown in Appendix~\ref{app:trained}.
\FloatBarrier

\section{Related work}
\label{sec:related}

We screened roughly two hundred benchmarks and methods. Table~\ref{tab:related-full} (Appendix~\ref{app:related}) sets the closest against the five properties that this benchmark combines: a fresh instance on every exposure, recurrence after variable gaps, a task set that grows through the stream, binary feedback that is charged, and a protocol open to prompting, memory and weight updates. We found no work that has all five.

\paragraph{ARC.}
\citet{chollet2019measure} notes that ARC's format is close-ended, a weakness that motivates this work, and suggests that a better one would let the test-taker query an example generator and submit answers for feedback repeatedly, with the feedback consumed as the score. To our knowledge, this has not been implemented, although Re-ARC \citep{hodel2024rearc} supplies the generators and verifiers it needs. The strongest ARC-AGI-1 systems instead adapt their weights at test time and discard the adaptation, per task \citep{akyurek2024ttt,franzen2025poe} or over the evaluation set \citep{pourcel2025soar}, never meeting a fresh instance of a task they have adapted to. ArcMemo \citep{ho2025arcmemo} writes concepts to memory, but over a single fixed pass. Over a stream of generated ARC-style tasks, \citet{zhang2026faulty} find that a memory the model keeps rewriting first helps and then degrades. ARC-AGI-3 \citep{arcprize2026arcagi3} moves to learning over time within a single game, and a harness that keeps per-game notes across runs \citep{bedi2026learningmachines} has saturated its public games, untested on others.

\paragraph{Continual learning.}
The classic protocol \citep{lopezpaz2017gradient,vandeven2019three} is a fixed sequence of tasks with free labels, scored on fixed test sets after each task boundary. Where tasks come back at all, they return as a planned second pass \citep{wolczyk2021continual,powers2022cora,veniat2021efficient} or as unannounced revisits \citep{caccia2020online,hemati2023class}. Fresh instances are standard in reinforcement learning environments \citep{cobbe2020procgen,kuttler2020nethack} and in generated reasoning suites with verifiers \citep{stojanovski2025reasoninggym,chen2025enigmata,zeng2026rlve}, but these suites train first and test afterwards, and supervision is free.

\paragraph{Agent benchmarks.}
StreamBench \citep{wu2024streambench} gives binary feedback in one shuffled pass, MemoryBench \citep{ai2025memorybench} feedback from simulated users, LifelongAgentBench \citep{zheng2025lifelongagentbench} brings skills back but not instances, and SWE-Bench-CL \citep{joshi2025swebenchcl} regroups SWE-Bench by repository. CL-Bench \citep{asawa2026clbench} scores a stateful system against a stateless one, and AgentCL \citep{shu2026agentcl} and PATH-Bench \citep{yang2026pathbench} rerun a stream or reinsert a probe task. All draw from a fixed pool of instances that appear once or recur unchanged. None grows the task set, charges for supervision or serves many users, and most evaluate prompting and memory only. Concurrent work \citep{anonymous2026continualautoresearch,zheng2026servelearnbench,zhao2026cle} neither charges for feedback nor trains weights.

\paragraph{Memory against weights.}
The comparison between memory and weights exists only in pieces. Memory-based agents \citep{shinn2023reflexion,zhao2024expel,suzgun2025dynamiccheatsheet,zhou2025memento,nekoei2025jefhinter,zhang2026ace} are motivated by the cost or forgetting of fine-tuning but seldom compare against it, and the closest to our program library keeps a skill as a program once it runs correctly \citep{wang2025asi}. On-policy reinforcement learning is reported to forget less than supervised fine-tuning \citep{shenfeld2025rlsrazor,chen2026retaining}, but to forget severely over a sequence of tasks \citep{luo2026rlforgets}, as does harness learning as tasks drift \citep{chen2026selfevolving}. SEAL \citep{zweiger2025seal} runs self-directed weight updates on ARC, test-time curricula \citep{hubotter2025testtimecurricula} train on tasks chosen for the target, and self-improving harnesses \citep{karten2026primeagent} rewrite their memory and skills, but each is evaluated on its own terms. Concurrently, Self-Direction \citep{kansal2026selfdirection} distills an agent's reflections into a LoRA adapter, on task sets recurring at most three times. Memory and weights have yet to be compared on tasks that keep returning.
\section{Discussion and Future Work}
\label{sec:discussion}

To the best of our knowledge, Continual-ARC is the first benchmark to compare harness-based and weight-based learners under one score, in a domain where both can memorize and generalize. We have shown that most harnesses as shipped did not learn measurably in this domain. How well they learn in other domains remains open. Simple expert guidance greatly improved how much they memorized (Figure~\ref{fig:plane-luna}), which shows that the default harness prompts were not sufficient to inform the models what to store, though more powerful models may fare better.

Training the weights proved important in this domain, as DAgger learned from a short stream of tasks to solve new ones where in-context learning through the harness did not. Continual-ARC targets the continual learning that one person or a small team meets over a short horizon, from 200 instances rather than a post-training corpus. We are therefore most excited about weight-based algorithms that learn from sparse data, model-based reinforcement learning above all \citep{kaiser2020model}. The DAgger algorithm we used is a relatively simple form of weight-based learning, as it needs a stronger teacher whose reasoning tokens are available. Teacher-free reinforcement learning from the sparse reward \citep{shao2024deepseekmath}, on-policy self-distillation \citep{shenfeld2026sdft,hubotter2026sdpo} and on-policy distillation \citep{agarwal2024gkd} are natural next steps (Appendix~\ref{app:other-weights} reports our first attempts). Trust-region updates \citep{schulman2015trpo} would also keep such a learner from degrading what its users receive, which is essential in real deployments.

One weakness of this study is that most of its runs used relatively small models on ARC-AGI-1, the easiest of our three tracks. We leave it to future work to test stronger models on the ARC-AGI-2 and ARC-AGI-3 tracks (Appendix~\ref{app:tracks}). This will let us investigate how much guidance frontier models need in order to store what matters in a harness, and how efficiently they learn it in their weights, both for memorization and generalization. In the long run, we expect the harness to become less the system itself and more the infrastructure that trains the inner model, continually distilling content into the weights and staying thin rather than accumulating an ever-growing list of files.

\section*{AI use statement}
\label{sec:ai-use}

We used an AI coding assistant (Claude, Anthropic) throughout this work. It drafted the abstract and every section of this paper from \leadauthor's notes, which record each design decision, the reason for it, the objection a reviewer would raise, and the answer. \Leadauthor{} edited every draft, and then rewrote the main text line by line. It carried out the literature search behind Section~\ref{sec:related} and Appendix~\ref{app:related}, screening roughly two hundred works. \Leadauthor{} checked every entry in the tables against the work's own abstract, and every comparative claim in the paper against those notes. It proposed parts of the experimental grid of Section~\ref{sec:baselines}, and \leadauthor{} decided on them. It wrote most of the code in the repositories that we release, including the ARC-AGI-2 and ARC-AGI-3 environments, the harness adapters and DAgger's training, under tests and smoke runs that \leadauthor{} reviewed. It also launched and monitored the runs on remote machines at \leadauthor's direction. Finally, it computed the analyses of Section~\ref{sec:results} and Appendices~\ref{app:persist} and~\ref{app:results} from the run records, including the paired comparisons and their intervals, and it read the harnesses' stored state for Appendix~\ref{app:persist}. Every table and figure is regenerated from the run directories when the paper is built. Scripts in the paper's repository recompute the numbers quoted in Section~\ref{sec:results} and the paired comparisons of Appendices~\ref{app:persist} and~\ref{app:ablations}.

No AI was used to generate data. Every instance in the stream comes from the Re-ARC generators, ARC-GEN's generators or our level generators for the ARC-AGI game engine. The rules of the private set are the exception (Appendix~\ref{app:heldout}). Six agents on Claude Opus 5.5, one for each family of rules, wrote the 98 candidates as generators and solvers in code, and models calibrated them, Claude Opus 5.5, GPT-6 Luna, GLM-5.3 Flash and Qwen3.5-9B answering fresh instances at first sight. The filters, the screens and the selection that kept 50 of them are code written with the assistant, and no person has tried the rules. No AI was used to prove claims, and the equations in Section~\ref{sec:benchmark} are definitions. We take responsibility for the final content of this work, including text, claims or artifacts produced with the aid of generative AI.

\section*{Ethics statement}
\label{sec:ethics}

This work involves no human subjects and no personal data. Every public task of the benchmark comes from the public ARC-AGI tasks and games or from the Re-ARC and ARC-GEN generators, which the software fetches from their sources and does not redistribute. ARC-AGI-3 levels are charged against the human baselines that ARC Prize publishes. The rules of the private set are our own, and we do not release them (Appendix~\ref{app:heldout}).

The harnesses we evaluate write and run their own code, and one reached the evaluation itself in an early run. Section~\ref{sec:baselines} and Appendix~\ref{app:harness} describe how we confine each one, and we recommend the same to anyone running agents on the benchmark.

Our findings are negative for three publicly available harnesses, none of them from \leadauthor's employer, which builds agent systems. We run each as its authors ship it, with its persistence switched on, and we trace each shortfall to a specific step in its design (Appendix~\ref{app:persist}). Appendix~\ref{app:harness} gives each one's configuration and reports that prime-agent's Python skills, \texttt{refine} among them, did not load in our runs. The adapters are released with the code, so that the harnesses' authors can check how their systems were run.

Continual learning in deployment also raises questions of privacy that this benchmark does not address. In the collective regime, one system learns from many users at once, and what it keeps from one user could reach another. Our streams are abstract puzzles with no user data, and the benchmark does not measure leakage between users.

\section*{Reproducibility statement}
\label{sec:reproducibility}

Section~\ref{sec:benchmark} defines the streams and their measures, which the benchmark computes from a run's records, the share learned from a run and the reset's, and Appendices~\ref{app:bench}, \ref{app:schedules} and~\ref{app:tracks} give every stream's parameters and the process that generates each stream (Algorithm~\ref{alg:stream}). Streams are deterministic, so every system meets the same instances in the same order. Appendix~\ref{app:prompts} gives every text that a learner is shown, or the file in the code that holds it, and Appendix~\ref{app:harness} describes how we run each harness, with the versions of every component. Appendix~\ref{sec:models} and Appendix~\ref{app:bench} give the model presets and the rule that chose them, and Appendix~\ref{app:trained} gives DAgger's settings. Every adapter DAgger played with is kept, so any round's student can be played again. The code of the benchmark and of every learner, including the harness adapters, is released with the records of every run this paper reports.  The rules of the private set are withheld by design, since a released rule could be learned before a run, and Appendix~\ref{app:heldout} gives how they were made and chosen. Each verified run is requested on the benchmark repository's public issues, and its numbers are posted there.  The models are sampled through provider APIs and on our own servers, so a rerun will differ from ours by about the spread we report, such as how far chance moves $S$, which the learners that keep nothing between instances show (Section~\ref{sec:results}).

\section*{Acknowledgments}

We thank Nassim Berrada, Dhruva Bhardwaj, Mark Chen, Aadya Chinubhai, Mateusz Dziemian, Yasser Mahmoud, Ved Patwardhan, Saturnin Pugnet, Charles Songhurst, Shreyas Sudhanva, Anna Tzatzopoulou, Daniel Warfield, Phil Winder and Jiahan Xie for comments on an early draft.

\bibliography{references}

\begin{thebibliography}{98}
\providecommand{\natexlab}[1]{#1}
\providecommand{\url}[1]{\texttt{#1}}
\expandafter\ifx\csname urlstyle\endcsname\relax
  \providecommand{\doi}[1]{doi: #1}\else
  \providecommand{\doi}{doi: \begingroup \urlstyle{rm}\Url}\fi

\bibitem[{Adaptive Agent Team} et~al.(2023){Adaptive Agent Team}, Bauer,
  Baumli, Baveja, Behbahani, Bhoopchand, Bradley-Schmieg, Chang, Clay,
  Collister, Dasagi, Gonzalez, Gregor, Hughes, Kashem, Loks-Thompson, Openshaw,
  Parker-Holder, Pathak, Perez-Nieves, Rakicevic, Rockt{\"a}schel, Schroecker,
  Sygnowski, Tuyls, York, Zacherl, and Zhang]{ada2023humantimescale}
{Adaptive Agent Team}, Jakob Bauer, Kate Baumli, Satinder Baveja, Feryal
  Behbahani, Avishkar Bhoopchand, Nathalie Bradley-Schmieg, Michael Chang,
  Natalie Clay, Adrian Collister, Vibhavari Dasagi, Lucy Gonzalez, Karol
  Gregor, Edward Hughes, Sheleem Kashem, Maria Loks-Thompson, Hannah Openshaw,
  Jack Parker-Holder, Shreya Pathak, Nicolas Perez-Nieves, Nemanja Rakicevic,
  Tim Rockt{\"a}schel, Yannick Schroecker, Jakub Sygnowski, Karl Tuyls, Sarah
  York, Alexander Zacherl, and Lei Zhang.
\newblock Human-timescale adaptation in an open-ended task space.
\newblock \emph{arXiv preprint arXiv:2301.07608}, 2023.

\bibitem[Agarwal et~al.(2024)Agarwal, Vieillard, Zhou, Stanczyk, Ramos, Geist,
  and Bachem]{agarwal2024gkd}
Rishabh Agarwal, Nino Vieillard, Yongchao Zhou, Piotr Stanczyk, Sabela Ramos,
  Matthieu Geist, and Olivier Bachem.
\newblock On-policy distillation of language models: Learning from
  self-generated mistakes.
\newblock In \emph{International Conference on Learning Representations}, 2024.

\bibitem[Ai et~al.(2025)Ai, Tang, Wang, Long, Su, and Liu]{ai2025memorybench}
Qingyao Ai, Yichen Tang, Changyue Wang, Jianming Long, Weihang Su, and Yiqun
  Liu.
\newblock {MemoryBench}: A benchmark for memory and continual learning in {LLM}
  systems.
\newblock \emph{arXiv preprint arXiv:2510.17281}, 2025.

\bibitem[Aky{\"u}rek et~al.(2024)Aky{\"u}rek, Damani, Zweiger, Qiu, Guo, Pari,
  Kim, and Andreas]{akyurek2024ttt}
Ekin Aky{\"u}rek, Mehul Damani, Adam Zweiger, Linlu Qiu, Han Guo, Jyothish
  Pari, Yoon Kim, and Jacob Andreas.
\newblock The surprising effectiveness of test-time training for few-shot
  learning.
\newblock \emph{arXiv preprint arXiv:2411.07279}, 2024.
\newblock v1 titled ``The Surprising Effectiveness of Test-Time Training for
  Abstract Reasoning''.

\bibitem[Alford et~al.(2021)Alford, Gandhi, Rangamani, Banburski, Wang,
  Dandekar, Chin, Poggio, and Chin]{alford2021bidirectional}
Simon Alford, Anshula Gandhi, Akshay Rangamani, Andrzej Banburski, Tony Wang,
  Sylee Dandekar, John Chin, Tomaso Poggio, and Peter Chin.
\newblock Neural-guided, bidirectional program search for abstraction and
  reasoning.
\newblock In \emph{Complex Networks \& Their Applications X}, 2021.
\newblock arXiv:2110.11536.

\bibitem[Anonymous(2026{\natexlab{a}})]{anonymous2026comate}
Anonymous.
\newblock {CoMATE}: Consolidating memory across trajectories for long-horizon
  interactive exploration on {ARC-AGI-3}.
\newblock Submitted to the International Conference on Learning Representations
  (ICLR 2027), under review, 2026{\natexlab{a}}.
\newblock URL \url{https://openreview.net/forum?id=X9BcfhZsuB}.

\bibitem[Anonymous(2026{\natexlab{b}})]{anonymous2026compiledworlds}
Anonymous.
\newblock {Compiled Worlds}: What adaptation costs on transformed {ARC-AGI-3}
  games.
\newblock Submitted to the International Conference on Learning Representations
  (ICLR 2027), under review, 2026{\natexlab{b}}.
\newblock URL \url{https://openreview.net/forum?id=8u4GwD0Jwm}.

\bibitem[Anonymous(2026{\natexlab{c}})]{anonymous2026continualautoresearch}
Anonymous.
\newblock From {AutoResearch} to {Continual AutoResearch}: Do agents learn
  across tasks?
\newblock Submitted to the International Conference on Learning Representations
  (ICLR 2027), under review, 2026{\natexlab{c}}.
\newblock URL \url{https://openreview.net/forum?id=OPHp9Aei9U}.

\bibitem[{ARC Prize Foundation}(2026{\natexlab{a}})]{arcprize2026arcagi3}
{ARC Prize Foundation}.
\newblock {ARC-AGI-3}: A new challenge for frontier agentic intelligence.
\newblock \emph{arXiv preprint arXiv:2603.24621}, 2026{\natexlab{a}}.

\bibitem[{ARC Prize Foundation}(2026{\natexlab{b}})]{arcprize2026leaderboard}
{ARC Prize Foundation}.
\newblock {ARC-AGI} leaderboard.
\newblock \url{https://arcprize.org/leaderboard}, 2026{\natexlab{b}}.
\newblock Kimi K3's results on ARC-AGI-2 as of 31 July 2026, read 6 November
  2026.

\bibitem[{Artificial Analysis}(2026)]{artificialanalysis2026index}
{Artificial Analysis}.
\newblock {GPT-6 Luna} (high) vs {GLM-5.3-Flash}: model comparison.
\newblock
  \url{https://artificialanalysis.ai/models/comparisons/gpt-6-luna-high-vs-glm-5-3-flash},
  2026.
\newblock Artificial Analysis Intelligence Index v4.3.2, accessed 26 October
  2026.

\bibitem[Asawa et~al.(2026)Asawa, Glaze, Orlanski, Ramakrishnan, Xu, Biswal,
  Chen, Sala, Zaharia, and Gonzalez]{asawa2026clbench}
Parth Asawa, Christopher~M. Glaze, Gabriel Orlanski, Ramya Ramakrishnan, Benji
  Xu, Asim Biswal, Vincent~Sunn Chen, Frederic Sala, Matei Zaharia, and
  Joseph~E. Gonzalez.
\newblock Continual learning bench: Evaluating frontier {AI} systems in
  real-world stateful environments.
\newblock \emph{arXiv preprint arXiv:2606.05661}, 2026.

\bibitem[Bedi(2026)]{bedi2026learningmachines}
Ashpreet Bedi.
\newblock Learning machines: the memory that cracked {ARC-AGI-3} public set.
\newblock Agno blog, 2026.
\newblock URL \url{https://www.agno.com/articles/learning-machines}.

\bibitem[Bornschein et~al.(2022)Bornschein, Galashov, Hemsley, Rannen-Triki,
  Chen, Chaudhry, He, Douillard, Caccia, Feng, Shen, Rebuffi, Stacpoole, de~las
  Casas, Hawkins, Lazaridou, Teh, Rusu, Pascanu, and
  Ranzato]{bornschein2022nevis}
J{\"o}rg Bornschein, Alexandre Galashov, Ross Hemsley, Amal Rannen-Triki,
  Yutian Chen, Arslan Chaudhry, Xu~Owen He, Arthur Douillard, Massimo Caccia,
  Qixuang Feng, Jiajun Shen, Sylvestre-Alvise Rebuffi, Kitty Stacpoole, Diego
  de~las Casas, Will Hawkins, Angeliki Lazaridou, Yee~Whye Teh, Andrei~A. Rusu,
  Razvan Pascanu, and Marc'Aurelio Ranzato.
\newblock {NEVIS}'22: A stream of 100 tasks sampled from 30 years of computer
  vision research.
\newblock \emph{arXiv preprint arXiv:2211.11747}, 2022.

\bibitem[Butt et~al.(2024)Butt, Manczak, Wiggers, Rainone, Zhang, Defferrard,
  and Cohen]{butt2024codeit}
Natasha Butt, Blazej Manczak, Auke Wiggers, Corrado Rainone, David~W. Zhang,
  Micha{\"e}l Defferrard, and Taco Cohen.
\newblock {CodeIt}: Self-improving language models with prioritized hindsight
  replay.
\newblock In \emph{International Conference on Machine Learning}, 2024.
\newblock arXiv:2402.04858.

\bibitem[Caccia et~al.(2020)Caccia, Rodriguez, Ostapenko, Normandin, Lin,
  Caccia, Laradji, Rish, Lacoste, Vazquez, and Charlin]{caccia2020online}
Massimo Caccia, Pau Rodriguez, Oleksiy Ostapenko, Fabrice Normandin, Min Lin,
  Lucas Caccia, Issam Laradji, Irina Rish, Alexandre Lacoste, David Vazquez,
  and Laurent Charlin.
\newblock Online fast adaptation and knowledge accumulation ({OSAKA}): a new
  approach to continual learning.
\newblock In \emph{Advances in Neural Information Processing Systems 33
  (NeurIPS)}, 2020.
\newblock arXiv:2003.05856.

\bibitem[Cai et~al.(2025)Cai, Hao, Zhou, Yan, Lei, Zhen, Han, Yang, Li, Pan,
  Huai, Chen, Li, Chen, Zhang, Qiu, and He]{cai2025ell}
Yuxuan Cai, Yipeng Hao, Jie Zhou, Hang Yan, Zhikai Lei, Rui Zhen, Zhenhua Han,
  Yutao Yang, Junsong Li, Qianjun Pan, Tianyu Huai, Qin Chen, Xin Li, Kai Chen,
  Bo~Zhang, Xipeng Qiu, and Liang He.
\newblock Building self-evolving agents via experience-driven lifelong
  learning: A framework and benchmark.
\newblock \emph{arXiv preprint arXiv:2508.19005}, 2025.

\bibitem[Chaudhry et~al.(2019)Chaudhry, Ranzato, Rohrbach, and
  Elhoseiny]{chaudhry2019efficient}
Arslan Chaudhry, Marc'Aurelio Ranzato, Marcus Rohrbach, and Mohamed Elhoseiny.
\newblock Efficient lifelong learning with a-gem.
\newblock In \emph{International Conference on Learning Representations
  (ICLR)}, 2019.
\newblock arXiv:1812.00420.

\bibitem[Chen et~al.(2026{\natexlab{a}})Chen, Razin, Narasimhan, and
  Chen]{chen2026retaining}
Howard Chen, Noam Razin, Karthik Narasimhan, and Danqi Chen.
\newblock Retaining by doing: The role of on-policy data in mitigating
  forgetting.
\newblock In \emph{International Conference on Machine Learning},
  2026{\natexlab{a}}.
\newblock arXiv:2510.18874.

\bibitem[Chen et~al.(2025)Chen, He, Yuan, Chen, Cai, Dai, Yu, Yu, Li, Chen,
  Zhou, and Wang]{chen2025enigmata}
Jiangjie Chen, Qianyu He, Siyu Yuan, Aili Chen, Zhicheng Cai, Weinan Dai,
  Hongli Yu, Qiying Yu, Xuefeng Li, Jiaze Chen, Hao Zhou, and Mingxuan Wang.
\newblock Enigmata: Scaling logical reasoning in large language models with
  synthetic verifiable puzzles.
\newblock In \emph{Advances in Neural Information Processing Systems}, 2025.
\newblock arXiv:2505.19914.

\bibitem[Chen et~al.(2026{\natexlab{b}})Chen, Huang, Xu, Zhao, McQuinn, Liu,
  and Yao]{chen2026selfevolving}
Xiaoyin Chen, Yizheng Huang, Yinglun Xu, Ziyun Zhao, Emmett McQuinn, Julie Liu,
  and Hope Yao.
\newblock Are self-evolving agents continual learners?
\newblock \emph{arXiv preprint arXiv:2610.25858}, 2026{\natexlab{b}}.

\bibitem[Chevalier-Boisvert et~al.(2019)Chevalier-Boisvert, Bahdanau, Lahlou,
  Willems, Saharia, Nguyen, and Bengio]{chevalierboisvert2019babyai}
Maxime Chevalier-Boisvert, Dzmitry Bahdanau, Salem Lahlou, Lucas Willems,
  Chitwan Saharia, Thien~Huu Nguyen, and Yoshua Bengio.
\newblock Babyai: A platform to study the sample efficiency of grounded
  language learning.
\newblock In \emph{International Conference on Learning Representations}, 2019.
\newblock arXiv:1810.08272.

\bibitem[Chollet(2019)]{chollet2019measure}
Fran{\c{c}}ois Chollet.
\newblock On the measure of intelligence.
\newblock \emph{arXiv preprint arXiv:1911.01547}, 2019.

\bibitem[Chollet et~al.(2025)Chollet, Knoop, Kamradt, Landers, and
  Pinkard]{chollet2025arcagi2}
Fran{\c{c}}ois Chollet, Mike Knoop, Gregory Kamradt, Bryan Landers, and Henry
  Pinkard.
\newblock {ARC-AGI-2}: A new challenge for frontier {AI} reasoning systems.
\newblock \emph{arXiv preprint arXiv:2505.11831}, 2025.

\bibitem[Chollet et~al.(2026)Chollet, Knoop, Kamradt, and
  Landers]{chollet2026arcprize2025}
Fran{\c{c}}ois Chollet, Mike Knoop, Gregory Kamradt, and Bryan Landers.
\newblock {ARC Prize 2025}: Technical report.
\newblock \emph{arXiv preprint arXiv:2601.10904}, 2026.

\bibitem[Cobbe et~al.(2020)Cobbe, Hesse, Hilton, and
  Schulman]{cobbe2020procgen}
Karl Cobbe, Christopher Hesse, Jacob Hilton, and John Schulman.
\newblock Leveraging procedural generation to benchmark reinforcement learning.
\newblock In \emph{Proceedings of the 37th International Conference on Machine
  Learning (ICML)}, 2020.
\newblock arXiv:1912.01588.

\bibitem[Dohare et~al.(2024)Dohare, Hernandez-Garcia, Lan, Rahman, Mahmood, and
  Sutton]{dohare2024loss}
Shibhansh Dohare, J.~Fernando Hernandez-Garcia, Qingfeng Lan, Parash Rahman,
  A.~Rupam Mahmood, and Richard~S. Sutton.
\newblock Loss of plasticity in deep continual learning.
\newblock \emph{Nature}, 632\penalty0 (8026):\penalty0 768--774, 2024.
\newblock \doi{10.1038/s41586-024-07711-7}.

\bibitem[Franzen et~al.(2025)Franzen, Disselhoff, and Hartmann]{franzen2025poe}
Daniel Franzen, Jan Disselhoff, and David Hartmann.
\newblock Product of experts with {LLMs}: Boosting performance on {ARC} is a
  matter of perspective.
\newblock In \emph{Proceedings of the 42nd International Conference on Machine
  Learning (ICML)}, 2025.
\newblock arXiv:2505.07859.

\bibitem[Goodfellow et~al.(2013)Goodfellow, Mirza, Xiao, Courville, and
  Bengio]{goodfellow2013empirical}
Ian~J. Goodfellow, Mehdi Mirza, Da~Xiao, Aaron Courville, and Yoshua Bengio.
\newblock An empirical investigation of catastrophic forgetting in
  gradient-based neural networks.
\newblock \emph{arXiv preprint arXiv:1312.6211}, 2013.

\bibitem[Guan et~al.(2026)Guan, Wang, Yang, Cao, Liu, Hu, Li, and
  Zhang]{guan2026continualskillbench}
Tianyi Guan, Yiding Wang, Haotong Yang, Siyuan Cao, Shirui Liu, Yi~Hu, Jiaqi
  Li, and Muhan Zhang.
\newblock {ContinualSkillBench}: Can {LLM} agents truly evolve their
  capabilities?
\newblock \emph{arXiv preprint arXiv:2608.03874}, 2026.

\bibitem[Guo et~al.(2025)Guo, Hou, Sun, He, Chen, Zhou, Jia, Wang, and
  Chua]{guo2025mllmctbench}
Haiyun Guo, Zhiyan Hou, Yandu Sun, Jinghan He, Yu~Chen, Yuzhe Zhou, Yuheng Jia,
  Jinqiao Wang, and Tat-Seng Chua.
\newblock {MLLM-CTBench}: A benchmark for continual instruction tuning with
  reasoning process diagnosis.
\newblock \emph{arXiv preprint arXiv:2508.08275}, 2025.

\bibitem[Hemati et~al.(2023)Hemati, Cossu, Carta, Hurtado, Pellegrini, Bacciu,
  Lomonaco, and Borth]{hemati2023class}
Hamed Hemati, Andrea Cossu, Antonio Carta, Julio Hurtado, Lorenzo Pellegrini,
  Davide Bacciu, Vincenzo Lomonaco, and Damian Borth.
\newblock Class-incremental learning with repetition.
\newblock In \emph{Proceedings of the 2nd Conference on Lifelong Learning
  Agents (CoLLAs)}, 2023.
\newblock arXiv:2301.11396.

\bibitem[Hemati et~al.(2025)Hemati, Pellegrini, Duan, Zhao, Xia, Masana,
  Tscheschner, Veas, Zheng, Zhao, Li, Huang, Lomonaco, and van~de
  Ven]{hemati2025repetition}
Hamed Hemati, Lorenzo Pellegrini, Xiaotian Duan, Zixuan Zhao, Fangfang Xia,
  Marc Masana, Benedikt Tscheschner, Eduardo Veas, Yuxiang Zheng, Shiji Zhao,
  Shao-Yuan Li, Sheng-Jun Huang, Vincenzo Lomonaco, and Gido~M. van~de Ven.
\newblock Continual learning in the presence of repetition.
\newblock \emph{Neural Networks}, 2025.
\newblock arXiv:2405.04101; CLVision 2023 challenge report.

\bibitem[Ho et~al.(2025)Ho, Si, Feng, Yu, et~al.]{ho2025arcmemo}
Matthew Ho, Chen Si, Zhaoxiang Feng, Fangxu Yu, et~al.
\newblock {ArcMemo}: Abstract reasoning composition with lifelong {LLM} memory.
\newblock \emph{arXiv preprint arXiv:2509.04439}, 2025.

\bibitem[Hodel(2023)]{hodel2023arcdsl}
Michael Hodel.
\newblock Domain specific language for the abstraction and reasoning corpus
  ({ARC-DSL}).
\newblock \url{https://github.com/michaelhodel/arc-dsl}, 2023.

\bibitem[Hodel(2024)]{hodel2024rearc}
Michael Hodel.
\newblock Addressing the abstraction and reasoning corpus via procedural
  example generation.
\newblock \emph{arXiv preprint arXiv:2404.07353}, 2024.

\bibitem[Hu et~al.(2022)Hu, Shen, Wallis, Allen-Zhu, Li, Wang, Wang, and
  Chen]{hu2022lora}
Edward~J. Hu, Yelong Shen, Phillip Wallis, Zeyuan Allen-Zhu, Yuanzhi Li, Shean
  Wang, Lu~Wang, and Weizhu Chen.
\newblock {LoRA}: Low-rank adaptation of large language models.
\newblock In \emph{International Conference on Learning Representations
  (ICLR)}, 2022.

\bibitem[H{\"u}botter et~al.(2025)H{\"u}botter, Diaz-Bone, Hakimi, Krause, and
  Hardt]{hubotter2025testtimecurricula}
Jonas H{\"u}botter, Leander Diaz-Bone, Ido Hakimi, Andreas Krause, and Moritz
  Hardt.
\newblock Learning on the job: Test-time curricula for targeted reinforcement
  learning.
\newblock \emph{arXiv preprint arXiv:2510.04786}, 2025.

\bibitem[H{\"u}botter et~al.(2026)H{\"u}botter, L{\"u}beck, Behric, Baumann,
  Bagatella, Marta, Hakimi, Shenfeld, Kleine~Buening, Guestrin, and
  Krause]{hubotter2026sdpo}
Jonas H{\"u}botter, Frederike L{\"u}beck, Lejs Behric, Anton Baumann, Marco
  Bagatella, Daniel Marta, Ido Hakimi, Idan Shenfeld, Thomas Kleine~Buening,
  Carlos Guestrin, and Andreas Krause.
\newblock Reinforcement learning via self-distillation.
\newblock \emph{arXiv preprint arXiv:2601.20802}, 2026.

\bibitem[Jiang et~al.(2026)Jiang, Ma, Hong, Wang, Lu, Wang, Chen, Zhang, Pan,
  Li, Liang, and Xiao]{jiang2026seaeval}
Sihang Jiang, Lipeng Ma, Zhonghua Hong, Keyi Wang, Zhiyu Lu, Tengfei Wang,
  Shisong Chen, Jinghao Zhang, Tianjun Pan, Weijia Li, Jiaqing Liang, and
  Yanghua Xiao.
\newblock {SEA-Eval}: A benchmark for evaluating self-evolving agents beyond
  episodic assessment.
\newblock \emph{arXiv preprint arXiv:2604.08988}, 2026.

\bibitem[Joshi et~al.(2025)Joshi, Chowdhury, and Uysal]{joshi2025swebenchcl}
Thomas Joshi, Shayan Chowdhury, and Fatih Uysal.
\newblock {SWE-Bench-CL}: Continual learning for coding agents.
\newblock \emph{arXiv preprint arXiv:2507.00014}, 2025.

\bibitem[Kaiser et~al.(2020)Kaiser, Babaeizadeh, Mi{\l}os, Osi{\'n}ski,
  Campbell, Czechowski, Erhan, Finn, Kozakowski, Levine, Mohiuddin, Sepassi,
  Tucker, and Michalewski]{kaiser2020model}
{\L}ukasz Kaiser, Mohammad Babaeizadeh, Piotr Mi{\l}os, B{\l}a{\.z}ej
  Osi{\'n}ski, Roy~H. Campbell, Konrad Czechowski, Dumitru Erhan, Chelsea Finn,
  Piotr Kozakowski, Sergey Levine, Afroz Mohiuddin, Ryan Sepassi, George
  Tucker, and Henryk Michalewski.
\newblock Model-based reinforcement learning for atari.
\newblock In \emph{International Conference on Learning Representations
  (ICLR)}, 2020.
\newblock arXiv:1903.00374; defines the Atari 100k protocol.

\bibitem[Kansal et~al.(2026)Kansal, Ullrich, Kirichenko, Kulikov, Jha,
  Pezeshki, and Ibrahim]{kansal2026selfdirection}
Yuval Kansal, Karen Ullrich, Polina Kirichenko, Ilia Kulikov, Niraj~K. Jha,
  Mohammad Pezeshki, and Mark Ibrahim.
\newblock Don't forget what you just learned: Self-directed repair turns agent
  mistakes into lasting lessons.
\newblock \emph{arXiv preprint arXiv:2610.31846}, 2026.

\bibitem[Karten et~al.(2026)Karten, Zhang, Thomas, M{\"u}ller, Bakouch, Auras,
  Senghaas, Obeid, Dunas, Hagemann, and Jaghouar]{karten2026primeagent}
Seth Karten, Alex~L. Zhang, Kevin Thomas, Sebastian M{\"u}ller, Elie Bakouch,
  Daniel Auras, Mika Senghaas, Fares Obeid, Konstantin Dunas, Johannes
  Hagemann, and Sami Jaghouar.
\newblock Prime agent: A self-improving {RLM} harness.
\newblock \emph{arXiv preprint arXiv:2608.23552}, 2026.

\bibitem[K{\"u}ttler et~al.(2020)K{\"u}ttler, Nardelli, Miller, Raileanu,
  Selvatici, Grefenstette, and Rockt{\"a}schel]{kuttler2020nethack}
Heinrich K{\"u}ttler, Nantas Nardelli, Alexander~H. Miller, Roberta Raileanu,
  Marco Selvatici, Edward Grefenstette, and Tim Rockt{\"a}schel.
\newblock The nethack learning environment.
\newblock In \emph{Advances in Neural Information Processing Systems 33
  (NeurIPS)}, 2020.
\newblock arXiv:2006.13760.

\bibitem[Lee et~al.(2024)Lee, Kim, Lee, Hwang, Lee, Lee, and Kim]{lee2024arcle}
Hosung Lee, Sejin Kim, Seungpil Lee, Sanha Hwang, Jihwan Lee, Byung-Jun Lee,
  and Sundong Kim.
\newblock {ARCLE}: The abstraction and reasoning corpus learning environment
  for reinforcement learning.
\newblock In \emph{Conference on Lifelong Learning Agents (CoLLAs)}, 2024.
\newblock arXiv:2407.20806.

\bibitem[Lehmann et~al.(2026)Lehmann, Khushbakht, Salehfard, Nishat, Bhandiwad,
  Aioanei, and Vahdati]{lehmann2026arctgi}
Jens Lehmann, Syeda Khushbakht, Nikoo Salehfard, Nur A~Zarin Nishat, Dhananjay
  Bhandiwad, Andrei Aioanei, and Sahar Vahdati.
\newblock {ARC-TGI}: Human-validated task generators with reasoning chain
  templates for {ARC-AGI}.
\newblock In \emph{Proceedings of the 32nd ACM SIGKDD Conference on Knowledge
  Discovery and Data Mining V.2}, 2026.
\newblock arXiv:2603.05099.

\bibitem[Lin et~al.(2022)Lin, Shi, Pathak, and Ramanan]{lin2022clear}
Zhiqiu Lin, Jia Shi, Deepak Pathak, and Deva Ramanan.
\newblock The clear benchmark: Continual learning on real-world imagery.
\newblock In \emph{Advances in Neural Information Processing Systems (NeurIPS)
  Datasets and Benchmarks Track}, 2022.
\newblock arXiv:2201.06289.

\bibitem[Liu et~al.(2023)Liu, Zhu, Gao, Feng, Liu, Zhu, and
  Stone]{liu2023libero}
Bo~Liu, Yifeng Zhu, Chongkai Gao, Yihao Feng, Qiang Liu, Yuke Zhu, and Peter
  Stone.
\newblock {LIBERO}: Benchmarking knowledge transfer for lifelong robot
  learning.
\newblock In \emph{Advances in Neural Information Processing Systems (NeurIPS)
  Datasets and Benchmarks Track}, 2023.
\newblock arXiv:2306.03310.

\bibitem[Lomonaco \& Maltoni(2017)Lomonaco and Maltoni]{lomonaco2017core50}
Vincenzo Lomonaco and Davide Maltoni.
\newblock Core50: a new dataset and benchmark for continuous object
  recognition.
\newblock In \emph{Proceedings of the 1st Conference on Robot Learning (CoRL)},
  2017.
\newblock arXiv:1705.03550.

\bibitem[Lopez-Paz \& Ranzato(2017)Lopez-Paz and Ranzato]{lopezpaz2017gradient}
David Lopez-Paz and Marc'Aurelio Ranzato.
\newblock Gradient episodic memory for continual learning.
\newblock In \emph{Advances in Neural Information Processing Systems 30
  (NeurIPS)}, 2017.
\newblock arXiv:1706.08840.

\bibitem[Luo et~al.(2026)Luo, Wang, Zhou, Ye, Zhao, Zhang, and
  Wei]{luo2026rlforgets}
Mao-Lin Luo, Zhe-Xu Wang, Zi-Hao Zhou, Bo~Ye, Jian Zhao, Min-Ling Zhang, and
  Tong Wei.
\newblock Rl forgets! towards continual policy optimization.
\newblock \emph{arXiv preprint arXiv:2607.04364}, 2026.

\bibitem[Miller et~al.(2021)Miller, Taori, Raghunathan, Sagawa, Koh, Shankar,
  Liang, Carmon, and Schmidt]{miller2021accuracy}
John Miller, Rohan Taori, Aditi Raghunathan, Shiori Sagawa, Pang~Wei Koh,
  Vaishaal Shankar, Percy Liang, Yair Carmon, and Ludwig Schmidt.
\newblock Accuracy on the line: on the strong correlation between
  out-of-distribution and in-distribution generalization.
\newblock In \emph{International Conference on Machine Learning (ICML)}, 2021.
\newblock arXiv:2107.04649.

\bibitem[Moffitt(2025)]{moffitt2025arcgen}
Michael~D. Moffitt.
\newblock {ARC-GEN}: A mimetic procedural benchmark generator for the
  abstraction and reasoning corpus.
\newblock \emph{arXiv preprint arXiv:2511.00162}, 2025.

\bibitem[Nekoei et~al.(2025)Nekoei, Jaiswal, Bechard, Shliazhko, Marquez~Ayala,
  Reymond, Caccia, Drouin, Chandar, and Lacoste]{nekoei2025jefhinter}
Hadi Nekoei, Aman Jaiswal, Patrice Bechard, Oleh Shliazhko, Orlando
  Marquez~Ayala, Mathieu Reymond, Massimo Caccia, Alexandre Drouin, Sarath
  Chandar, and Alexandre Lacoste.
\newblock {JEF-Hinter}: Leveraging offline knowledge for improving web agents
  adaptation.
\newblock \emph{arXiv preprint arXiv:2510.04373}, 2025.

\bibitem[New et~al.(2022)New, Baker, Nguyen, and Vallabha]{new2022lifelong}
Alexander New, Megan Baker, Eric Nguyen, and Gautam Vallabha.
\newblock Lifelong learning metrics.
\newblock \emph{arXiv preprint arXiv:2201.08278}, 2022.

\bibitem[{Nous Research}(2026)]{nous2026hermesagent}
{Nous Research}.
\newblock Hermes agent.
\newblock \url{https://github.com/NousResearch/hermes-agent}, 2026.
\newblock Version 0.20.2, accessed September 2026.

\bibitem[{Open Ended Learning Team} et~al.(2021){Open Ended Learning Team},
  Stooke, Mahajan, Barros, Deck, Bauer, Sygnowski, Trebacz, Jaderberg, Mathieu,
  McAleese, Bradley-Schmieg, Wong, Porcel, Raileanu, Hughes-Fitt, Dalibard, and
  Czarnecki]{oel2021xland}
{Open Ended Learning Team}, Adam Stooke, Anuj Mahajan, Catarina Barros, Charlie
  Deck, Jakob Bauer, Jakub Sygnowski, Maja Trebacz, Max Jaderberg, Michael
  Mathieu, Nat McAleese, Nathalie Bradley-Schmieg, Nathaniel Wong, Nicolas
  Porcel, Roberta Raileanu, Steph Hughes-Fitt, Valentin Dalibard, and
  Wojciech~Marian Czarnecki.
\newblock Open-ended learning leads to generally capable agents.
\newblock \emph{arXiv preprint arXiv:2107.12808}, 2021.

\bibitem[{OpenClaw contributors}(2026)]{openclaw2026}
{OpenClaw contributors}.
\newblock {OpenClaw}.
\newblock \url{https://openclaw.ai}, 2026.
\newblock Version 2026.9.5, accessed September 2026.

\bibitem[Platanios et~al.(2020)Platanios, Saparov, and
  Mitchell]{platanios2020jelly}
Emmanouil~Antonios Platanios, Abulhair Saparov, and Tom Mitchell.
\newblock Jelly bean world: A testbed for never-ending learning.
\newblock In \emph{International Conference on Learning Representations
  (ICLR)}, 2020.
\newblock arXiv:2002.06306.

\bibitem[Pourcel et~al.(2025)Pourcel, Colas, and Oudeyer]{pourcel2025soar}
Julien Pourcel, C{\'e}dric Colas, and Pierre-Yves Oudeyer.
\newblock Self-improving language models for evolutionary program synthesis: A
  case study on {ARC-AGI}.
\newblock In \emph{Proceedings of the 42nd International Conference on Machine
  Learning (ICML)}, 2025.
\newblock arXiv:2507.14172.

\bibitem[Powers et~al.(2022)Powers, Xing, Kolve, Mottaghi, and
  Gupta]{powers2022cora}
Sam Powers, Eliot Xing, Eric Kolve, Roozbeh Mottaghi, and Abhinav Gupta.
\newblock {CORA}: Benchmarks, baselines, and metrics as a platform for
  continual reinforcement learning agents.
\newblock In \emph{Proceedings of The 1st Conference on Lifelong Learning
  Agents (CoLLAs)}, volume 199 of \emph{Proceedings of Machine Learning
  Research}, pp.\  705--743, 2022.
\newblock arXiv:2110.10067.

\bibitem[Ross et~al.(2011)Ross, Gordon, and Bagnell]{ross2011dagger}
St{\'e}phane Ross, Geoffrey~J. Gordon, and J.~Andrew Bagnell.
\newblock A reduction of imitation learning and structured prediction to
  no-regret online learning.
\newblock In \emph{Proceedings of the Fourteenth International Conference on
  Artificial Intelligence and Statistics (AISTATS)}, 2011.
\newblock arXiv:1011.0686.

\bibitem[Schulman et~al.(2015)Schulman, Levine, Moritz, Jordan, and
  Abbeel]{schulman2015trpo}
John Schulman, Sergey Levine, Philipp Moritz, Michael Jordan, and Pieter
  Abbeel.
\newblock Trust region policy optimization.
\newblock In \emph{Proceedings of the 32nd International Conference on Machine
  Learning}, volume~37 of \emph{Proceedings of Machine Learning Research}, pp.\
   1889--1897, 2015.

\bibitem[Schulman et~al.(2017)Schulman, Wolski, Dhariwal, Radford, and
  Klimov]{schulman2017ppo}
John Schulman, Filip Wolski, Prafulla Dhariwal, Alec Radford, and Oleg Klimov.
\newblock Proximal policy optimization algorithms.
\newblock \emph{arXiv preprint arXiv:1707.06347}, 2017.

\bibitem[Shao et~al.(2024)Shao, Wang, Zhu, Xu, Song, Bi, Zhang, Zhang, Li, Wu,
  and Guo]{shao2024deepseekmath}
Zhihong Shao, Peiyi Wang, Qihao Zhu, Runxin Xu, Junxiao Song, Xiao Bi, Haowei
  Zhang, Mingchuan Zhang, Y.~K. Li, Y.~Wu, and Daya Guo.
\newblock Deepseekmath: Pushing the limits of mathematical reasoning in open
  language models.
\newblock \emph{arXiv preprint arXiv:2402.03300}, 2024.

\bibitem[Shenfeld et~al.(2025)Shenfeld, Pari, and
  Agrawal]{shenfeld2025rlsrazor}
Idan Shenfeld, Jyothish Pari, and Pulkit Agrawal.
\newblock Rl's razor: Why online reinforcement learning forgets less.
\newblock \emph{arXiv preprint arXiv:2509.04259}, 2025.

\bibitem[Shenfeld et~al.(2026)Shenfeld, Damani, H{\"u}botter, and
  Agrawal]{shenfeld2026sdft}
Idan Shenfeld, Mehul Damani, Jonas H{\"u}botter, and Pulkit Agrawal.
\newblock Self-distillation enables continual learning.
\newblock \emph{arXiv preprint arXiv:2601.19897}, 2026.

\bibitem[Shinn et~al.(2023)Shinn, Cassano, Berman, Gopinath, Narasimhan, and
  Yao]{shinn2023reflexion}
Noah Shinn, Federico Cassano, Edward Berman, Ashwin Gopinath, Karthik
  Narasimhan, and Shunyu Yao.
\newblock Reflexion: Language agents with verbal reinforcement learning.
\newblock In \emph{Advances in Neural Information Processing Systems}, 2023.
\newblock arXiv:2303.11366.

\bibitem[Shu et~al.(2026)Shu, Jim{\'e}nez~Guti{\'e}rrez, Jonnalagedda, Yao,
  Sun, and Su]{shu2026agentcl}
Yiheng Shu, Bernal Jim{\'e}nez~Guti{\'e}rrez, Saisri~Padmaja Jonnalagedda,
  Yuguang Yao, Huan Sun, and Yu~Su.
\newblock {AgentCL}: Toward rigorous evaluation of continual learning in
  language agents.
\newblock \emph{arXiv preprint arXiv:2606.02461}, 2026.

\bibitem[Stojanovski et~al.(2025)Stojanovski, Stanley, Sharratt, Jones,
  Adefioye, Kaddour, and K{\"o}pf]{stojanovski2025reasoninggym}
Zafir Stojanovski, Oliver Stanley, Joe Sharratt, Richard Jones, Abdulhakeem
  Adefioye, Jean Kaddour, and Andreas K{\"o}pf.
\newblock Reasoning gym: Reasoning environments for reinforcement learning with
  verifiable rewards.
\newblock In \emph{Advances in Neural Information Processing Systems}, 2025.
\newblock arXiv:2505.24760.

\bibitem[Suzgun et~al.(2025)Suzgun, Yuksekgonul, Bianchi, Jurafsky, and
  Zou]{suzgun2025dynamiccheatsheet}
Mirac Suzgun, Mert Yuksekgonul, Federico Bianchi, Dan Jurafsky, and James Zou.
\newblock Dynamic cheatsheet: Test-time learning with adaptive memory.
\newblock \emph{arXiv preprint arXiv:2504.07952}, 2025.

\bibitem[Tomilin et~al.(2023)Tomilin, Fang, Zhang, and
  Pechenizkiy]{tomilin2023coom}
Tristan Tomilin, Meng Fang, Yudi Zhang, and Mykola Pechenizkiy.
\newblock {COOM}: A game benchmark for continual reinforcement learning.
\newblock In \emph{Advances in Neural Information Processing Systems 36
  (NeurIPS) Datasets and Benchmarks Track}, 2023.
\newblock URL
  \url{https://proceedings.neurips.cc/paper_files/paper/2023/hash/d61d9f4fe4357296cb658795fd7999f0-Abstract-Datasets_and_Benchmarks.html}.

\bibitem[Tomilin et~al.(2026)Tomilin, van~den Boogaard, Garcin, Ruhdorfer,
  Grooten, Kusters, Du, Bulling, Pechenizkiy, and Fang]{tomilin2026meal}
Tristan Tomilin, Luka van~den Boogaard, Samuel Garcin, Constantin Ruhdorfer,
  Bram Grooten, Fabrice Kusters, Yali Du, Andreas Bulling, Mykola Pechenizkiy,
  and Meng Fang.
\newblock {MEAL}: A benchmark for continual multi-agent reinforcement learning.
\newblock In \emph{Proceedings of the 43rd International Conference on Machine
  Learning (ICML)}, 2026.
\newblock arXiv:2506.14990.

\bibitem[van~de Ven \& Tolias(2019)van~de Ven and Tolias]{vandeven2019three}
Gido~M. van~de Ven and Andreas~S. Tolias.
\newblock Three scenarios for continual learning.
\newblock \emph{arXiv preprint arXiv:1904.07734}, 2019.

\bibitem[Veniat et~al.(2021)Veniat, Denoyer, and Ranzato]{veniat2021efficient}
Tom Veniat, Ludovic Denoyer, and Marc'Aurelio Ranzato.
\newblock Efficient continual learning with modular networks and task-driven
  priors.
\newblock In \emph{International Conference on Learning Representations
  (ICLR)}, 2021.
\newblock arXiv:2012.12631; introduces the CTrL benchmark.

\bibitem[Wang et~al.(2023)Wang, Zhang, Chen, Gao, Jin, Yang, Xi, Zheng, Zou,
  Gui, Zhang, and Huang]{wang2023trace}
Xiao Wang, Yuansen Zhang, Tianze Chen, Songyang Gao, Senjie Jin, Xianjun Yang,
  Zhiheng Xi, Rui Zheng, Yicheng Zou, Tao Gui, Qi~Zhang, and Xuanjing Huang.
\newblock {TRACE}: A comprehensive benchmark for continual learning in large
  language models.
\newblock \emph{arXiv preprint arXiv:2310.06762}, 2023.

\bibitem[Wang et~al.(2025)Wang, Gandhi, Neubig, and Fried]{wang2025asi}
Zora~Zhiruo Wang, Apurva Gandhi, Graham Neubig, and Daniel Fried.
\newblock Inducing programmatic skills for agentic tasks.
\newblock \emph{arXiv preprint arXiv:2504.06821}, 2025.

\bibitem[Wei et~al.(2025)Wei, Sachdeva, Coleman, He, Bei, Ning, Ai, Li, He,
  Chi, Wang, Chen, Pereira, Kang, and Cheng]{wei2025evomemory}
Tianxin Wei, Noveen Sachdeva, Benjamin Coleman, Zhankui He, Yuanchen Bei,
  Xuying Ning, Mengting Ai, Yunzhe Li, Jingrui He, Ed~H. Chi, Chi Wang, Shuo
  Chen, Fernando Pereira, Wang-Cheng Kang, and Derek~Zhiyuan Cheng.
\newblock {Evo-Memory}: Benchmarking {LLM} agent test-time learning with
  self-evolving memory.
\newblock \emph{arXiv preprint arXiv:2511.20857}, 2025.

\bibitem[Wo{\l}czyk et~al.(2021)Wo{\l}czyk, Zaj{\k{a}}c, Pascanu, Kuci{\'n}ski,
  and Mi{\l}o{\'s}]{wolczyk2021continual}
Maciej Wo{\l}czyk, Micha{\l} Zaj{\k{a}}c, Razvan Pascanu, {\L}ukasz
  Kuci{\'n}ski, and Piotr Mi{\l}o{\'s}.
\newblock Continual world: A robotic benchmark for continual reinforcement
  learning.
\newblock In \emph{Advances in Neural Information Processing Systems 34
  (NeurIPS)}, 2021.
\newblock arXiv:2105.10919.

\bibitem[Wu et~al.(2024)Wu, Tam, Lin, Chen, and Lee]{wu2024streambench}
Cheng-Kuang Wu, Zhi~Rui Tam, Chieh-Yen Lin, Yun-Nung Chen, and Hung-yi Lee.
\newblock {StreamBench}: Towards benchmarking continuous improvement of
  language agents.
\newblock In \emph{Advances in Neural Information Processing Systems 37
  (NeurIPS 2024), Datasets and Benchmarks Track}, 2024.
\newblock arXiv:2406.08747.

\bibitem[Wu(2026)]{wu2026kepler}
Wensen Wu.
\newblock {Kepler}: Auditable world models for {ARC-AGI-3}.
\newblock \emph{arXiv preprint arXiv:2610.00834}, 2026.

\bibitem[Xu et~al.(2026)Xu, Li, Wu, Lan, Li, Zhou, Jiang, Wang, Wang, Luu,
  Xiong, Park, Hooi, and Hu]{xu2026evoarena}
Jundong Xu, Qingchuan Li, Jiaying Wu, Yihuai Lan, Shuyue~Stella Li, Huichi
  Zhou, Bowen Jiang, Lei Wang, Jun Wang, Anh~Tuan Luu, Caiming Xiong, Hae~Won
  Park, Bryan Hooi, and Zhiyuan Hu.
\newblock {EvoArena}: Tracking memory evolution for robust {LLM} agents in
  dynamic environments.
\newblock \emph{arXiv preprint arXiv:2606.13681}, 2026.

\bibitem[Yan et~al.(2026)Yan, Liang, Hu, He, Yuan, Zhang, and
  Tan]{yan2026agentstream}
Dong Yan, Jian Liang, Dapeng Hu, Ran He, Nicholas~Jing Yuan, Qi~Zhang, and
  Tieniu Tan.
\newblock {AgentStream}: How well do self-evolving {LLM} agents perform under
  streaming tasks?
\newblock \emph{arXiv preprint arXiv:2608.00155}, 2026.

\bibitem[Yang et~al.(2026)Yang, Zhang, Li, Liu, Sheng, Hua, Yin, Fang, Shen,
  and Wang]{yang2026pathbench}
Xidong Yang, Xingyi Zhang, Wenhao Li, Wenyan Liu, Junjie Sheng, Yun Hua, Wei
  Yin, Tao Fang, Chuyun Shen, and Xiangfeng Wang.
\newblock {PATH-Bench}: Path-dependent evaluation of lifelong agents.
\newblock \emph{arXiv preprint arXiv:2608.01149}, 2026.

\bibitem[Zeng et~al.(2026)Zeng, Ivison, Wang, Yuan, Li, Ye, Li, He, Zhou, Chen,
  Zhao, Tsvetkov, Du, Jaques, Peng, Koh, and Hajishirzi]{zeng2026rlve}
Zhiyuan Zeng, Hamish Ivison, Yiping Wang, Lifan Yuan, Shuyue~Stella Li, Zhuorui
  Ye, Siting Li, Jacqueline He, Runlong Zhou, Tong Chen, Chenyang Zhao, Yulia
  Tsvetkov, Simon~Shaolei Du, Natasha Jaques, Hao Peng, Pang~Wei Koh, and
  Hannaneh Hajishirzi.
\newblock Rlve: Scaling up reinforcement learning for language models with
  adaptive verifiable environments.
\newblock In \emph{International Conference on Machine Learning}, 2026.
\newblock arXiv:2511.07317.

\bibitem[Zenke et~al.(2017)Zenke, Poole, and Ganguli]{zenke2017continual}
Friedemann Zenke, Ben Poole, and Surya Ganguli.
\newblock Continual learning through synaptic intelligence.
\newblock In \emph{Proceedings of the 34th International Conference on Machine
  Learning (ICML)}, 2017.
\newblock arXiv:1703.04200.

\bibitem[Zhang et~al.(2026{\natexlab{a}})Zhang, Lin, Wu, Sun, Li, Li, and
  Peng]{zhang2026faulty}
Dylan Zhang, Yanshan Lin, Zhengkun Wu, Yihang Sun, Bingxuan Li, Dianqi Li, and
  Hao Peng.
\newblock Useful memories become faulty when continuously updated by {LLMs}.
\newblock In \emph{Advances in Neural Information Processing Systems
  (NeurIPS)}, 2026{\natexlab{a}}.
\newblock arXiv:2605.12978.

\bibitem[Zhang et~al.(2026{\natexlab{b}})Zhang, Hu, Upasani, Ma, Hong,
  Kamanuru, Rainton, Wu, Ji, Li, Thakker, Zou, and Olukotun]{zhang2026ace}
Qizheng Zhang, Changran Hu, Shubhangi Upasani, Boyuan Ma, Fenglu Hong,
  Vamsidhar Kamanuru, Jay Rainton, Chen Wu, Mengmeng Ji, Hanchen Li, Urmish
  Thakker, James Zou, and Kunle Olukotun.
\newblock Agentic context engineering: Evolving contexts for self-improving
  language models.
\newblock In \emph{International Conference on Learning Representations},
  2026{\natexlab{b}}.
\newblock arXiv:2510.04618.

\bibitem[Zhang et~al.(2023)Zhang, Fang, Chen, and Namazi-Rad]{zhang2023citb}
Zihan Zhang, Meng Fang, Ling Chen, and Mohammad-Reza Namazi-Rad.
\newblock {CITB}: A benchmark for continual instruction tuning.
\newblock In \emph{Findings of the Association for Computational Linguistics:
  EMNLP 2023}, 2023.
\newblock arXiv:2310.14510.

\bibitem[Zhao et~al.(2024)Zhao, Huang, Xu, Lin, Liu, and Huang]{zhao2024expel}
Andrew Zhao, Daniel Huang, Quentin Xu, Matthieu Lin, Yong-Jin Liu, and Gao
  Huang.
\newblock Expel: Llm agents are experiential learners.
\newblock In \emph{Proceedings of the AAAI Conference on Artificial
  Intelligence}, 2024.
\newblock arXiv:2308.10144.

\bibitem[Zhao et~al.(2026{\natexlab{a}})Zhao, Tu, Huang, Zhou, Yin, Huang, Xu,
  Shi, Shi, Zhang, Zhang, Chan, Hu, Sheng, Luo, Chen, Yao, Fang, and
  Wang]{zhao2026cle}
Haoyu Zhao, Sitian Tu, Yuxuan Huang, Huichi Zhou, Zhangyue Yin, Jiahao Huang,
  Jingyuan Xu, Tianmiao Shi, Jin Shi, Nianxi Zhang, Yangmiao Zhang, Zhuolang
  Chan, Huiyan Hu, Chenchao Sheng, Yaqing Luo, Yue Chen, Huaxiu Yao, Meng Fang,
  and Jun Wang.
\newblock {Continual Learning Exam}: Benchmarking long-horizon {RSI} and
  experiential learning in {LLM} agents.
\newblock \emph{arXiv preprint arXiv:2610.33956}, 2026{\natexlab{a}}.

\bibitem[Zhao et~al.(2026{\natexlab{b}})Zhao, Yu, He, Fang, Tutunov, Bou-Ammar,
  Luo, and Wang]{zhao2026memento3}
Haoyu Zhao, Zhengxu Yu, Zhiyuan He, Meng Fang, Rasul Tutunov, Haitham
  Bou-Ammar, Weilin Luo, and Jun Wang.
\newblock {Memento 3}: Model-based recursive self-improvement through
  reflective rulebooks.
\newblock \emph{arXiv preprint arXiv:2610.11794}, 2026{\natexlab{b}}.

\bibitem[Zheng et~al.(2026)Zheng, Di, Sadhukhan, Jin, and
  Chen]{zheng2026servelearnbench}
Haizhong Zheng, Yizhuo Di, Ranajoy Sadhukhan, Shuowei Jin, and Beidi Chen.
\newblock {ServeLearnBench}: How well can agents self-improve from serving
  experience?
\newblock \emph{arXiv preprint arXiv:2610.07792}, 2026.

\bibitem[Zheng et~al.(2025)Zheng, Cai, Li, Zhang, Li, Zhang, Song, and
  Ma]{zheng2025lifelongagentbench}
Junhao Zheng, Xidi Cai, Qiuke Li, Duzhen Zhang, ZhongZhi Li, Yingying Zhang,
  Le~Song, and Qianli Ma.
\newblock {LifelongAgentBench}: Evaluating {LLM} agents as lifelong learners.
\newblock \emph{arXiv preprint arXiv:2505.11942}, 2025.

\bibitem[Zhong et~al.(2026)Zhong, Lu, Ning, Wan, Feng, Ao, Ribeiro, Dreyer,
  Ammirati, and Xiong]{zhong2026skilllearnbench}
Shanshan Zhong, Yi~Lu, Jingjie Ning, Yibing Wan, Lihan Feng, Yuyi Ao, Leonardo
  F.~R. Ribeiro, Markus Dreyer, Sean Ammirati, and Chenyan Xiong.
\newblock {SkillLearnBench}: Benchmarking continual learning methods for agent
  skill generation on real-world tasks.
\newblock \emph{arXiv preprint arXiv:2604.20087}, 2026.

\bibitem[Zhou et~al.(2025)Zhou, Chen, Guo, Yan, Lee, Wang, Lee, Zhang, Shao,
  Yang, and Wang]{zhou2025memento}
Huichi Zhou, Yihang Chen, Siyuan Guo, Xue Yan, Kin~Hei Lee, Zihan Wang, Ka~Yiu
  Lee, Guchun Zhang, Kun Shao, Linyi Yang, and Jun Wang.
\newblock Memento: Fine-tuning llm agents without fine-tuning llms.
\newblock \emph{arXiv preprint arXiv:2508.16153}, 2025.

\bibitem[Zweiger et~al.(2025)Zweiger, Pari, Guo, Aky{\"u}rek, Kim, and
  Agrawal]{zweiger2025seal}
Adam Zweiger, Jyothish Pari, Han Guo, Ekin Aky{\"u}rek, Yoon Kim, and Pulkit
  Agrawal.
\newblock Self-adapting language models.
\newblock \emph{arXiv preprint arXiv:2506.10943}, 2025.

\end{thebibliography}
\bibliographystyle{preprint}

\appendix
\AtBeginEnvironment{table}{\setlength{\belowcaptionskip}{6pt}}
\renewcommand{\topfraction}{0.95}\renewcommand{\bottomfraction}{0.95}
\renewcommand{\textfraction}{0.05}\renewcommand{\floatpagefraction}{0.85}
\setcounter{topnumber}{3}\setcounter{bottomnumber}{2}\setcounter{totalnumber}{4}
\section{Benchmark details}
\label{app:bench}

\paragraph{ARC-AGI-1.} We draw the stream from the 400 ARC-AGI-1 training tasks through the Re-ARC generators \citep{hodel2024rearc}, pinned by commit. Given a task id and a seed, the generator produces an input grid and the verifier produces its output, and we keep only the pairs that the verifier accepts, discarding any demonstration whose input equals the test input. Each demonstration request draws one more such pair. A learner cannot cache outputs, since every instance is fresh, and caching demonstrations has a price. The held-out test draws on the 400 ARC-AGI-1 evaluation tasks, which have no generators, and each of them appears once, with its original demonstrations available on request. A request returns them one at a time, two to seven of them and three or four for most tasks of the track, and nothing after them.

\paragraph{The held-out tasks are public.} The held-out test's 100 tasks come from ARC-AGI-1's public evaluation set, and 90 of them are also among ARC-AGI-2's training tasks, so a model may have met them in training. Each learner is read against itself reset, which met them as often, so what it saves there is what the stream added, not what the model brought.

\paragraph{Why the held-out test can show transfer.} ARC tasks differ in their rules but share the operations from which the rules are built. The Re-ARC generators \citep{hodel2024rearc}, from which the ARC-AGI-1 stream is drawn, implement all 400 training rules with one library of 160 primitive functions, and 53 of these primitives appear in ten or more of the generators, among them filling cells, shifting objects and filtering sets of cells (\texttt{scripts/rearc\_primitives.py}). The same library reaches the evaluation tasks, on which the held-out test draws: it is the domain-specific language of \citet{hodel2023arcdsl}, designed from the training tasks alone, and CodeIt, writing its programs in it, solved 59 of the 400 evaluation tasks \citep{butt2024codeit}. A learner that refactors what it learns into task-specific compositions of reusable low-level primitives, and keeps the primitives, should therefore meet a held-out task with much of the machinery that the task needs already in place. Library learning has shown this on ARC, where abstractions that DreamCoder built from solved tasks let it solve progressively harder ones \citep{alford2021bidirectional}, and concept-level memory distilled from solution traces raised a language model's score over a no-memory baseline \citep{ho2025arcmemo}. None of the learners in this paper keeps such primitives. The program library stores one whole program per task, and the harnesses as shipped store prose or general advice (Appendix~\ref{app:persist}), and so we attribute the absence of transfer from notes and programs in Section~\ref{sec:results} to these learners.

\begin{table}[!h]
\centering
\footnotesize\setlength{\tabcolsep}{4pt}
\begin{tabular}{p{0.2\linewidth}p{0.36\linewidth}p{0.36\linewidth}}
\toprule
 & Kept in the harness's notes & Kept in the student's weights \\
\midrule
Taught by the student & Q: the harness as shipped \newline QE: given expert guidance, the program library's procedure and playing rules & -- \\
\addlinespace
Taught by GLM-5.3 Flash & QT: the teacher plays the stream and writes the notes, and the student plays from them \newline QTE: the teacher given the guidance \newline QETE: both given it & QDE: imitates the teacher's move in each state the student met, the teacher reading expert guidance \\
\bottomrule
\end{tabular}
\caption{The comparison on Qwen3.5-9B. Every method plays the same stream in the same harness, prime-agent, beside its wiped twin and the baselines, and the methods on each line share their teacher. The weights taught by the student alone are not shown (Appendix~\ref{app:other-weights}).}
\label{tab:design}
\end{table}

\begin{figure}[t]
\centering
\includegraphics[width=0.66\linewidth]{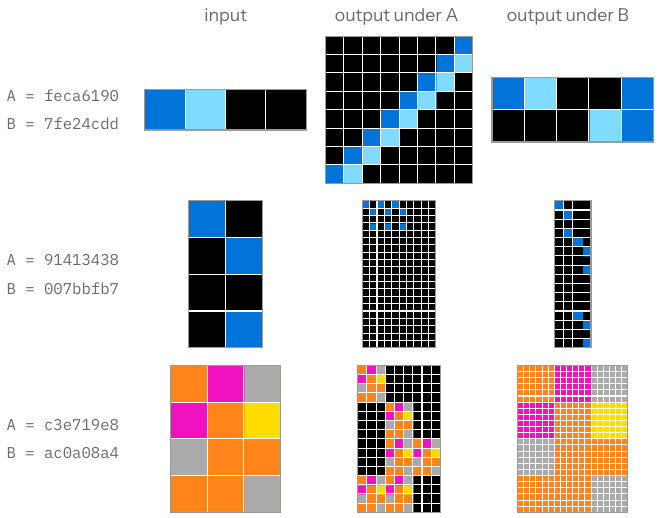}
\caption{An input grid does not determine its task. Each row shows one input generated for training task A, with the output of task A's rule and the output of task B's rule on the same grid. Both are valid ARC tasks, and only the demonstrations or the id say which rule applies. Top: A (\texttt{feca6190}) draws diagonal lines from each colored cell on a larger black square, and B (\texttt{7fe24cdd}) lays out the input and its three rotations as four quadrants. Middle: A (\texttt{91413438}) tiles copies of the input, one per colored cell, and B (\texttt{007bbfb7}) places the input's pattern in every tile whose cell is colored. Bottom: A (\texttt{c3e719e8}) places the whole input in every tile whose cell has the most frequent color. B (\texttt{ac0a08a4}) enlarges every cell into a block. The pairs are selected automatically: an input generated for A on which B's verifier runs and returns a different valid grid.}
\label{fig:taskid}
\end{figure}

\paragraph{Task ids.} In continual learning, exposing the task id corresponds to the task-oracle setting. We expose it because recognizing that today's grid is the same kind of problem as one from last month is a separate skill from remembering how to solve it, and mixing the two would make the measurement harder to read. An input grid alone also does not determine the task, as Figure~\ref{fig:taskid} shows with pairs of training tasks whose inputs are drawn from overlapping distributions. The same grid is then a valid input to both tasks, and only the demonstrations or the id say which rule applies. A learner may key its memory on the id, as the floors, the program library and the harnesses given expert guidance do (Section~\ref{sec:baselines}). It sees the id under an alias drawn for each run. The real id would be a lookup key: on the ARC-AGI-1 track, a learner with no model that runs the Re-ARC solver named by the id, and finds each held-out input among the public evaluation files, solves every instance of the stream and the held-out tasks at no cost, and uses none of the budget without learning anything. The alias makes such a lookup cost at least one demonstration per task, since a learner can still find the task by running every solver on a pair it has bought, and so what a learner may bring to the stream is stated with the code, and a result is verified by rerunning it from its code. A task-free variant would be a different benchmark, in which the learner must first buy enough demonstrations to identify the task, and we leave it to future work.

\paragraph{Observation horizon.} As the generators produce fresh instances without end, a configuration's stream length is only the window over which we observe a learner. It places no limit on how much the learner may learn. Within a run, a checkpoint is a prefix, and it never changes when the run is carried further. Every instance is charged, so the budget used against instances is the experience curve and the budget used at a horizon is its mean (Equation~\ref{eq:score}). A system that becomes competent after fewer instances then uses less of the budget at every checkpoint, which is the kind of efficiency that \citet{chollet2019measure} asked for, and the benchmark sets no cap on data. While a learner cannot choose how many instances arrive, it can choose how many demonstrations to buy, which the score charges, and how much of its experience to train on, which we count as compute and log without scoring.

\begin{table}[t]
\centering
\small
\setlength{\tabcolsep}{4pt}
\begin{tabular}{lrrrrrrr}
\toprule
track & tasks & stream & held-out & $n_c$ & $b$ & $m$ & $a$ \\
\midrule
ARC-AGI-1 & 25  & 200       & 100 & 4  & 3      & 30      & 2   \\
\bottomrule
\end{tabular}
\caption{The ARC-AGI-1 track's configuration. Each track is its configuration, in the individual regime, at seed 0. The stream length is per user, and the held-out tasks, those of the held-out test, are the same for every user. $n_c$, $b$, $m$ and $a$ are the stream's parameters (Appendix~\ref{app:schedules}), in instances, with $\sigma = 1$ throughout. The number of users $n_u \in \{1, 2, 4\}$ is a setting of the run, not of the configuration. At four users, the stream has 200 rounds over 100 distinct tasks, with eight instances per task and four graded instances per round. Every configuration has an attempt cap of 8 and a demonstration cap of 8, one pair per request, with $c_d = 1$.}
\label{tab:configs}
\end{table}

\paragraph{Intervals and chance.} Tables report pointwise 95\% intervals from 2{,}000 paired task resamples, after averaging each instance's spend over its plays; $L$ is resampled over stream tasks. The two-test planes instead show approximate simultaneous 95\% intervals across the displayed learners, separately for each model, reasoning effort, user count, track, test and measure. For $k$ learners, we take quantiles $0.025/k$ and $1-0.025/k$ from 200{,}000 paired task resamples (Bonferroni correction). A change is detected when this interval excludes zero for share saved, or one for the solve-rate ratio. These intervals condition on the learned state: repeated tests do not measure variation across independent learning runs. Task resampling includes independent play noise, but does not capture variation shared across a whole play. Two learners differ when the pointwise 95\% interval of their paired task-by-task difference excludes zero.

\paragraph{The exposure curve.} We write $e_i$ for the exposure index of instance $i$, the number of earlier rounds in which its task appeared (in the individual regime, a round is one instance), from 0 to 7, since the tracks serve every task eight times. Over the stream records $\mathcal{S}$, we define
\begin{equation}
U(k) = \operatorname{mean}\{\, u_i : e_i = k \,\},
\label{eq:metrics}
\end{equation}
and Figure~\ref{fig:notes-weights} plots it against the visit to the task, $e_i + 1$. $U(0)$ is the acquisition cost and $U(k)$ the exposure curve. A system that replays stored demonstrations to a stateless solver moves down this curve only by the demonstrations it no longer pays for, while a system that has learned a task falls below the curve of such a replay. The curve also has a solve-rate twin, which replaces $u_i$ with $s_i$. The standard measures of average accuracy, backward transfer and forgetting \citep{lopezpaz2017gradient} assume task boundaries and a fixed test set per task, neither of which a recurring stream scored by attempts has.

\paragraph{Why a simpler reading fails.} The drop from a system's first exposures to its later ones rewards a poor start, and a stream's returning instances need not be as hard as its first ones. On the ARC-AGI-1 track at one user, the stateless floor, which cannot learn, uses more of the budget on returns than on first exposures with every model, from 3 points on Qwen3.5-9B to 16 on GPT-6 Luna at medium effort. The share learned (Equation~\ref{eq:learned}) instead pairs a learner with its own reset on the same instances. Keeping the demonstrations a task was shown is a memory like any other, and the demo-cache floor, read against the stateless floor, scores above zero. The bare model is no reference. A learner read against its own reset could score by playing worse whenever its memory is empty. On Qwen3.5-9B every reset in prime-agent does play both tests below the bare model, since the harness gives up most of their instances after replies with no readable action (Appendix~\ref{app:harness}), so Table~\ref{tab:tests-absolute} marks each such reset and gives every learner's budget beside it, and a verified result is also played bare and flagged when the reset plays below the bare model (Appendix~\ref{app:heldout}). A learner at more than one user is read against the reset at the same number of users, which met the same instances (Appendix~\ref{app:trained}).

\paragraph{Generating data at will.} \citet{chollet2019measure} also wrote that an ideal benchmark should avoid tasks for which new data can be generated at will, because performance can then be bought with data. We generate data at will deliberately, for two reasons. In the stream, every demonstration and every wrong attempt is charged, and so the measure is the feedback consumed, as he proposed, which leaves no room for the unpaid practice on the test distribution that concerned him. In the held-out test there are no generators at all, which rules out buying final skill with data where we measure transfer. The concern does survive in a weaker form, since Re-ARC covers the public training split, which any developer knows.

\paragraph{Compute.} Each record carries whatever the learner reports, as a per-instance difference in cumulative counters, including tokens, wall-clock seconds, bytes of persistent state and dollars, and, for a learner that trains on its traces, the number of instances it has trained on. The baselines report these through the accounting layer so that they do not count for themselves. We log compute without scoring it, because we do not know the exchange rate between a weight update and a hundred thousand prompt tokens, and any fixed rate would decide the comparison in advance.

\section{Schedule generator details}
\label{app:schedules}

We generate every schedule deterministically from the configuration name and the seed, and a track is the seed-0 stream of its configuration (Section~\ref{sec:stream}), whose process this appendix gives. The same process, set otherwise, also serves the stream's visits in two other orders, which only a side experiment plays (below); they are not part of the benchmark. Since we derive instance seeds by hashing the schedule seed, the phase and the slot index, slot $t$ carries the same instance seed in every order and always yields the same instance. We also ensure that the test input of a slot never appears among the demonstrations served for it.

\paragraph{The stream process.}
Algorithm~\ref{alg:stream} generates every stream, keeping up to $n_c$ tasks active and, at each step, sampling one active task uniformly and emitting a slot for it. A task becomes active for a spell of $\beta$ instances, and when the spell is spent it rests for $\rho$ instances before it is ready to be re-activated, unless $\rho = \infty$, in which case it is finished. Never-seen tasks arrive every $\alpha$ instances and take priority over waiting tasks for the next free place. When no task is ready and no arrival is due, we wake the soonest-waking dormant task early so that the stream never idles, and we pull unseen tasks ahead of schedule only when no other task exists. Every task has a supply of $s$ instances, and a spell never exceeds what the task has left. A task whose supply is spent is finished and does not rest. Every spell, rest and inter-arrival time is a fresh draw from the distribution that the order sets for it, and all draws come from one generator seeded by the schedule seed and the order's name.

The stream and the side experiment's two orders are settings of $(n_c, \beta, \rho, \alpha)$, the parameters of Equation~\ref{eq:stream}:
\begin{itemize}
  \item \textbf{one task at a time} (sequential): $n_c = 1$, $\beta$ is the constant $L / n$ (the stream length over the number of tasks), $\rho = \infty$ and $\alpha$ is the constant 0, so the process reduces to the tasks in a shuffled order, each for $L/n$ consecutive instances.
  \item \textbf{the stream} (workstreams): $n_c$ from Table~\ref{tab:configs}, $\beta = \max(1, \lfloor \mathrm{Exp}(b) + \tfrac12 \rfloor)$, $\rho = \max(1, \lfloor \mathrm{LogNormal}(\ln m, \sigma) + \tfrac12 \rfloor)$ and $\alpha \sim \mathrm{Exp}(a)$.
  \item \textbf{fully shuffled} (random): $n_c = n$, $\beta = \infty$ and $\alpha = 0$ ($\rho$ is never drawn), so every task is introduced at $t = 0$ and stays until its supply is spent, and each step is a uniform draw over the tasks with instances left.
\end{itemize}
Figure~\ref{fig:modes} shows the stream of the ARC-AGI-1 track, and the code and the records name its order \texttt{workstreams}.

\begin{algorithm}[h]
\caption{The stream. Parameters: concurrency $n_c$; spell $\beta$, rest $\rho$ (possibly $\infty$) and inter-arrival time $\alpha$, each a fresh draw at every use; supply $s$ per task; length $L$.}
\label{alg:stream}
\begin{algorithmic}[1]
\STATE $\mathcal{Q} \gets$ shuffled task list (never seen); $s_\tau \gets s$ for every task $\tau$; $\mathcal{A} \gets \varnothing$ (active, with remaining spell); $\mathcal{D} \gets \varnothing$ (dormant, with wake step); $\mathcal{W} \gets \varnothing$ (ready); $t_{\mathrm{next}} \gets 0$
\FOR{$t = 0, \dots, L-1$}
  \STATE move every task in $\mathcal{D}$ with wake step $\le t$ to $\mathcal{W}$
  \WHILE{$|\mathcal{A}| < n_c$}
    \IF{$\mathcal{Q} \ne \varnothing$ and $t \ge t_{\mathrm{next}}$} \STATE $\tau \gets \operatorname{pop}(\mathcal{Q})$; \ $t_{\mathrm{next}} \gets t + \alpha$
    \ELSIF{$\mathcal{W} \ne \varnothing$} \STATE $\tau \gets$ uniform draw from $\mathcal{W}$, removed
    \ELSIF{$\mathcal{D} \ne \varnothing$} \STATE $\tau \gets$ the task in $\mathcal{D}$ with the earliest wake step, removed
    \ELSIF{$\mathcal{Q} \ne \varnothing$} \STATE $\tau \gets \operatorname{pop}(\mathcal{Q})$
    \ELSE \STATE \textbf{break}
    \ENDIF
    \STATE $\mathcal{A}[\tau] \gets \min(\beta, s_\tau)$
  \ENDWHILE
  \STATE $\tau \gets$ uniform draw from $\mathcal{A}$; \ emit slot $(\tau, \text{seed}(t))$; \ $\mathcal{A}[\tau] \gets \mathcal{A}[\tau] - 1$; \ $s_\tau \gets s_\tau - 1$
  \IF{$\mathcal{A}[\tau] = 0$} \STATE remove $\tau$ from $\mathcal{A}$; \ \textbf{if} $\rho < \infty$ and $s_\tau > 0$ \textbf{then} $\mathcal{D}[\tau] \gets t + 1 + \rho$
  \ENDIF
\ENDFOR
\end{algorithmic}
\end{algorithm}

On the ARC-AGI-1 track, a learner has met 14 tasks by step 50 and all 25 by step 100, and the gap before a re-exposure has a median of 5 instances, a 90th percentile of 49 and a maximum of 141.

\paragraph{Users and rounds.}
A run sets a number of users $n_u \in \{1, 2, 4\}$. With $n_u = 1$, the schedule is one stream and every slot forms a round. With $n_u > 1$, every user has a separate schedule, drawn by the same process with a separate seed over tasks of their own and the held-out tasks drawn at the configuration's seed. User $k$ plays block $k$ of 25 from a nested order of 100 tasks: the track's 25 sorted by id, then 75 more from a seeded draw of the rest, sorted. User 0, which keeps the configuration's seed and its tasks, meets exactly the individual stream at that seed. Round $r$ serves slot $r$ of every user at once, and the learner acts on every user's instance with the state that the round began with. That state is one for all users, a harness's store included: every instance of a round starts from it, and what each instance writes is merged into it in user order before the next round begins, by a rule for files and databases that two users changed alike which the benchmark's code gives. Each instance runs from its first action to its verdict at its user's pace, and the round ends when every user's instance has. We compute the exposure index and gap of a slot over rounds and over the whole team. Every record also carries the count for its user. At $n_u = 1$, both are the definitions of Appendix~\ref{app:bench}, unchanged.

\paragraph{The two tests and their references.}
After the stream, each test is one round whose every slot is a user of its own: the seen test serves four fresh instances of each stream task at one user, and at two and four users the same 100 spread evenly over every user's tasks (two of each of 50 at two users, one of each of 100 at four), at seeds the stream did not use, with instance $i$ a task of user $i \bmod n_u$. The held-out test serves one instance of each held-out task, the instances user 0 would meet, at any number of users. Every user starts from the learner's state at the end of the stream, so the round serves 100 users at once and nothing reaches one test instance from another. A learner that cannot serve them all at once plays the test narrower, with the reset flag on every slot restoring its end state before the instance. The learner reset before each instance, its reference, plays the same slots from an empty state, and so does the bare model (Appendix~\ref{app:heldout}).

\paragraph{A side experiment on the order.}\label{app:orders}
We were curious whether the order in which the stream serves its visits changes what the harnesses do, so, outside the benchmark, GPT-6 Luna's systems also played the same visits one task at a time and fully shuffled (above). The trends were the same in every order and at every effort: the demo-cache floor uses less of the budget than the stateless floor, each persistent harness uses about what it uses wiped, and the program library uses the least, 6.2 to 26.1\%. Since the three orders serve the same visits, a comparison in these appendices that needs more slots pools them.

\paragraph{A sweep of the stream's settings.}\label{app:sweep}
To see how far the measures depend on the stream's settings, GLM-5.3 Flash's four methods (G, GE, GC and GP) and their resets also played twelve other settings at one user, outside the benchmark. Two play the default stream again from scratch (below), and two play the side experiment's orders. Four change one parameter of Equation~\ref{eq:stream} from its value in Table~\ref{tab:configs}: $n_c$ from 4 to 12, $a$ from 2 to 8, and $m$ from 30 to 5 or to 90. The last four keep the default stream and set the demonstration cap or the attempt cap to 4 or to 16 in place of 8. Each method played its stream once and then each test four times. Another stream serves other instances, so $L$ reads a method against its reset on the same stream. The tests serve the same instances after any stream, so every uncapped stream's tests are read against the reset's four plays beside the default stream's first run. A cap changes what the reset spends, so a capped setting is read against its own reset under the same cap, which played each test four times.

\begin{figure}[htbp]
\centering
\includegraphics[width=\linewidth]{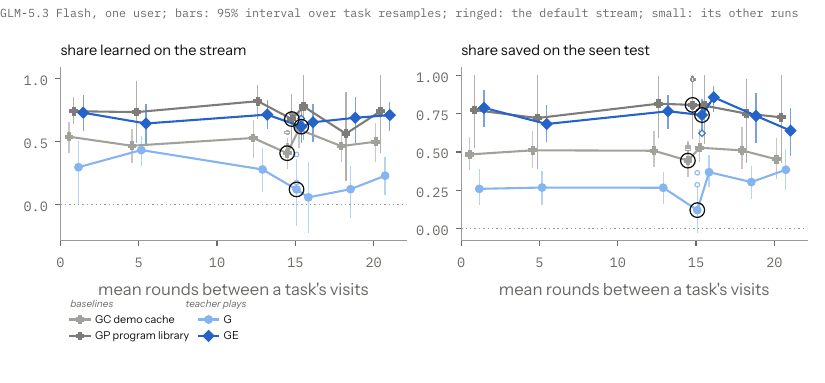}
\caption{The stream sweep on GLM-5.3 Flash at one user: the share learned on the stream (left) and the share saved on the seen test (right) of G, GE, GC and GP, each against its reset, by the mean gap between a task's visits on its stream, from one task at a time to fully shuffled. The default stream's first run is ringed, and its two other runs are the small markers beside it.}
\label{fig:stream-sweep}
\end{figure}

The mean gap between a task's visits ran from 1.0 instances with one task at a time to 20.6 fully shuffled, against 14.9 by default. The rest length hardly moved it, since a task returns when one of the stream's four places frees up, not when its rest ends. Figure~\ref{fig:stream-sweep} draws $L$ and $S_{\mathrm{seen}}$ against the mean gap, and Table~\ref{tab:sweep} gives both tests on every setting. On the seen test all four methods saved on all nine uncapped streams, in the same order on each: GP and GE saved 0.62 to 0.98 of their resets' spend, GC 0.44 to 0.54 and G 0.12 to 0.38. G's interval included zero only on the default stream's first run. Under each of the four caps the methods saved in that order again, with every interval above zero, and each of a method's four plays cost less than every play of its reset (\texttt{scripts/read\_caps\_plays.py}). On the held-out test five of the 52 intervals exclude zero, in both directions, from GE under a demonstration cap of 4, which spent 0.10 [0.02, 0.19] more than its reset, to G under an attempt cap of 4, which saved 0.18 [0.09, 0.27]. Each of the five is one stream against one reset, and every uncapped stream is read against the same four plays of its method's reset, so we draw no conclusion from any one of them (\texttt{scripts/read\_sweep.py}).

\begin{table}[htbp]
\centering
\footnotesize\setlength{\tabcolsep}{3pt}
\begin{tabular}{lrr@{\hspace{0.4em}}lr@{\hspace{0.4em}}lr@{\hspace{0.4em}}lr@{\hspace{0.4em}}l}
\toprule
setting & gap & \multicolumn{2}{c}{G} & \multicolumn{2}{c}{GE} & \multicolumn{2}{c}{GP} & \multicolumn{2}{c}{GC} \\
\midrule
\multicolumn{10}{l}{\emph{share saved on the seen test, $S_{\mathrm{seen}}$}} \\
default & 14.9 & +0.12 & {\scriptsize [-0.03, 0.27]} & +0.74 & {\scriptsize [0.64, 0.84]} & \textbf{+0.81} & {\scriptsize [0.59, 1.00]} & +0.44 & {\scriptsize [0.34, 0.57]} \\
default, second run & 14.9 & +0.29 & {\scriptsize [0.15, 0.41]} & +0.75 & {\scriptsize [0.65, 0.83]} & \textbf{+0.98} & {\scriptsize [0.92, 1.00]} & +0.54 & {\scriptsize [0.42, 0.68]} \\
default, third run & 14.9 & +0.37 & {\scriptsize [0.25, 0.48]} & +0.62 & {\scriptsize [0.50, 0.73]} & \textbf{+0.80} & {\scriptsize [0.58, 1.00]} & +0.52 & {\scriptsize [0.41, 0.64]} \\
one task at a time & 1.0 & +0.26 & {\scriptsize [0.16, 0.38]} & \textbf{+0.79} & {\scriptsize [0.67, 0.90]} & +0.77 & {\scriptsize [0.53, 1.00]} & +0.48 & {\scriptsize [0.39, 0.60]} \\
$a = 8$ & 5.0 & +0.27 & {\scriptsize [0.16, 0.37]} & +0.68 & {\scriptsize [0.57, 0.80]} & \textbf{+0.72} & {\scriptsize [0.39, 1.00]} & +0.51 & {\scriptsize [0.40, 0.64]} \\
$m = 90$ & 12.8 & +0.27 & {\scriptsize [0.16, 0.37]} & +0.77 & {\scriptsize [0.65, 0.87]} & \textbf{+0.82} & {\scriptsize [0.60, 0.99]} & +0.51 & {\scriptsize [0.41, 0.63]} \\
$m = 5$ & 15.7 & +0.37 & {\scriptsize [0.28, 0.48]} & \textbf{+0.86} & {\scriptsize [0.76, 0.93]} & +0.80 & {\scriptsize [0.56, 1.00]} & +0.53 & {\scriptsize [0.44, 0.65]} \\
$n_c = 12$ & 18.4 & +0.30 & {\scriptsize [0.20, 0.40]} & +0.74 & {\scriptsize [0.56, 0.88]} & \textbf{+0.75} & {\scriptsize [0.43, 0.97]} & +0.51 & {\scriptsize [0.39, 0.66]} \\
fully shuffled & 20.6 & +0.38 & {\scriptsize [0.26, 0.52]} & +0.64 & {\scriptsize [0.48, 0.78]} & \textbf{+0.73} & {\scriptsize [0.43, 1.00]} & +0.45 & {\scriptsize [0.33, 0.59]} \\
demonstration cap 4 & 14.9 & +0.20 & {\scriptsize [0.06, 0.32]} & +0.78 & {\scriptsize [0.70, 0.87]} & \textbf{+0.80} & {\scriptsize [0.61, 0.97]} & +0.47 & {\scriptsize [0.38, 0.60]} \\
demonstration cap 16 & 14.9 & +0.26 & {\scriptsize [0.09, 0.44]} & +0.72 & {\scriptsize [0.58, 0.84]} & \textbf{+0.73} & {\scriptsize [0.50, 1.00]} & +0.46 & {\scriptsize [0.37, 0.58]} \\
attempt cap 4 & 14.9 & +0.20 & {\scriptsize [0.12, 0.29]} & +0.51 & {\scriptsize [0.22, 0.76]} & \textbf{+0.80} & {\scriptsize [0.57, 0.97]} & +0.51 & {\scriptsize [0.39, 0.63]} \\
attempt cap 16 & 14.9 & +0.24 & {\scriptsize [0.09, 0.36]} & \textbf{+0.72} & {\scriptsize [0.61, 0.82]} & +0.71 & {\scriptsize [0.36, 0.99]} & +0.53 & {\scriptsize [0.41, 0.66]} \\
\addlinespace[2pt]
\emph{lowest to highest} & & \multicolumn{2}{c}{+0.12 to +0.38} & \multicolumn{2}{c}{+0.51 to +0.86} & \multicolumn{2}{c}{+0.71 to +0.98} & \multicolumn{2}{c}{+0.44 to +0.54} \\
\midrule
\multicolumn{10}{l}{\emph{share saved on the held-out test, $S_{\mathrm{held}}$}} \\
default & 14.9 & 0.00 & {\scriptsize [-0.11, 0.10]} & +0.07 & {\scriptsize [-0.02, 0.16]} & +0.05 & {\scriptsize [-0.06, 0.14]} & -0.02 & {\scriptsize [-0.11, 0.06]} \\
default, second run & 14.9 & +0.05 & {\scriptsize [-0.06, 0.16]} & \textbf{+0.14} & {\scriptsize [0.04, 0.23]} & -0.07 & {\scriptsize [-0.21, 0.05]} & +0.02 & {\scriptsize [-0.05, 0.08]} \\
default, third run & 14.9 & +0.09 & {\scriptsize [-0.01, 0.17]} & +0.09 & {\scriptsize [-0.02, 0.19]} & -0.07 & {\scriptsize [-0.19, 0.04]} & +0.06 & {\scriptsize [-0.02, 0.12]} \\
one task at a time & 1.0 & +0.01 & {\scriptsize [-0.10, 0.10]} & +0.06 & {\scriptsize [-0.06, 0.16]} & +0.01 & {\scriptsize [-0.11, 0.12]} & +0.02 & {\scriptsize [-0.06, 0.10]} \\
$a = 8$ & 5.0 & +0.03 & {\scriptsize [-0.08, 0.14]} & +0.07 & {\scriptsize [-0.05, 0.17]} & -0.04 & {\scriptsize [-0.17, 0.08]} & \textbf{+0.07} & {\scriptsize [0.01, 0.15]} \\
$m = 90$ & 12.8 & +0.04 & {\scriptsize [-0.07, 0.14]} & +0.11 & {\scriptsize [-0.02, 0.21]} & -0.02 & {\scriptsize [-0.13, 0.08]} & +0.03 & {\scriptsize [-0.07, 0.12]} \\
$m = 5$ & 15.7 & +0.07 & {\scriptsize [-0.03, 0.17]} & +0.04 & {\scriptsize [-0.08, 0.15]} & +0.02 & {\scriptsize [-0.11, 0.14]} & +0.03 & {\scriptsize [-0.05, 0.10]} \\
$n_c = 12$ & 18.4 & +0.06 & {\scriptsize [-0.05, 0.16]} & +0.09 & {\scriptsize [-0.02, 0.19]} & +0.01 & {\scriptsize [-0.11, 0.12]} & +0.01 & {\scriptsize [-0.09, 0.10]} \\
fully shuffled & 20.6 & +0.05 & {\scriptsize [-0.06, 0.14]} & \textbf{+0.11} & {\scriptsize [0.01, 0.19]} & -0.03 & {\scriptsize [-0.17, 0.09]} & 0.00 & {\scriptsize [-0.09, 0.08]} \\
demonstration cap 4 & 14.9 & +0.11 & {\scriptsize [0.00, 0.20]} & -0.10 & {\scriptsize [-0.19, -0.02]} & -0.05 & {\scriptsize [-0.16, 0.04]} & -0.01 & {\scriptsize [-0.08, 0.05]} \\
demonstration cap 16 & 14.9 & +0.06 & {\scriptsize [-0.13, 0.21]} & -0.09 & {\scriptsize [-0.24, 0.05]} & +0.08 & {\scriptsize [-0.05, 0.21]} & +0.07 & {\scriptsize [-0.03, 0.17]} \\
attempt cap 4 & 14.9 & \textbf{+0.18} & {\scriptsize [0.09, 0.27]} & +0.04 & {\scriptsize [-0.06, 0.12]} & +0.01 & {\scriptsize [-0.08, 0.10]} & -0.03 & {\scriptsize [-0.12, 0.05]} \\
attempt cap 16 & 14.9 & -0.03 & {\scriptsize [-0.19, 0.12]} & -0.06 & {\scriptsize [-0.18, 0.06]} & -0.04 & {\scriptsize [-0.20, 0.10]} & 0.00 & {\scriptsize [-0.12, 0.10]} \\
\addlinespace[2pt]
\emph{lowest to highest} & & \multicolumn{2}{c}{-0.03 to +0.18} & \multicolumn{2}{c}{-0.10 to +0.14} & \multicolumn{2}{c}{-0.07 to +0.08} & \multicolumn{2}{c}{-0.03 to +0.07} \\
\bottomrule
\end{tabular}

\caption{The stream sweep on GLM-5.3 Flash at one user: the share of its reset's spend that each method saved on each test (Equation~\ref{eq:transfer}, higher is better) on each setting, four plays against four, with its 95\% interval over task resamples in small type. The gap is the mean number of instances between a task's visits on the setting's stream. An uncapped stream is read against the reset's plays beside the default stream's first run, and a capped setting against its own reset under the same cap. Bold marks each row's best among the values whose 95\% interval lies above zero. The last line under each test gives each method's lowest and highest value over the thirteen settings.}
\label{tab:sweep}
\end{table}

\paragraph{Three runs from scratch.}\label{app:runs}
With its two replays, the default stream was played three times from scratch by each of the four methods and by its reset. Table~\ref{tab:sweep-runs} reads the three runs together, with each side's spend on an instance averaged over its plays. The model's provider changed how it serves GLM-5.3 Flash after the first run and before the later runs' tests: on the same endpoint and model snapshot, the plays we sampled decoded at 34 to 79 tokens per second before the change and at 83 to 118 after it. A reset played before the change is therefore not known to be the system that the later runs played on, so the resets of G and GE played both tests four more times beside the later runs, and each run is read against the reset of its own week. The baselines' resets, whose streams cost about the same in both weeks, were not played again. Over the three runs G saved on both tests, with $S_{\mathrm{seen}} = 0.25$ [0.17, 0.33], half of what GC saved, and $S_{\mathrm{held}} = 0.10$ [0.03, 0.17] over the runs' twelve plays, or 0.08 [0.03, 0.12] with the 42 further plays read below, which is the table's value. The choice of reset moves both harness methods on the held-out test. With the first run's reset standing for all three runs, GE reads 0.10 [0.01, 0.18] where the table has 0.01, and G reads 0.05 [$-0.03$, 0.12]. Against all eight plays of its reset pooled G reads 0.09 [0.02, 0.15], and the two weeks of that reset differ by no more than its plays do, 30 of the 70 ways to split the eight into two fours giving a gap as large. The intervals resample tasks, so they also leave out how much one whole play of a test differs from the next. Read as whole plays, G's twelve held-out plays saved 0.09 against its reset's eight, and 7,008 of the 125,970 ways to choose eight of the twenty plays as the reset's give a gap as large ($p = 0.06$), where 12 do on the seen test (\texttt{scripts/read\_drift\_weeks.py}). Since whole plays left the held-out saving just short, G's held-out test was played 42 more times in a week of its own, seven times from each run's end state beside 21 plays of its reset, and read as fixed before any of them had ended (\texttt{scripts/read\_more\_held\_out.py}). The new plays alone give $S_{\mathrm{held}} = 0.06$ [0.01, 0.12] and, as whole plays, 0.06 saved with $p = 0.009$ over a million random splits. All 62 plays give 0.07 [0.03, 0.12] and $p = 0.0015$, and read as the table reads them, each week against its own reset, 0.08 [0.03, 0.12]. So G's saving on unseen tasks is small, and measurable both ways.

\begin{table}[htbp]
\centering
\footnotesize\setlength{\tabcolsep}{3pt}
\begin{tabular}{lr@{\hspace{0.4em}}lr@{\hspace{0.4em}}lr@{\hspace{0.4em}}ll}
\toprule
method & \multicolumn{2}{c}{$L$} & \multicolumn{2}{c}{$S_{\mathrm{seen}}$} & \multicolumn{2}{c}{$S_{\mathrm{held}}$} & resets on the tests \\
\midrule
G & +0.23 & {\scriptsize [0.10, 0.35]} & +0.25 & {\scriptsize [0.17, 0.33]} & \textbf{+0.08} & {\scriptsize [0.03, 0.12]} & of each run's week \\
GE & +0.63 & {\scriptsize [0.58, 0.69]} & +0.68 & {\scriptsize [0.60, 0.76]} & +0.01 & {\scriptsize [-0.04, 0.07]} & of each run's week \\
\midrule
GP & \textbf{+0.69} & {\scriptsize [0.47, 0.90]} & \textbf{+0.86} & {\scriptsize [0.72, 0.99]} & -0.03 & {\scriptsize [-0.13, 0.06]} & of the first run's week \\
GC & +0.48 & {\scriptsize [0.39, 0.58]} & +0.50 & {\scriptsize [0.40, 0.62]} & +0.02 & {\scriptsize [-0.04, 0.07]} & of the first run's week \\
\bottomrule
\end{tabular}

\caption{GLM-5.3 Flash's methods over three runs of the default stream from scratch, read as in Table~\ref{tab:tests}, with twelve plays of each test for each method, and 21 more of G's held-out test beside 21 more of its reset. The last column gives the resets that the tests are read against: for G and GE the reset played in each run's own week, eight plays in all, and for GP and GC the first run's four plays. Bold marks each column's best among the values whose 95\% interval lies above zero.}
\label{tab:sweep-runs}
\end{table}

\section{Prompts}
\label{app:prompts}

The texts below describe the prompts used for the reported experiments, preserved at the repository versions listed below. Appendix~\ref{app:harness} describes how we run each harness, and Appendix~\ref{sec:models} gives the model presets. The rules below state the charge of Equation~\ref{eq:score}, in place of the earlier charge they were first written for (Appendix~\ref{sec:charge}).

\paragraph{Expert guidance in the current code.}
The code now uses \href{https://github.com/unifyai/continual-arc/blob/main/baselines/src/continual_arc_baselines/systems/guidance.py}{one expert-guidance prompt}{} for every harness and DAgger teacher, a deduplicated superset of the playing rules and the program-library procedure. The historical prompts are preserved in commits \href{https://github.com/unifyai/continual-arc/tree/b2fd5a94572fa83cde9ecd20b38d41d1548d64b9}{\texttt{b2fd5a9}} (GPT-6 Luna) and \href{https://github.com/unifyai/continual-arc/tree/9616cbd2c1d5741142ca7f2590ecb3535028daf9}{\texttt{9616cbd}} (the open models).  QDE's teacher read the playing rules alone, QE and GE read both texts, and GPT-6 Luna's guided harnesses read an earlier version of the library procedure. The results reported here use those earlier prompts, and the consolidated prompt has not been evaluated.

Every learner on the first two tracks is shown the grid protocol's rules below once per session, and each instance opens with a message like the one after them. Both are reproduced from the benchmark's repository, the same at tag \texttt{v0.3.0}, commit \texttt{b2fd5a9}, which GPT-6 Luna's methods read, and at tag \texttt{v0.5.0}, commit \texttt{fa4b33e}, which the rest read, with lines longer than the page wrapped at spaces; placeholders in braces are filled at run time with the values named in Section~\ref{sec:benchmark}. Every other text a learner is shown, from the replies to its actions to what each method adds to the rules and the third track's protocol, is in the code\ released with the paper, in the files of Table~\ref{tab:prompts}.

\paragraph{Grid protocol: rules shown once per session (demo cost 1)}
\begin{scriptsize}\begin{verbatim}
You are being evaluated on Continual-ARC, a lifelong stream of ARC-style grid puzzles.
Each puzzle instance has a task id and a test input grid. Tasks recur across the
stream with fresh inputs. Each instance is a separate conversation: the next one
starts without this one's messages.

On every message you must choose exactly one action and put it as a JSON object
on the last line of your reply:
  {"action": "submit", "grid": [[0, 1, 2], [3, 4, 5]]}   submit an output grid
  {"action": "request_demos"}                            ask for one more demonstration pair
    for this task
  {"action": "finish"}                                   end the instance, once it is over
Grids are lists of rows; cells are integers 0-9; dimensions are 1-30.

Each demo request returns one more input-output pair of the task, up to the demo cap shown
with the input.

Scoring: an instance you solve costs 1 for each wrong submission and 1 for
each demo request. An instance is over when a submission is correct or when the attempt
cap is reached, and one that ends unsolved costs the whole budget, {max_cost}, however
many submissions and demo requests it used. Feedback on a submission is only
correct/incorrect. There is no way to abandon an instance, and finish is refused until
the instance is over. Each instance is scored on its own, and lower cost is better.
\end{verbatim}\end{scriptsize}

\paragraph{Grid protocol: an instance's first message, after the rules (an example)}
\begin{scriptsize}\begin{verbatim}
New instance. Task id: task-3f2a9c1e
Test input (2x2):
2 0
0 2
Wrong attempts used: 0/8. Demo requests used: 0/8.
Reply with exactly one action JSON on the last line.
\end{verbatim}\end{scriptsize}

\begin{table}[htbp]
\centering
\footnotesize
\begin{tabular}{@{}lp{0.6\linewidth}@{}}
\toprule
file & the texts it holds \\
\midrule
\texttt{systems/protocol.py} & the grid protocol's other messages: the replies to an answer or a request, the verdicts, the re-prompts and the budget notice; the ask after every verdict of methods QA, GA and \mbox{H-A}; the expert guidance of methods QE, GE, \mbox{H-E}, \mbox{H-E\textsubscript{0}} and QETE and its asks \\
\texttt{systems/prime\_agent.py} & prime-agent's one addition to the rules, its fact about its store \\
\texttt{systems/bare.py} & the demo-cache floor's cached pairs, and the program library's addition to the rules, its kept program, its checks and its re-prompt \\
\texttt{systems/arc3\_protocol.py} & the third track's protocol: its rules, messages, feedback and re-prompt \\
\texttt{references.py} & c1, the page of playing rules that DAgger's teacher reads \\
\texttt{oracle.py} & the rules and the note with which DAgger's teacher labels the student's calls \\
\bottomrule
\end{tabular}
\caption{Where the benchmark's code keeps every other text a learner is shown: the first four files under \texttt{baselines/src/continual\_arc\_baselines/} and the rest under \texttt{training/src/continual\_arc\_training/}, at tag \texttt{v0.3.0} for GPT-6 Luna's methods and \texttt{v0.5.0} for the rest, which alone has DAgger's teacher and the third track.}
\label{tab:prompts}
\end{table}

\section{Harness implementation details}
\label{app:harness}

This appendix gives the details of how we run the three harnesses. Each harness's command line, configuration, budgets and software versions are in the adapters' code in the benchmark's repository (\texttt{baselines/}), at its public tag. The open models' methods of Section~\ref{sec:results} ran prime-agent 0.9.6, whose digest gives the agent the same guidance on what to keep as 0.10.0's (Appendix~\ref{app:persist}).

\subsection{A conversation per instance}
\label{app:per-instance}

Each instance is a new piece of work. At the start of every instance, we open a fresh conversation with the harness and send the protocol rules again, and at the end of the instance, once the harness's end-of-work steps have run (below), we close the conversation. Nothing carries from one instance to the next except what the harness persisted itself, which is the global harness store, skills and workspace files for prime-agent, the memory and skill files for hermes-agent, and the memory notes, skills and dream diary for openclaw. This protocol follows the intended use of these systems, one session per job with anything worth keeping written to files, and under it a method measures only what the harness chose to keep.

We also ran the alternative, one conversation per cell for the whole stream, and found that it measures something else. The transcript then carried everything a harness knew about a recurring task, so its re-exposure gains were in-context learning that every compaction and restart wiped, and prime-agent's refinements lived in a session-local store that a restart discarded. Two harnesses also got worse as their context grew over the stream, which reflects compaction more than learning. A short run of the per-instance protocol then confirmed, from each harness's files, that a session opens with only our observation and that the next session loads what was persisted.

\subsection{How each harness is driven}

\paragraph{The end of an instance.} Every instance ends with one more turn in its conversation, carrying the verdict on the final submission (``Your submission was CORRECT. Instance solved with 1 wrong attempt(s).'', or the cap-reached line). The harness may then do what it wants, including storing what it learned with its own tools, and it ends the instance itself with the finish action of the rules, which is refused while the instance can still be played, so that no message of ours tells it what to do after the verdict. Any other reply after the verdict is answered with the fact that the instance is over, and after two such re-prompts, or when the harness spends its turn's budget, the instance ends there. A harness method and its wiped twin end their instances alike, and the rules state the protocol and give no advice (Appendix~\ref{app:prompts}), apart from the one fact about prime-agent's store that its agent cannot know (below). Our first version of the protocol had no such turn. prime-agent's refine and the post-turn reviews of hermes-agent and openclaw stored trajectories without their outcome, and we added the turn to fix this. Without it, a harness that stores only when its agent calls a storage tool could not store knowing that the answer was right.

\paragraph{Isolation.} Each harness runs as one long-lived process per cell, with a fresh session per instance, in a bubblewrap sandbox whose whole filesystem is read-only, with the benchmark source, its data, every virtual environment and the run tree of every cell masked by empty tmpfs mounts. Only the harness's private state directory, a private HOME and a private TMPDIR are writable. The sandbox has a private PID namespace and a private network namespace, whose only route out is a UNIX-socket bridge to a local reverse proxy that holds the real credential, journals every request and reply with its tokens and charge, and forwards the body unchanged apart from the run's reasoning effort and session id. Every measure of cost in this paper uses the proxy's counters. A harness sees each task as \texttt{task-<8 hex>}, derived from the real id with a salt drawn once per run. The proxy forwarded a request to any path, so a harness's own code could reach any of the provider's endpoints on the run's key. In the open-model methods, where it also set each chat completion's model to the method's, the agents' own code sent it 5{,}302 other requests, mostly to paths they had invented, and nearly all were refused. Apart from reads of the model list, 20 succeeded, in five games. Eight of them, in a stream game of method Q\textsubscript{0} at two users, went to OpenRouter's responses endpoint and were served by the other model they named; the agent asked that model for the task's demonstrations, which it could not know, and then solved the game from the demonstrations the game gave. The other four games were charged the whole budget. The released proxy forwards chat completions and reads of the model list alone, and sends every chat completion to the method's model.

\paragraph{Wiped methods.} Before every instance, we stop the harness and delete everything it wrote, from its home directory, memory and skill files to its sessions, workspace and temporary files. Only the adapter's provider configuration survives, and we then start the harness again. We also skip its idle gap, since nothing could survive it.

\paragraph{What each harness keeps by design.} prime-agent's automatic refinement runs at its default, every 25 assistant turns, and writes the session-local store, while the global store that later sessions load changes only when the agent calls the continual harness with \texttt{global\_=True} (Appendix~\ref{app:persist-prime}), and its rules add that entries created without \texttt{global\_=True} belong to the session and are lost with it. hermes-agent's post-turn background review of memory and skill files stays on, as shipped, and fires after any turn once ten tool iterations have accumulated, and its \texttt{MEMORY.md}, \texttt{USER.md} and skills index are injected into every new session's system prompt. openclaw keeps one session, reset in place at every instance, which drops the transcript and keeps the workspace, memory files and skills. We enable its memory plugin's \texttt{dreaming} and set its sweep to hourly, from a nightly default, to match the time scale of the stream, and leave memory search unset, so openclaw looks for OpenAI embeddings, which the sandbox cannot reach (Appendix~\ref{app:persist-openclaw}). After every instance we idle for 40 seconds, since openclaw's experience review, the step that turns a session into a skill proposal, starts only after 30 seconds of quiet with no run active.

\paragraph{prime-agent's Python skills.} Several of prime-agent's bundled skills are Python packages that its kernel imports, among them \texttt{refine}, which hands the conversation to prime-agent's host to review and update the harness store, and \texttt{edit}, \texttt{compact} and \texttt{goal}. prime-agent installs them only into a kernel environment it manages itself, and our adapter points it at an interpreter of its own instead (\texttt{PRIME\_AGENT\_KERNEL\_PYTHON}), so in every prime-agent run of the first track, the trained methods' included, they failed to import. Calling one raised an error, and releases 0.9.6 and 0.10.0, which those runs used, also told the agent so once its kernel started. The store itself stayed writable through the calls that prime-agent's system prompt documents beside \texttt{refine} (\texttt{rlm.harness.create\_memory} and the rest), and the agents stored through them: a call to \texttt{refine.run} failed in at most a tenth of any harness method's games, while asked to store (methods QA, GA and \mbox{H-A}, Appendix~\ref{app:ablations}), GLM-5.3 Flash wrote to its store directly in 79\% of its games, GPT-6 Luna in 52 to 83\% by effort and Qwen3.5-9B in 18\%. On the second and third tracks, Kimi K3.1 Pro called none of the skills in its streams. Played again with the skills installed, beside the same games without them, method G used 13.0 against 15.4\% of the budget over one pair of 200 games and 15.8 against 17.5 over a second, neither difference beyond chance by a sign test over the games ($p = 0.25$ and $0.11$), and method Q 89.2 against 85.4 over its 200 games ($p = 0.31$). The released adapter installs the skills, and stops a run in which they do not import.

\subsection{Interruptions, the server's window and give-ups}

\paragraph{Pauses and restarts.} Every cell stopped after its first 25 instances while we checked it, and 53 of the sequential and random cells were started again, some several times, after their runner's disk filled. On the runners' code, the first instance after a start did not always begin as every other did. Methods H\textsubscript{0}, H-E\textsubscript{0} and PL\textsubscript{0} skipped their wipe there, so 66 of their 15{,}692 instances began on what the instance before had left, 23 of them on the same task and 6 of those solved without a demonstration, all at the 26th instance of a sequential cell. openclaw's bridge takes up its one session at every start, so 75 instances of its methods H, \mbox{H-A} and \mbox{H-E}, after such a start or a restart between instances, went on in an earlier instance's conversation; 29 of them were on that instance's task, and 14 of those were solved without a demonstration, against a quarter of the other instances that followed one on their task. While the disk was full, openclaw gave no answer at 67 instances of its sequential and random cells, which were given up under the adapters' restart rule. The released adapters begin every instance alike.

\paragraph{The window on our own server.} The server that plays the trained methods and their references serves Qwen3.5-9B with a window of 131{,}072 tokens, and our player asked on every call for replies of up to 65{,}536, so the server refused every prompt longer than 65{,}536 tokens. Once a long tool result had carried a game's conversation that far, prime-agent's compactions were refused too, and most such games were given up and charged the whole budget. Every frozen-test game this cut, 98 over DAgger, its reset and the untrained student given only its teacher's guidance, was played again with the player fixed, on the same adapter and instance, and three were refused again and ended as they had. Replayed, DAgger saves 0.29 of the reset's spend on the held-out test rather than 0.28. On the streams the cut games stand as they ended, 7 of DAgger's 200 at one user, 19 of 400 at two and 69 of 800 at four. Scored as the same task's other visits went, its share learned would be 0.44 rather than 0.43 at one user, 0.48 rather than 0.47 at two and 0.55 rather than 0.53 at four. Through OpenRouter, whose providers allow longer windows, the window cut three games of the other methods, which stand. We fix the released player so that no request overruns the window.

\paragraph{Refused requests.} An instance played while the model provider refused requests, as OpenRouter does for an account out of credit, measures the outage rather than its learner. Every instance with a model call in such a window, which the proxy's call log gives, is left out of every measure whatever its outcome, and, since the measures pair instances across runs, out of every comparison it entered; the runs themselves played on. Two plays of a frozen test in which the provider answered nothing at all to 26 and 12 instances are left out whole, and each method played the test once more in their place. The three methods in which the teacher writes the notes played each test four times more after their first four plays met a storm of rate-limited requests, in which 9 to 11\% of their requests failed or were cancelled against 2 to 3\% in the four plays after, and their measures read the later four.

\paragraph{Readable replies.} Since the benchmark is not a test of output formatting, we count a reply as readable if it holds an action object anywhere, parsed leniently (stray or missing brackets, single quotes, trailing commas), with the action and the grid under any of the usual keys or inside a runtime's argument object or envelope. Qwen3.5-9B still often wrote none: a give-up ended 48 to 68\% of the stream's instances in every one of its harness methods, the wiped ones among them, nearly all after two re-prompts had drawn no readable action, about four actions into the instance, and about two thirds of the untrained student's instances on the tests. DAgger gave up 28\% of its stream's instances, 42\% of the held-out test's and 18\% of the seen test's; the floors, which play without a harness, 1 to 2\%; and GLM-5.3 Flash at most one of its 200 in any harness method. Every learner on Qwen3.5-9B in the harness is read against a reset that gives up as often, so a share it saves includes what it learned about writing a readable action.

\section{What the harnesses persist, and what of it carries}
\label{app:persist}

Every reading in this appendix but those of Appendices~\ref{app:persist-glm} and~\ref{app:persist-teacher} comes from GPT-6 Luna's ARC-AGI-1 cells of Appendix~\ref{app:results}, which run with the finish action of Appendix~\ref{app:harness} so that the verdict of an instance reaches the harness before it finishes the instance. No message of ours asks a harness to store. We examined each harness in its three persistent cells on the stream, one per effort, hermes-agent through its session database, and openclaw through its session log, which keeps the last 85 to 157 of each cell's 200 instances. We read the stores of these cells as the stream left them (\texttt{scripts/store\_readings.py}), so that the behavior a reading explains and the store that explains it come from the same cells.

\paragraph{Stores and retrieval.}
For each method, Table~\ref{tab:persist} lists the store that its design writes between instances and the path by which the store reaches the model when a task returns. These are properties of the designs, which we took from their code and configuration (Appendix~\ref{app:harness}), and they do not depend on any run. The three harnesses are built for lessons that transfer across unrelated work, and none of them ties its store to the task id, as the program library does. The rules of the stream tell each harness that tasks recur and that each demonstration request is charged, and we examine below what each does with this.

\begin{table}[htbp]
\centering
\footnotesize
\begin{adjustbox}{max width=\linewidth}
\begin{tabular}{@{}p{2.2cm}p{4.2cm}p{3.6cm}p{3.8cm}@{}}
\toprule
system & store written between instances & written by & how it reaches the model on a re-exposure \\
\midrule
program library (P) & one Python program per task, kept only if it reproduces every demonstration the model has, plus the demonstrations bought, up to eight per task & the learner, whenever a program passes its checks & run first; a model call only when it fails \\
prime-agent & memories, prompt notes, functions, skills, sub-agents & the agent, at will; an automatic refinement every 25 turns & a digest of the store, capped in size, injected at session start; a search tool \\
hermes-agent & skill files and a user memory & a background review after a turn with ten or more tool iterations; the agent, at will & the agent reads its skills at session start or on demand \\
openclaw & a consolidated memory, skills, workspace files & a scheduled consolidation over transcripts and a skill workshop; the agent, at will & skills listed at session start; search over the consolidated memory and over the conversation of every earlier instance \\
\bottomrule
\end{tabular}
\end{adjustbox}
\caption{What each method is built to persist between instances and how it reaches the model when a task returns, read from the designs. The subsections below give what the stores held and delivered on the cells.}
\label{tab:persist}
\end{table}

\paragraph{Harnesses without persistence.}
The stateless floor makes one model call per attempt and asks for the whole grid in that call. A harness has tools that the floor lacks, code execution above all, so it can write the candidate rule as a program and check it against the demonstrations before it costs an attempt. This advantage lies within the instance, and so it would appear at the first exposure, not grow with exposure, and survive the wipe of method H\textsubscript{0}. Paired with the stateless floor on the stream's 25 first exposures, the stateless harnesses save between $-3.8$ and 2.5 points of the cap at medium and high effort, within noise. At low effort they save between 1.8 and 17.0, within noise but for hermes-agent's 17.0 [3.0, 31.8].

\paragraph{The value of persistence.}
Whether a store can answer a task depends on what kind of store it is. An executable artifact per task, tested before it is stored, answers the task at zero model cost and cannot be misapplied, and the program library keeps such an artifact. A rule written in words answers the task only through a model that trusts and applies it. If the rule ends by telling the agent to confirm it against the demonstrations, it also costs the very demonstration request it was meant to save. A lesson about method, such as how to analyze demonstration pairs or when to stop guessing, improves every instance a little, without favoring re-exposures. Finally, a retrieval path that shows a recent window of a capped digest, or that depends on a search the agent must think to make, may not deliver even a stored rule when the task returns. From the designs alone, we would expect only a store that holds a tested, executable rule per task to bend the exposure curve below the demo-cache reference.

\subsection{prime-agent on GPT-6 Luna: nothing crossed a session}
\label{app:persist-prime}

prime-agent keeps its lessons as entries of a harness store, memories, prompt notes, functions, skills and sub-agents, and each session writes to a store of its own, which dies with it. Only an entry written with \texttt{global\_=True}, by the agent or by a refinement asked to go global, enters the global store that later sessions see, as a digest injected at their start, and every default points to the session. In the 600 sessions of its three persistent cells, the global store was never created, and after the verdict the agent wrote to its store in none of the 600 instances. Each of those sessions carried prime-agent's notice that \texttt{refine} and its other Python skills had failed to import (Appendix~\ref{app:harness}), and the automatic refinement ran in none of them on the release the cells ran, 0.10.0. Every re-exposure therefore opened on the digest of the empty store, ``No saved harness entries yet.'', as the stateless method's sessions do, and on re-exposures the two methods use the same budget within noise at every effort. prime-agent's learning across sessions is opted into write by write, and GPT-6 Luna never opted in.

\subsection{prime-agent on GLM-5.3 Flash: rules the digest seldom shows}
\label{app:persist-glm}

GLM-5.3 Flash, unlike GPT-6 Luna, wrote to prime-agent's global store. We read method G at one user from its store as the stream left it, the digest the harness injected at the start of each instance, and the events of all 200 instances (\texttt{scripts/glm\_store\_readings.py}). G saved 106 memories, a rule in words for each of the 25 tasks, and kept some tasks' rules as up to nine entries, one for each rewrite. None held a program. The harness listed at most six entries at the start of an instance, each cut off after a few lines, and it chose them before the instance's task was shown. Of the 175 returns, 154 came back to a task that already had an entry, and the digest listed it on 40 of them. On those 40, G bought 0.88 demonstrations per instance, spent 10.6\% of the budget and solved every one. On the other 114 it bought 2.11, spent 23.7\% and solved 108, close to G\textsubscript{0}, which bought 2.29 and spent 24.7\% on its returns. The agent searched or read its store on 43 of the 154 returns. Returns that the digest reached thus cost G less than half as much as those it missed, and the harness, not the model, decided which those were.

\subsection{hermes-agent: advice that changes every instance}
\label{app:persist-hermes}

hermes-agent keeps skills, markdown files with reference files beside them, and two memory files, one of notes and one about its user, which a background review writes after a turn with ten or more tool iterations. The memory files and an index of the skills sit in the system prompt of every session, with an instruction that a skill matching the task must be loaded. Its store holds general method, and files about individual tasks only at low effort, where 41 of its files are named for a task. Its skills reach the model reliably, a skill having been loaded on 175, 175 and 174 of the 175 re-exposures at low, medium and high effort, and something naming the task at hand came back on 46 of them at low effort. These retrievals did not make those re-exposures cheaper, since the 46 were solved 52\% of the time, at 72.7\% of the budget, against 81\% and 37.8 on the other 129. Its persistence instead changes how it attempts a task, with 0.72 more pairs bought per re-exposure than its stateless method at low effort (SE 0.21), and the mandatory skills displace its own tool use: the persistent method calls its code tool 2.7, 1.8 and 2.0 times per instance at low, medium and high effort, against 15.4, 13.2 and 8.4 in the stateless method. On the stream's re-exposures, persistence neither pays nor costs measurably at any effort. We suspect that advice on method helps only a model that is able to act on it, although we have not tested this directly.

\subsection{openclaw: a consolidation that cannot promote}
\label{app:persist-openclaw}

openclaw keeps one conversation for its agent, in which each of our instances opens with a reset, in a database searchable by text. Dreaming, openclaw's scheduled consolidation, ran every hour of the stream and never promoted a memory into \texttt{MEMORY.md}, which every later session reads: its gate needs three recall signals and three distinct queries from the agent's own memory searches, and no entry of its short-term store had more than two signals or a recall by the agent. The skill workshop wrote three skills in the low-effort cell and none at medium and high effort. Of the 73 re-exposures of a task it had solved before at low effort, the earlier correct answer came back on none, and at medium and high effort on none of 122 and 138. At low effort, 10 of its 11 memory searches failed, on our configuration, which left memory search unset. With memory search set, in the first 50 instances of three cells played under the earlier charge (Appendix~\ref{app:results}), nothing was promoted either, and neither its share solved nor the pairs it bought changed beyond noise. On the stream's re-exposures, persistence neither pays nor costs measurably at any effort, and what the persistent method carries is general.

\subsection{Across the three}

\paragraph{Demonstrations bought again.}
None of the harnesses keeps the demonstrations it has already paid for, as the demo-cache floor does, and so each pays for them again. On the stream, where 175 of the 200 instances are re-exposures, hermes-agent and openclaw buy 2.3 to 3.9 pairs per instance, and prime-agent 3.5, 2.9 and 2.8 at low, medium and high effort. The demo-cache floor, which sees its cached pairs again for free, still buys 2.6, 2.0 and 1.4 per instance.

\subsection{The teacher's notes on Qwen3.5-9B}
\label{app:persist-teacher}

This subsection reads Qwen3.5-9B's methods at one user in which the teacher writes the notes (Section~\ref{sec:baselines}), from their captured calls and the state the teacher's harness kept after every round, in the stream's last 50 games. QT's teacher, given nothing, wrote its notes into prime-agent's own store, 101 entries by round 190, and left 5 files in the harness's workspace. Given the expert guidance, the teacher of QTE and QETE kept each task's pairs and program in a folder and left the student a memory in the store naming them, 35 entries by round 190, beside 69 files in the workspace: the 25 tasks' programs and pairs, and 19 other notes and drafts. In all, the teacher left the student 106 notes and files in QT and 104 in QTE and QETE. The harness shows the student six of the store's entries at the start of every instance and counts the rest, so QTE's student, not given the guidance, saw its own task's memory in 10 of the 50 games; it opened its task's program in seven of the 50, six times by its first answer, and solved two without buying a demonstration. QETE's student, whose guidance tells it to look in the folder when the task's memory is not among those shown, opened one in 30, 29 times by its first answer, and solved 17 of the 50 without a demonstration. Run on each game's test input, the kept program it opened returned the right grid in 27 of those 30 games, and it solved 17 of them, 15 without a demonstration. Of the 22 games in which it opened a correct program before buying a demonstration it solved 14 at no cost, and in two of the others it ran the program on a test grid it had retyped with an error. Given to the teacher alone, the guidance put the programs where the student was told of them but seldom looked. On the seen test QETE saves 0.33 [0.23, 0.44] more of the reset's spend than QTE, while QT and QTE save no more than each other or than the harness as shipped (QT less QTE, 0.05 [$-0.03$, 0.14]; QTE less method Q, 0.02 [$-0.04$, 0.07]). On the held-out test QT and QTE read as their reset does, since notes about tasks already met cannot help with new ones. QETE reads a little below the reset, $-0.05$ [$-0.10$, 0.00]. Each of these methods reads its stream against a reset played beside it at the same hours. Four earlier plays of each test and one of each stream, in which 8 to 14\% of the model's requests failed or were cancelled against 2 to 5\% in those we read, are left out: on them QETE saves 0.11 [0.01, 0.19] on the seen test and learns 0.30 [0.17, 0.43] on the stream, and no other reading moves by more than 0.06.

\paragraph{QETE with the teacher's letter.}
\label{app:persist-letter}
To see whether what the student found in the harness held QETE back, we also played it with the teacher's teaching in the student's prompt, where it could not be missed. Given every file its state held after the stream's last round, the teacher wrote its student a letter, teaching whatever would help most on future tasks, unseen ones included. It could not decline and could write up to 131,072 tokens, the room left in the student's context, and it wrote 1,784 of the student's tokens. The letter ended the student's system prompt in every request of its own conversation. The student played both tests four times from the teacher's state at the stream's end, read against QETE's reset as QETE is. On the held-out test it saved $S_{\mathrm{held}} = -0.05$ [$-0.10$, $-0.01$], as QETE did (less QETE, $-0.01$ [$-0.05$, 0.05]), and DAgger at one user saved 0.34 [0.26, 0.43] more. On the seen test it saved 0.25 [0.15, 0.35], and QETE 0.09 [0.01, 0.18] more, a gap we have not traced. So the teacher's teaching, in every request, carried the student no further on new tasks, and on tasks already met it saved less than QETE. The plays failed or cancelled 2.0 to 2.4\% of their requests, against 1.9 to 2.8\% in QETE's, and in 17 of the 800 instances the student opened a child agent, whose prompt lacks the letter (\texttt{scripts/read\_taught.py}).

\paragraph{QETE with every file the teacher saved.}
\label{app:persist-dump}
The letter holds what the teacher chose to say, so we also played QETE with everything the teacher had saved, unselected. All 71 files that the teacher's state held after the stream's last round ended the student's system prompt in every request of its own conversation, after the line ``Here's everything that has been saved up to now. Use any of it that might be useful.'' With each grid written as rows of digits, they came to about 89,000 of the student's tokens, and the student's context window was raised by as many, so that its conversation kept the room it has in QETE. The student again played both tests four times from the teacher's state at the stream's end, read against QETE's reset. On the seen test it saved $S_{\mathrm{seen}} = 0.64$ [0.55, 0.72], which is 0.30 [0.21, 0.39] more than QETE and about what DAgger at one user saved (less DAgger, 0.06 [$-0.03$, 0.16]). On the held-out test it saved $S_{\mathrm{held}} = 0.02$ [$-0.03$, 0.07], nothing measurable, and DAgger saved 0.27 [0.19, 0.35] more. It did save 0.07 [0.01, 0.13] more there than QETE, which read a little below its reset. So with everything the teacher saved in every request, the student memorized about as much as DAgger did and still saved nothing measurable on new tasks. The plays failed or cancelled 1.0 to 1.9\% of their requests, against 1.9 to 2.8\% in QETE's, and in 5 of the 800 instances the student opened a child agent, whose prompt lacks the files (\texttt{scripts/read\_taught.py}).

\section{Harnesses asked to store, and given expert guidance}
\label{app:ablations}

Method H runs each harness as it ships: no message asks it to store, and none tells it what to keep (Section~\ref{sec:baselines}). This appendix reports methods that change that, on GPT-6 Luna's ARC-AGI-1 cells of methods H and H\textsubscript{0}, for the three harnesses at three efforts, and one method that removes the memory from the program library, method PL\textsubscript{0}, these methods named for the harness since GPT-6 Luna plays three; Qwen3.5-9B and GLM-5.3 Flash also play the first of them, and GLM-5.3 Flash the expert guidance with a library of building blocks, kept (GB) or given (the library ceiling). Section~\ref{sec:results} reads their main results, and this appendix gives them in full. On GPT-6 Luna every comparison pools the stream with the side experiment's two orders, which serve the same visits (Appendix~\ref{app:orders}), while the share learned reads the stream alone, and the paragraph on why asking pays reads the stream's cells, whose stores we read (Appendix~\ref{app:persist}). In every harness method of this appendix, openclaw runs with its memory search set, as keyword search over its memory files and session transcripts (Appendix~\ref{app:persist-openclaw}). Its H-A comparison with H\textsubscript{0} therefore also changes the search configuration.

\paragraph{Asked to store (\mbox{H-A}).}
Method H, with one message after every verdict. It asks the harness to save liberally anything that could help with the task's future instances, adds that only demonstrations and wrong attempts are scored, and gives prime-agent the fact about its global store that its rules already carry (Appendix~\ref{app:prompts}). The rules are unchanged, and a stateless method is never asked. \mbox{H-A} is read against H\textsubscript{0} on the same slots.

\paragraph{Given expert guidance (\mbox{H-E} and \mbox{H-E\textsubscript{0}}).}
Method H, with expert guidance added to its rules: the program-library policy, as a procedure that the harness carries out with its own tools. It keeps a program per task in a folder of its workspace and runs the task's program first when the task returns. Before submitting, it checks a program on every demonstration it has received, and it submits only the program's output. After the verdict it saves the program, and a message asks it to (Appendix~\ref{app:prompts}). \mbox{H-E\textsubscript{0}} receives the same messages, with its state wiped before every instance.

\paragraph{The program library with nothing kept (PL\textsubscript{0}).}
Method PL, with its kept programs and its cache of demonstrations wiped before every instance, and the one sentence of its rules about keeping programs removed. Every submission is still a program, checked free on the instance's demonstrations.

\paragraph{What the methods show.}
Table~\ref{tab:ablation-results} gives what each method saves against its reference. Asked to store, prime-agent saved 6.3 to 7.4 points of the budget against its stateless method at every effort, where method H saved nothing, and hermes-agent saved 8.0 and 7.2 points at medium and high effort. What pays is the keeping. Against \mbox{H-E\textsubscript{0}}, which follows the same procedure with nothing kept, \mbox{H-E} saved 7.6 to 13.3 points, about what keeping saves the program library, 12.5 to 14.4 points against PL\textsubscript{0}, while what the harnesses keep of their own accord saves at most 1.7 points (method H). Read as the share learned on the stream (Equation~\ref{eq:learned}), a harness given expert guidance saves 0.17 to 0.62 of its learnable budget, method PL 0.49 to 0.62, and a harness asked to store from $-0.22$ to 0.50, measurably for prime-agent at every effort and for hermes-agent at medium and high (\texttt{scripts/text\_numbers.py}).

\paragraph{Why asking pays on the stream for two harnesses and hardly for openclaw.}
A returning task is solved at cost 0, with no demonstration and the first answer right, almost only by a learner that uses something it kept about the task. On the stream's cells, whose stores we read, the stateless methods never do so, and method H on at most one of a cell's 175 returning instances, while \mbox{H-A} does so on 18, 31 and 47\% at low, medium and high effort for prime-agent, 15, 41 and 38 for hermes-agent, and on none for openclaw. \mbox{H-E} does so on 46 to 80\% for every harness, and PL on 72 to 86. openclaw did store when asked: its \mbox{H-A} cells wrote 47, 59 and 84 notes at low, medium and high effort, 40, 39 and 50 of them named for a task. What it lacks is a path back. It loads only its \texttt{MEMORY.md} at the start of a session, wrote none, and read a note named for the task on none of the 67 and 131 returning instances its replay log holds at low and medium effort, and on 1 of 131 at high. \mbox{H-E}'s rules name the file and tell the agent to look first, and every harness does so, touching the task's program before its first action on 75 to 100\% of returning instances. So openclaw follows either instruction, and only one of them tells it where its knowledge is.

\paragraph{Asked to store, on the open models.}
Qwen3.5-9B and GLM-5.3 Flash also play methods QA and GA in prime-agent, read as the main text reads its methods (Table~\ref{tab:tests-open}). Asked to store, the student's harness still saved nothing measurable, on the stream or on either test. The teacher's saved a measurable share on the stream and on the seen test, 0.31 [0.20, 0.39] and 0.26 [0.13, 0.39], where as shipped it did only over three runs (Appendix~\ref{app:runs}), but by no measurable margin over method G (GA less G, 0.19 [$-0.04$, 0.45] and 0.14 [$-0.03$, 0.29]), and nothing on the held-out test. On the instances it solved once every task had come up, method GA spent \sbGlmHaCost{}\% of the budget, against \sbGlmHCost{} as shipped and \sbGlmHzeroCost{} wiped (Table~\ref{tab:solved}). Neither model's harness kept measurably more for being asked than as shipped (\texttt{scripts/ablations\_open.py}).

\paragraph{Expert guidance as building blocks (GB).}
\label{app:blocks}
The Re-ARC generators compose the stream's rules from one library of primitives, 122 of its 160 at one user, so a library of small, shared functions is what could carry a harness from the seen tasks to the held-out ones (Section~\ref{sec:results}). Method GB gives GLM-5.3 Flash the expert guidance's procedure with its library kept as building blocks: a file \texttt{lib.py} of small, general grid functions, each task's program a short composition of them, and after each verdict on a working program, the parts another task could reuse moved into \texttt{lib.py}. As examples it is shown six Re-ARC primitives that no rule of the stream uses, and two other tasks' programs composed from them. GB\textsubscript{0}, the same wiped before every instance, is shown the same, and GB is read against it. GB saved nothing on the held-out test, $S_{\mathrm{held}} = -0.04$ [$-0.16$, 0.07], where GE saved 0.07 [$-0.02$, 0.16], and on the seen test it saved 0.64 [0.48, 0.78] against GE's 0.74 [0.64, 0.84], by no measurable margin either way (GB less GE, $-0.11$ [$-0.27$, 0.05] and $-0.10$ [$-0.30$, 0.06]). GLM followed the guidance in letter only. At the end of the stream \texttt{lib.py} held 75 functions and all 25 kept programs imported it, but 21 of them were a single call to their task's whole solution saved there, and only one function, \texttt{background}, was imported by two programs. Told to keep building blocks, GLM kept one whole program per task all the same (\texttt{scripts/read\_blocks.py}, \texttt{scripts/block\_library.py}).

\paragraph{The library ceiling.}
\label{app:ceiling}
GB's library never became building blocks, so we also gave GLM-5.3 Flash the library the stream's rules are built from, the most a harness could keep of them. To GE\textsubscript{0}, which follows the expert guidance with its state wiped before every instance, we handed the Re-ARC primitives that the stream's rules use, with every primitive they call: 131 functions at one user and 153 at four. Each instance's opening message showed the library whole, about 8,400 and 10,000 tokens, with the line that imports it, and the workspace held it as a module. In one held-out play prime-agent compacted the conversation of 22 of the 100 instances, and its summaries kept none of the source. Each library played the held-out test four times, and the one-user library the seen test too, against GE\textsubscript{0} without a library played in the same hours. Neither library saved anything measurable on the held-out test, $S_{\mathrm{held}} = -0.06$ [$-0.17$, 0.04] and $-0.09$ [$-0.20$, 0.01]. Nor did the one-user library save on the seen test, $S_{\mathrm{seen}} = -0.04$ [$-0.16$, 0.07], although that test's rules are built from its own functions. GLM used the library in 33 to 36 of each test's 400 instances, wrote its other programs in plain Python, and spent no less when it did use it. Every instance buys at least one demonstration, and GE\textsubscript{0} spent 1.86 of its 2.60 per held-out instance on demonstrations, which no library replaces. So handed the stream's building blocks, GLM saved nothing on new tasks, which bounds what a harness storing them could give it here, but not what they could give a model that used them (\texttt{scripts/read\_ceiling.py}, \texttt{scripts/ceiling\_use.py}).

\begin{table}[htbp]
\centering
\footnotesize\setlength{\tabcolsep}{3pt}
\begin{tabular}{lr@{\hspace{0.4em}}lr@{\hspace{0.4em}}lr@{\hspace{0.4em}}l}
\toprule
method & \multicolumn{2}{c}{$L$} & \multicolumn{2}{c}{$S_{\mathrm{seen}}$} & \multicolumn{2}{c}{$S_{\mathrm{held}}$} \\
\midrule
\multicolumn{7}{l}{\emph{Qwen3.5-9B, notes in the harness}} \\
Q & -0.02 & {\scriptsize [-0.09, 0.04]} & -0.01 & {\scriptsize [-0.04, 0.03]} & -0.01 & {\scriptsize [-0.05, 0.03]} \\
QA & -0.05 & {\scriptsize [-0.11, 0.01]} & +0.02 & {\scriptsize [-0.02, 0.07]} & 0.00 & {\scriptsize [-0.05, 0.04]} \\
QE & +0.05 & {\scriptsize [-0.06, 0.16]} & +0.03 & {\scriptsize [-0.08, 0.13]} & -0.01 & {\scriptsize [-0.05, 0.03]} \\
\midrule
\multicolumn{7}{l}{\emph{Qwen3.5-9B, notes the teacher writes}} \\
QT & +0.05 & {\scriptsize [-0.02, 0.14]} & +0.06 & {\scriptsize [-0.01, 0.14]} & +0.03 & {\scriptsize [-0.02, 0.07]} \\
QTE & +0.07 & {\scriptsize [-0.03, 0.16]} & +0.01 & {\scriptsize [-0.05, 0.07]} & -0.02 & {\scriptsize [-0.07, 0.03]} \\
QETE & \textbf{+0.29} & {\scriptsize [0.19, 0.40]} & \textbf{+0.34} & {\scriptsize [0.26, 0.42]} & -0.05 & {\scriptsize [-0.10, 0.00]} \\
\bottomrule
\end{tabular}
\\[8pt]
\begin{tabular}{lr@{\hspace{0.4em}}lr@{\hspace{0.4em}}lr@{\hspace{0.4em}}l}
\toprule
method & \multicolumn{2}{c}{$L$} & \multicolumn{2}{c}{$S_{\mathrm{seen}}$} & \multicolumn{2}{c}{$S_{\mathrm{held}}$} \\
\midrule
\multicolumn{7}{l}{\emph{GLM-5.3 Flash, notes in the harness}} \\
G & +0.12 & {\scriptsize [-0.16, 0.36]} & +0.12 & {\scriptsize [-0.03, 0.27]} & 0.00 & {\scriptsize [-0.11, 0.10]} \\
GA & +0.31 & {\scriptsize [0.20, 0.39]} & +0.26 & {\scriptsize [0.13, 0.39]} & -0.03 & {\scriptsize [-0.14, 0.06]} \\
GE & +0.62 & {\scriptsize [0.49, 0.75]} & +0.74 & {\scriptsize [0.64, 0.84]} & +0.07 & {\scriptsize [-0.02, 0.16]} \\
\midrule
\multicolumn{7}{l}{\emph{GLM-5.3 Flash, baselines}} \\
GP & \textbf{+0.68} & {\scriptsize [0.49, 0.87]} & \textbf{+0.81} & {\scriptsize [0.59, 1.00]} & +0.05 & {\scriptsize [-0.06, 0.14]} \\
GC\textsubscript{0} & 0 & & -0.06 & {\scriptsize [-0.16, 0.04]} & +0.03 & {\scriptsize [-0.10, 0.14]} \\
GC & +0.41 & {\scriptsize [0.29, 0.53]} & +0.44 & {\scriptsize [0.34, 0.57]} & -0.02 & {\scriptsize [-0.11, 0.06]} \\
G\textsubscript{0} & 0 & & -0.05 & {\scriptsize [-0.23, 0.11]} & +0.13 & {\scriptsize [-0.01, 0.26]} \\
GE\textsubscript{0} & 0 & & -0.06 & {\scriptsize [-0.24, 0.07]} & -0.01 & {\scriptsize [-0.15, 0.12]} \\
GP\textsubscript{0} & 0 & & 0.00 & {\scriptsize [-0.16, 0.13]} & -0.05 & {\scriptsize [-0.22, 0.09]} \\
\bottomrule
\end{tabular}

\caption{Every method on Qwen3.5-9B that keeps notes in the harness at one user, QA among them (top), and every method on GLM-5.3 Flash at one user, GA and the methods that keep nothing among them (bottom), read as in Table~\ref{tab:tests}. Bold marks each column's best among the values whose 95\% interval lies above zero.}
\label{tab:tests-open}
\end{table}

\paragraph{The baselines of Figure~\ref{fig:plane}.}
\label{app:plane-baselines}
The program library is read against itself wiped (QP\textsubscript{0} and GP\textsubscript{0}) and the demo-cache floor against the stateless floor, and on GLM-5.3 Flash the two references differ. Checking each program free on the pairs before it is submitted cuts that model's wrong answers on the seen test from 2.97 per instance to 0.88 with nothing kept, so GP\textsubscript{0} spends 19.2\% of the budget there against the stateless floor's 39.0. For Qwen3.5-9B the check helps little (4.97 wrong answers per instance against 5.32), and its two references spend alike, 64.9 and 66.5\%. As a share of what the reset spends (Equation~\ref{eq:transfer}), the program library saves more than the demo-cache floor on both models, 0.81 against 0.44 on GLM-5.3 Flash and 0.65 against 0.37 on Qwen3.5-9B, by 0.36 [0.16, 0.57] and 0.29 [0.16, 0.43]. On a task it has not met, neither baseline can change anything, and on the held-out test neither reads beyond chance on either model, from $-0.02$ to 0.05 (Table~\ref{tab:tests}). Plays of one system differ about as their instances do: across the one-user methods' 46 sets of four plays, the variance of the plays' mean charges over what independent instances predict has a median of 0.5 to 1.1 by model and test, the largest 2.8 (Appendix~\ref{app:persist-teacher}), so the bootstrap over tasks carries the noise of playing a test (\texttt{scripts/ablations\_open.py}); since some of a panel's intervals would still exclude zero by chance alone, its methods are judged together (Section~\ref{sec:metrics}).

\begin{table}[htbp]
\centering
\small\setlength{\tabcolsep}{7pt}
\begin{adjustbox}{max width=\linewidth}
\begin{tabular}{llrrr}
\toprule
 & method & \multicolumn{1}{c}{low} & \multicolumn{1}{c}{medium} & \multicolumn{1}{c}{high} \\
\midrule
prime-agent & H, as it ships & \makebox[2.15em][r]{-1.0}\makebox[2.3em][l]{\hspace{0.45em}\scriptsize$\pm$0.8} & \makebox[2.15em][r]{-0.2}\makebox[2.3em][l]{\hspace{0.45em}\scriptsize$\pm$0.8} & \makebox[2.15em][r]{-0.3}\makebox[2.3em][l]{\hspace{0.45em}\scriptsize$\pm$0.6} \\
 & H-A, asked to store & \makebox[2.15em][r]{+6.3}\makebox[2.3em][l]{\hspace{0.45em}\scriptsize$\pm$1.3} & \makebox[2.15em][r]{+7.4}\makebox[2.3em][l]{\hspace{0.45em}\scriptsize$\pm$1.5} & \makebox[2.15em][r]{+6.6}\makebox[2.3em][l]{\hspace{0.45em}\scriptsize$\pm$1.4} \\
 & H-E\textsubscript{0}, expert guidance, nothing kept & \makebox[2.15em][r]{+5.0}\makebox[2.3em][l]{\hspace{0.45em}\scriptsize$\pm$1.7} & \makebox[2.15em][r]{+2.9}\makebox[2.3em][l]{\hspace{0.45em}\scriptsize$\pm$1.1} & \makebox[2.15em][r]{+2.4}\makebox[2.3em][l]{\hspace{0.45em}\scriptsize$\pm$0.9} \\
 & H-E, expert guidance & \makebox[2.15em][r]{+17.2}\makebox[2.3em][l]{\hspace{0.45em}\scriptsize$\pm$3.1} & \makebox[2.15em][r]{+13.1}\makebox[2.3em][l]{\hspace{0.45em}\scriptsize$\pm$1.6} & \makebox[2.15em][r]{+12.4}\makebox[2.3em][l]{\hspace{0.45em}\scriptsize$\pm$1.5} \\
 & H-E against H-E\textsubscript{0}, what keeping adds & \makebox[2.15em][r]{+12.2}\makebox[2.3em][l]{\hspace{0.45em}\scriptsize$\pm$2.6} & \makebox[2.15em][r]{+10.2}\makebox[2.3em][l]{\hspace{0.45em}\scriptsize$\pm$1.4} & \makebox[2.15em][r]{+10.1}\makebox[2.3em][l]{\hspace{0.45em}\scriptsize$\pm$1.1} \\
 & H-E against PL & \makebox[2.15em][r]{-5.3}\makebox[2.3em][l]{\hspace{0.45em}\scriptsize$\pm$2.5} & \makebox[2.15em][r]{-1.9}\makebox[2.3em][l]{\hspace{0.45em}\scriptsize$\pm$1.2} & \makebox[2.15em][r]{-1.2}\makebox[2.3em][l]{\hspace{0.45em}\scriptsize$\pm$1.8} \\
\midrule
hermes-agent & H, as it ships & \makebox[2.15em][r]{-3.2}\makebox[2.3em][l]{\hspace{0.45em}\scriptsize$\pm$1.7} & \makebox[2.15em][r]{+1.7}\makebox[2.3em][l]{\hspace{0.45em}\scriptsize$\pm$0.8} & \makebox[2.15em][r]{-1.7}\makebox[2.3em][l]{\hspace{0.45em}\scriptsize$\pm$1.0} \\
 & H-A, asked to store & \makebox[2.15em][r]{-2.7}\makebox[2.3em][l]{\hspace{0.45em}\scriptsize$\pm$2.2} & \makebox[2.15em][r]{+8.0}\makebox[2.3em][l]{\hspace{0.45em}\scriptsize$\pm$1.7} & \makebox[2.15em][r]{+7.2}\makebox[2.3em][l]{\hspace{0.45em}\scriptsize$\pm$1.3} \\
 & H-E\textsubscript{0}, expert guidance, nothing kept & \makebox[2.15em][r]{-5.1}\makebox[2.3em][l]{\hspace{0.45em}\scriptsize$\pm$1.7} & \makebox[2.15em][r]{+0.1}\makebox[2.3em][l]{\hspace{0.45em}\scriptsize$\pm$1.4} & \makebox[2.15em][r]{+0.4}\makebox[2.3em][l]{\hspace{0.45em}\scriptsize$\pm$1.1} \\
 & H-E, expert guidance & \makebox[2.15em][r]{+2.5}\makebox[2.3em][l]{\hspace{0.45em}\scriptsize$\pm$2.9} & \makebox[2.15em][r]{+13.4}\makebox[2.3em][l]{\hspace{0.45em}\scriptsize$\pm$2.6} & \makebox[2.15em][r]{+10.9}\makebox[2.3em][l]{\hspace{0.45em}\scriptsize$\pm$1.7} \\
 & H-E against H-E\textsubscript{0}, what keeping adds & \makebox[2.15em][r]{+7.6}\makebox[2.3em][l]{\hspace{0.45em}\scriptsize$\pm$2.1} & \makebox[2.15em][r]{+13.3}\makebox[2.3em][l]{\hspace{0.45em}\scriptsize$\pm$2.1} & \makebox[2.15em][r]{+10.5}\makebox[2.3em][l]{\hspace{0.45em}\scriptsize$\pm$1.6} \\
 & H-E against PL & \makebox[2.15em][r]{-17.1}\makebox[2.3em][l]{\hspace{0.45em}\scriptsize$\pm$2.8} & \makebox[2.15em][r]{-5.0}\makebox[2.3em][l]{\hspace{0.45em}\scriptsize$\pm$1.2} & \makebox[2.15em][r]{-4.7}\makebox[2.3em][l]{\hspace{0.45em}\scriptsize$\pm$1.3} \\
\midrule
openclaw & H, as it ships & \makebox[2.15em][r]{+0.8}\makebox[2.3em][l]{\hspace{0.45em}\scriptsize$\pm$1.3} & \makebox[2.15em][r]{+0.5}\makebox[2.3em][l]{\hspace{0.45em}\scriptsize$\pm$1.4} & \makebox[2.15em][r]{+0.1}\makebox[2.3em][l]{\hspace{0.45em}\scriptsize$\pm$0.9} \\
 & H-A, asked to store & \makebox[2.15em][r]{+1.1}\makebox[2.3em][l]{\hspace{0.45em}\scriptsize$\pm$1.3} & \makebox[2.15em][r]{+2.8}\makebox[2.3em][l]{\hspace{0.45em}\scriptsize$\pm$1.3} & \makebox[2.15em][r]{+1.2}\makebox[2.3em][l]{\hspace{0.45em}\scriptsize$\pm$1.2} \\
 & H-E\textsubscript{0}, expert guidance, nothing kept & \makebox[2.15em][r]{+7.2}\makebox[2.3em][l]{\hspace{0.45em}\scriptsize$\pm$2.3} & \makebox[2.15em][r]{+5.9}\makebox[2.3em][l]{\hspace{0.45em}\scriptsize$\pm$1.8} & \makebox[2.15em][r]{+1.2}\makebox[2.3em][l]{\hspace{0.45em}\scriptsize$\pm$1.8} \\
 & H-E, expert guidance & \makebox[2.15em][r]{+16.6}\makebox[2.3em][l]{\hspace{0.45em}\scriptsize$\pm$4.0} & \makebox[2.15em][r]{+13.7}\makebox[2.3em][l]{\hspace{0.45em}\scriptsize$\pm$2.8} & \makebox[2.15em][r]{+14.3}\makebox[2.3em][l]{\hspace{0.45em}\scriptsize$\pm$1.6} \\
 & H-E against H-E\textsubscript{0}, what keeping adds & \makebox[2.15em][r]{+9.3}\makebox[2.3em][l]{\hspace{0.45em}\scriptsize$\pm$2.6} & \makebox[2.15em][r]{+7.8}\makebox[2.3em][l]{\hspace{0.45em}\scriptsize$\pm$1.8} & \makebox[2.15em][r]{+13.1}\makebox[2.3em][l]{\hspace{0.45em}\scriptsize$\pm$1.9} \\
 & H-E against PL & \makebox[2.15em][r]{-4.7}\makebox[2.3em][l]{\hspace{0.45em}\scriptsize$\pm$2.2} & \makebox[2.15em][r]{-5.6}\makebox[2.3em][l]{\hspace{0.45em}\scriptsize$\pm$1.2} & \makebox[2.15em][r]{-1.6}\makebox[2.3em][l]{\hspace{0.45em}\scriptsize$\pm$1.5} \\
\midrule
bare model & PL\textsubscript{0} against the stateless floor & \makebox[2.15em][r]{+17.8}\makebox[2.3em][l]{\hspace{0.45em}\scriptsize$\pm$3.5} & \makebox[2.15em][r]{+10.1}\makebox[2.3em][l]{\hspace{0.45em}\scriptsize$\pm$2.7} & \makebox[2.15em][r]{+8.3}\makebox[2.3em][l]{\hspace{0.45em}\scriptsize$\pm$2.5} \\
 & PL against the stateless floor & \makebox[2.15em][r]{\textbf{+31.9}}\makebox[2.3em][l]{\hspace{0.45em}\textbf{\scriptsize$\pm$4.6}} & \makebox[2.15em][r]{\textbf{+24.5}}\makebox[2.3em][l]{\hspace{0.45em}\textbf{\scriptsize$\pm$3.9}} & \makebox[2.15em][r]{\textbf{+20.8}}\makebox[2.3em][l]{\hspace{0.45em}\textbf{\scriptsize$\pm$3.2}} \\
 & PL against PL\textsubscript{0} & \makebox[2.15em][r]{+14.1}\makebox[2.3em][l]{\hspace{0.45em}\scriptsize$\pm$2.0} & \makebox[2.15em][r]{+14.4}\makebox[2.3em][l]{\hspace{0.45em}\scriptsize$\pm$2.4} & \makebox[2.15em][r]{+12.5}\makebox[2.3em][l]{\hspace{0.45em}\scriptsize$\pm$1.5} \\
\bottomrule
\end{tabular}

\end{adjustbox}
\caption{ARC-AGI-1: budget saved by each method against its reference, in percent of the cap (higher is better), on the stream slots both methods have reached, pooled with the side experiment's two orders, with a bootstrap standard error over tasks in small type. A harness method is read against the same harness's stateless method H\textsubscript{0}, except for the last two lines of each harness, which read \mbox{H-E} against \mbox{H-E\textsubscript{0}} and against method PL; the bare model's methods are read as named. Bold marks each column's best among the savings whose 95\% interval, 1.96 standard errors either side, lies above zero. \ablationnote}
\label{tab:ablation-results}
\end{table}

\section{Related work: extended table}
\label{app:related}

In Table~\ref{tab:related-full}, we place the benchmarks and methods closest to this one, out of those we screened, against the properties that this benchmark combines.  Other benchmarks cover aspects that ours does not, such as drift in real domains \citep{lomonaco2017core50,lin2022clear}, natural-language feedback \citep{ai2025memorybench} and side effects on general ability \citep{wang2023trace}.

\begin{sidewaystable}[p]
\centering
\scriptsize
\setlength{\tabcolsep}{4pt}
\begin{adjustbox}{max width=\textheight}
\begin{tabular}{llllllll}
\toprule
benchmark & domain & instances & recurrence & growth & feedback & learners & score \\
\midrule
\multicolumn{8}{l}{\emph{Supervised continual learning}} \\
Permuted/Split MNIST \citep{goodfellow2013empirical,zenke2017continual} & image classes & fixed & none & no & labels & W & accuracy \\
GEM \citep{lopezpaz2017gradient} & image classes & fixed & none & no & labels & W & ACC, BWT, FWT \\
A-GEM \citep{chaudhry2019efficient} & image classes & fixed & none & no & labels & W & accuracy, forgetting, LCA \\
CTrL \citep{veniat2021efficient} & image streams & fixed & one revisit at end & no & labels & W & accuracy, transfer \\
OSAKA \citep{caccia2020online} & few-shot classes & fixed pools & hidden revisits & new tasks & labels after & W & cumulative online acc. \\
CIR \citep{hemati2023class,hemati2025repetition} & image classes & generated streams & class repetition & classes & labels & W & average accuracy \\
CORe50 \citep{lomonaco2017core50} & objects (video) & fixed & sessions & classes & labels & W & accuracy curve \\
CLEAR \citep{lin2022clear} & temporal images & fixed & none & no & labels & W & in/next-domain acc. \\
NEVIS'22 \citep{bornschein2022nevis} & 106 vision tasks & fixed & none & yes (chronological) & labels & any & accuracy vs compute \\
Loss of plasticity \citep{dohare2024loss} & thousands of tasks & fixed & none & unbounded & labels, reward & W & online accuracy \\
\midrule
\multicolumn{8}{l}{\emph{Continual RL and generated environments}} \\
Continual World \citep{wolczyk2021continual} & manipulation & random init & second pass (CW20) & no & dense reward & W & perf., forgetting, transfer \\
CORA \citep{powers2022cora} & Atari, Procgen, MiniHack & levels & fixed cycles & no & reward & W & continual eval, forgetting \\
COOM \citep{tomilin2023coom} & 3D ViZDoom & fixed & none & no & reward & W & perf., forgetting, transfer \\
MEAL \citep{tomilin2026meal} & Overcooked & generated layouts & none & no & reward & W & perf., forgetting, plasticity \\
L2M metrics \citep{new2022lifelong} & any & scenario & learning blocks & no & reward & any & PM, FT, BT, RP, SE \\
LIBERO \citep{liu2023libero} & manipulation & fixed & none & no & 50 demos, free & W & FWT, NBT, AUC \\
Jelly Bean World \citep{platanios2020jelly} & never-ending grid & generated & no tasks & no & reward & W & reward over time \\
Procgen \citep{cobbe2020procgen} & 16 games & generated levels & not continual & no & reward & W & sample eff., generalization \\
XLand \citep{oel2021xland,ada2023humantimescale} & 3D task space & generated & trials & no & reward & W & held-out task success \\
BabyAI \citep{chevalierboisvert2019babyai} & grounded language & generated & curriculum & no & reward, demos & W & demos per level \\
Reasoning Gym \citep{stojanovski2025reasoninggym} & 100+ reasoning tasks & generated & train/test & no & verifier & W & accuracy \\
Enigmata \citep{chen2025enigmata} & 36 puzzles & generated & train/test & no & verifier & W & accuracy \\
RLVE \citep{zeng2026rlve} & 400 verifiable envs & generated, adaptive & training curriculum & no & verifier & W & accuracy \\
\midrule
\multicolumn{8}{l}{\emph{Language agents}} \\
StreamBench \citep{wu2024streambench} & SQL, code, tools, QA & fixed, one pass & none & no & binary & P, M & final accuracy \\
LifelongAgentBench \citep{zheng2025lifelongagentbench} & DB, OS, KG & fixed & skills recur & no & success & P, M & success rate \\
SWE-Bench-CL \citep{joshi2025swebenchcl} & GitHub issues & fixed & none & no & tests & M & ACC, forgetting, BWT \\
MemoryBench \citep{ai2025memorybench} & QA, legal, academic & fixed & none & no & simulated user & M & normalized effectiveness \\
CL-Bench \citep{asawa2026clbench} & six real domains & fixed & poker stages & no & reward, tests & P, M & gain over stateless \\
SkillLearnBench \citep{zhong2026skilllearnbench} & real skill tasks & fixed, 2--6 each & none & no & verifier, self & P, M & skill quality, accuracy \\
AgentStream \citep{yan2026agentstream} & six agent benchmarks & fixed, one pass & none & no & execution, self-evaluation & P, M & task accuracy \\
AgentCL \citep{shu2026agentcl} & code, research, BabyAI & fixed & one second pass & no & none & M & plasticity, stability, generalization gains \\
PATH-Bench \citep{yang2026pathbench} & code, tools & fixed & probe every 6--12 & no & none & M & AP, FWT, BWT \\
Evo-Memory \citep{wei2025evomemory} & QA, math, embodied & fixed & similar tasks & no & correctness & M & accuracy, ordering \\
ContinualSkillBench \citep{guan2026continualskillbench} & five domains & fixed & none & no & evaluator & P, M & reward delta \\
SEA-Eval \citep{jiang2026seaeval} & web, OS & variable slots & repeats, variants & no & outcome & P, M & success, efficiency \\
EvoArena \citep{xu2026evoarena} & terminal, SWE & fixed chains & evolved versions & no & execution & M & chain accuracy \\
StuLife \citep{cai2025ell} & simulated college & fixed & exams revisit & no & scores, text & P, M & GPA, retention \\
TRACE \citep{wang2023trace} & 8 NLP tasks & fixed & none & no & labels & W & OP, BWT, side effects \\
CITB \citep{zhang2023citb} & dialogue tasks & fixed & none & no & labels & W & ROUGE, FWT, BWT \\
\midrule
\multicolumn{8}{l}{\emph{ARC}} \\
ARC-AGI-1 \citep{chollet2019measure} & grids & fixed & none & no & none (2 tries) & any & accuracy \\
ARC-AGI-2 \citep{chollet2025arcagi2} & grids & fixed & none & no & none (2 tries) & any & accuracy, cost/task \\
ARCLE \citep{lee2024arcle} & grid editing MDP & fixed & none & no & step reward & W & per-task success \\
ArcMemo \citep{ho2025arcmemo} & ARC-AGI-1 (100) & original & none & no & train pairs & M & accuracy \\
SOAR \citep{pourcel2025soar} & ARC-AGI-1 eval & original & every task, 2 test-time rounds & no & train pairs & W & accuracy \\
PoE \citep{franzen2025poe} & ARC-AGI-1 eval & original & none (per task) & no & train pairs & W & accuracy \\
ARC-AGI-3 \citep{arcprize2026arcagi3} & grid games & fixed games & none scored & no & observations & any & RHAE \\
\midrule
\bench{} & ARC-AGI-1 to 3 & generated & bursts, variable gaps & yes & binary, paid demos & P, M, W & share learned, share saved; budget used, solved \\
\bottomrule
\end{tabular}
\end{adjustbox}
\caption{Extended comparison. Instances: whether each exposure draws a fresh instance. Recurrence: whether and how tasks come back. Growth: whether the task set grows through the stream. Learners: which families the protocol admits, P prompting, M external memory, W weight updates; LCA is learning-curve area, RHAE is ARC-AGI-3's human-relative action efficiency.}
\label{tab:related-full}
\end{sidewaystable}

\section{The methods on the open models}
\label{app:trained}

DAgger trains Qwen3.5-9B on the stream, each run on a node of eight H100 GPUs rented for it. A vLLM server on the node serves the base model and every adapter by name, and the student plays every move through prime-agent 0.9.6 under the same rules as method Q. It samples at the setting of Qwen3.5-9B's model card for thinking mode (temperature 1, top-$p$ 0.95, top-$k$ 20, presence penalty 1.5), as do the untrained student that DAgger is read against and the student told the page alone (Appendix~\ref{app:told}): at temperature 1 with no truncation, as the methods served through OpenRouter sample, a trained adapter's replies ran into incoherent text up to the token cap, which they never did at the card's setting. Every model call is saved as it was served, and every game's outcome goes to the run's records, as for every other method.

\paragraph{The schedule.}
Training waits until the base model has played 24 games. From then on, after each round, a phase trains a LoRA adapter \citep{hu2022lora} of rank 128 and alpha 128 on the query, key, value, output, gate, up and down projections, starting from the last one, on every game the stream has given so far, with one game in ten held out, and it stops once the held-out measure stops improving. The round after next plays with the new adapter, a lag of one round, so that play never waits for training. Every phase's adapter is kept, so the student of any round can be restored and played again. It trains at a learning rate of $10^{-4}$, with 64 of the student's calls to each optimizer step. Its held-out measure is the likelihood of the teacher's labels on the held-out games, taken before the first step and every five steps after it; the phase keeps the adapter at the best of them, and it ends once six in a row have not improved on it, or at 120 steps.

\begin{figure}[htbp]
\centering
\includegraphics[width=\linewidth]{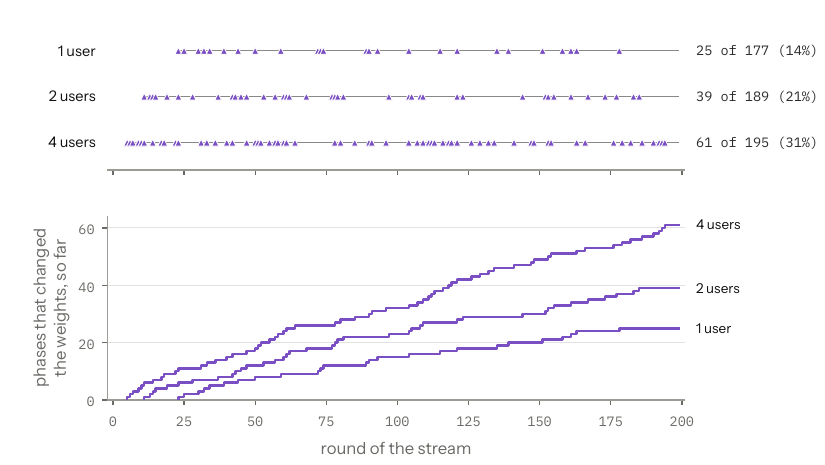}
\caption{How often DAgger's phases changed Qwen3.5-9B's adapter at one, two and four users. Above, a triangle marks each round after which the phase kept a later checkpoint than the adapter it started from, on a rule over the rounds that had a phase, which begin once the base model has played 24 games; on the right, how many of the phases did. Below, how many had by each round.}
\label{fig:weight-updates}
\end{figure}

Most of DAgger's phases kept the adapter they started from, since no later measure improved on the first: 152 of the 177 at one user, and 90 of the 100 from round 100. The held-out measure is doing its work there: once the adapter has learned what the games so far teach, further steps only fit the labels it has already seen, and from round 50 at one user the held-out games' negative log-likelihood per token averaged 0.14 to 0.20 above that of the labels the adapter trained on, against 0.02 before. How often a phase still found something to learn grew with the number of users (Figure~\ref{fig:weight-updates}): 25 of the 177 phases changed the adapter at one user, 39 of the 189 at two and about a third at four, where nearly one in three still did after round 100. A single user's stream soon brings little that it has not already taught, so most of its later phases spend their compute and change nothing, while more users bring more new games each round for the student to learn from. The variety of the stream and its size grow together with the users, so the figure does not tell the two apart.

\begin{figure}[htbp]
\centering
\includegraphics[width=\linewidth]{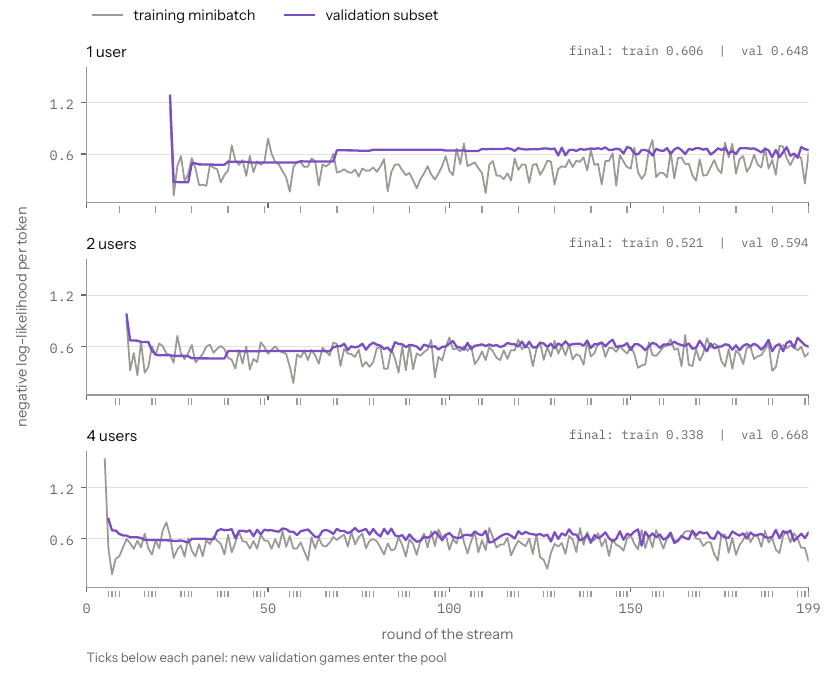}
\caption{QDE's training and validation token loss at the start of each phase, with common axes and no smoothing. Ticks mark new games entering the validation pool; its sampled subset can also change between ticks. Values above each panel are from the final phase, which kept its incoming adapter in all three runs.}
\label{fig:weight-losses}
\end{figure}

\paragraph{Training and validation loss.}
Figure~\ref{fig:weight-losses} compares the same adapter at the start of each phase: its first training minibatch's loss, before any update, and its validation loss. Validation draws whole games from the accumulated pool to reach about 1,000 labeled calls, with a new sample each phase. The training curve is a minibatch reading, since full training-set evaluations were not logged. Both the data and the student states differ across user counts, so these curves do not establish which learns more on a common test.

\paragraph{DAgger with an expert-guided teacher (DE).}
In each state that the student met, the teacher, GLM-5.3 Flash at medium effort, is asked for its own reply, reading the playing rules alone, the expert guidance without its program-library procedure. Its reply, reasoning included, is the target of a supervised loss on the student's own context \citep{ross2011dagger}. Methods QE and GE receive those rules plus the program-library procedure (Appendix~\ref{app:prompts}). DAgger's student sees neither directly. At one user the teacher was asked about 21,096 of the student's moments, and 21,047 of its replies became training examples: 39 moments got no label after every re-ask, and 10 labels did not render. It took 1.07 asks per label, and 94\% of labels came at the first ask. Those 21,047 replies come to 21.3 million of the student's tokens, reasoning included, which is 162 times the window of 131{,}072 tokens that our server serves it with (Appendix~\ref{app:harness}), far more than it could be shown in context.

\paragraph{Other weight-based methods.}
\label{app:other-weights}
We also trained the student by a policy gradient on each game's outcome \citep{schulman2017ppo} and by self-distillation from the teacher's move \citep{shenfeld2026sdft}, but got neither learning reliably. The policy gradient, learning only from the games the student wins, stalled after its few early wins; in self-distillation the student often read the teacher's move inside the tool's output as a move already made, learned that and played worse than untrained.

\paragraph{How the configuration was chosen.} The search that set DAgger's final configuration scored its candidates on 100 ARC-AGI-1 evaluation tasks that the stream never serves, 23 of which are among the held-out test's, and played some of them on both tests as well. Leaving those 23 tasks out, DAgger saves 0.27 [0.20, 0.34] of the reset's spend on the held-out test at one user and 0.39 [0.32, 0.46] at two, against 0.29 and 0.40 on all 100.

\paragraph{Training compute.}
At one user DAgger ran 177 training phases, one after each round from round 23, for 84.7 hours on two to four of its node's eight H100 GPUs: 263 GPU-hours, \$1,182 at the node's on-demand price of \$4.50 per GPU-hour, and the teacher's labels cost \$71 more. Both are apart from the cost of serving the student, which four other GPUs of the node do.

\paragraph{The page alone.}
\label{app:told}
The untrained student also plays both tests told only the playing rules its teacher reads, wiped before every instance as Q\textsubscript{0} is, four plays of each at the model card's setting, as DAgger and the reset play them. On the held-out test it solves 22 to 33 of the 100 tasks per play, against 13 to 26 untrained and 47 to 52 for DAgger. It answers before seeing a demonstration on 5\% of instances, against 35\% untrained, and DAgger never does; it asks for 1.6 demonstrations per instance and makes 2.2 wrong answers, against 1.1 and 2.5 untrained. It saves 0.09 [0.05, 0.14] of the untrained student's spend, and DAgger 0.19 [0.13, 0.26] more. On the seen test, where it keeps nothing of the stream, it saves 0.16 [0.08, 0.26], against DAgger's 0.57 [0.48, 0.67], DAgger 0.41 [0.32, 0.50] more.

\paragraph{The teacher on solved instances.}
Table~\ref{tab:solved} gives GLM-5.3 Flash's methods from round 100, when every task has been met, as Section~\ref{sec:results} reads them, beside its stateless floor.

\begin{table}[htbp]
\centering
\footnotesize\setlength{\tabcolsep}{3pt}
\begin{adjustbox}{max width=\linewidth}
\begin{tabular}{lrr@{\hspace{0.4em}}lrrr}
\toprule
system & solved (\%) & \multicolumn{2}{c}{budget a solve used (\%)} & solved at no cost (\%) & demonstrations & wrong answers \\
\midrule
\multicolumn{7}{l}{\emph{GLM-5.3 Flash (medium), prime-agent}} \\
GC\textsubscript{0} & 83 & 20.5 & {\scriptsize [16.0, 25.7]} & 0 & 1.8 & 1.5 \\
G\textsubscript{0} & 93 & 17.5 & {\scriptsize [14.3, 20.7]} & 0 & 1.9 & 0.9 \\
G & 94 & 13.6 & {\scriptsize [9.3, 18.3]} & 23 & 1.4 & 0.7 \\
GA & 98 & 16.0 & {\scriptsize [12.4, 20.0]} & 9 & 1.7 & 0.9 \\
GE & \textbf{99} & 4.6 & {\scriptsize [2.8, 6.3]} & 74 & 0.4 & 0.3 \\
GP & \textbf{99} & \textbf{1.4} & {\scriptsize [0.3, 2.9]} & \textbf{92} & \textbf{0.1} & \textbf{0.1} \\
\bottomrule
\end{tabular}

\end{adjustbox}
\caption{GLM-5.3 Flash in prime-agent, rounds 100 to 199 of the stream at one user: the share of instances solved, the budget a solve used in percent of the budget with its 95\% interval over task resamples, the share of solves that used none of it, and the demonstrations and wrong answers behind a solve. Perfect memory on any of these systems spends nothing here. Bold marks each column's best.}
\label{tab:solved}
\end{table}

\paragraph{The two tests in absolute terms.}
A share saved is a share of the reset's own spend, so a learner whose reset already spends little can save a large share of very little, and a share is read beside the budgets it comes from. Table~\ref{tab:tests-absolute} gives every method of Table~\ref{tab:tests} in those terms: the budget used on each test and the share of the held-out test solved, the learner's own beside the reset's.

\begin{table}[htbp]
\centering
\footnotesize\setlength{\tabcolsep}{4pt}
\begin{adjustbox}{max width=\linewidth}
\begin{tabular}{lrrrrrr}
\toprule
 & \multicolumn{3}{c}{Qwen3.5-9B} & \multicolumn{3}{c}{GLM-5.3 Flash} \\
\cmidrule(lr){2-4}\cmidrule(lr){5-7}
method & $U_{\mathrm{seen}}$ & $U_{\mathrm{held}}$ & $P_{\mathrm{held}}$ & $U_{\mathrm{seen}}$ & $U_{\mathrm{held}}$ & $P_{\mathrm{held}}$ \\
\midrule
\multicolumn{7}{l}{\emph{notes in the harness}} \\
Q, G & 84.7$^{*}$ {\tiny 84.2$^{*}$} & 88.0$^{*}$ {\tiny 87.2$^{*}$} & 15.5 {\tiny 16.5} & 19.2 {\tiny 21.8} & 22.9 {\tiny 22.9} & 94.2 {\tiny 94.2} \\
QE, GE & 71.8$^{*}$ {\tiny 73.8$^{*}$} & 80.6$^{*}$ {\tiny 79.7$^{*}$} & 23.0 {\tiny 24.8} & 4.6 {\tiny 17.7} & \textbf{16.7} {\tiny 18.0} & 97.0 {\tiny 96.5} \\
\midrule
\multicolumn{7}{l}{\emph{notes the teacher writes}} \\
QT & 79.2$^{*}$ {\tiny 84.2$^{*}$} & 85.0$^{*}$ {\tiny 87.2$^{*}$} & 18.5 {\tiny 16.5} & -- & -- & -- \\
QTE & 83.2$^{*}$ {\tiny 84.2$^{*}$} & 88.7$^{*}$ {\tiny 87.2$^{*}$} & 14.5 {\tiny 16.5} & -- & -- & -- \\
QETE & 48.9 {\tiny 73.8$^{*}$} & 83.3$^{*}$ {\tiny 79.7$^{*}$} & 20.5 {\tiny 24.8} & -- & -- & -- \\
\midrule
\multicolumn{7}{l}{\emph{weights}} \\
QDE & 36.0 {\tiny 84.0$^{*}$} & \textbf{60.4} {\tiny 84.7$^{*}$} & \textbf{49.0} {\tiny 19.2} & -- & -- & -- \\
\midrule
\multicolumn{7}{l}{\emph{baselines}} \\
QP, GP & \textbf{22.4} {\tiny 64.9} & 73.0 {\tiny 75.9$^{*}$} & 32.6 {\tiny 28.5} & \textbf{3.7} {\tiny 19.2} & 17.9 {\tiny 18.7} & \textbf{98.0} {\tiny 96.5} \\
QC\textsubscript{0}, GC\textsubscript{0} & 64.4 {\tiny 68.6} & 74.1 {\tiny 71.8} & 31.3 {\tiny 33.5} & 40.1 {\tiny 37.8} & 40.9 {\tiny 41.9} & 77.5 {\tiny 78.5} \\
QC, GC & 42.0 {\tiny 66.5} & 73.6 {\tiny 73.2} & 32.5 {\tiny 32.1} & 21.7 {\tiny 39.0} & 42.3 {\tiny 41.4} & 75.8 {\tiny 78.0} \\
Q\textsubscript{0}, G\textsubscript{0} & 84.0$^{*}$ {\tiny 84.4$^{*}$} & 86.0$^{*}$ {\tiny 88.5$^{*}$} & 18.0 {\tiny 15.0} & 22.4 {\tiny 21.2} & 21.3 {\tiny 24.5} & 96.0 {\tiny 92.5} \\
QE\textsubscript{0}, GE\textsubscript{0} & 75.8$^{*}$ {\tiny 71.8$^{*}$} & 79.5$^{*}$ {\tiny 79.8$^{*}$} & 24.5 {\tiny 25.0} & 18.2 {\tiny 17.1} & 18.1 {\tiny 17.9} & 96.5 {\tiny 96.5} \\
QP\textsubscript{0}, GP\textsubscript{0} & 63.9 {\tiny 65.8} & 75.6$^{*}$ {\tiny 76.1$^{*}$} & 29.0 {\tiny 28.0} & 19.2 {\tiny 19.2} & 19.2 {\tiny 18.2} & 96.0 {\tiny 97.0} \\
\bottomrule
\end{tabular}

\end{adjustbox}
\caption{Every method at one user in absolute terms, on Qwen3.5-9B (left) and on GLM-5.3 Flash (right): $U$ is the budget used on a test in percent, lower is better, and $P_{\mathrm{held}}$ the share of the held-out test solved in percent, each the mean of \testPlays{} plays, the learner's own in normal type and the reset's beside it in small type, each marked with an asterisk where it is above the bare model's, the stateless floor's, as Appendix~\ref{app:heldout} flags. Table~\ref{tab:tests} gives each method's share saved, which is one minus the ratio of the two budgets. Bold marks each column's best.}
\label{tab:tests-absolute}
\end{table}

\paragraph{DAgger at every number of users.}
Table~\ref{tab:tests-users} gives DAgger on Qwen3.5-9B with its floors and wiped learners at each number of users it served, as Table~\ref{tab:tests} gives the methods at one user. At two and four users every user brings 25 tasks of their own, and each method plays the stream and each test once, as does its wiped twin at the same number of users, which it is read against, except that DAgger plays each test \testPlays{} times and its twin the seen test \testPlays{} times. The seen test then serves every user's own tasks, so only that twin played its instances. The held-out test is the same at any number of users, so there the twin's one play joins the \testPlays{} plays of the method's reset at one user, the same wiped system. A method that keeps nothing is its own reset, with no second play of the seen test to be read against at two or four users, and the table gives it at one user alone.

\begin{table}[htbp]
\centering
\footnotesize\setlength{\tabcolsep}{3pt}
\begin{tabular}{lrr@{\hspace{0.4em}}lr@{\hspace{0.4em}}lr@{\hspace{0.4em}}l}
\toprule
method & users & \multicolumn{2}{c}{$L$} & \multicolumn{2}{c}{$S_{\mathrm{seen}}$} & \multicolumn{2}{c}{$S_{\mathrm{held}}$} \\
\midrule
\multicolumn{8}{l}{\emph{Qwen3.5-9B, weights}} \\
QDE & 1 & +0.43 & {\scriptsize [0.33, 0.54]} & +0.57 & {\scriptsize [0.48, 0.67]} & +0.29 & {\scriptsize [0.23, 0.35]} \\
QDE & 2 & +0.47 & {\scriptsize [0.38, 0.56]} & +0.55 & {\scriptsize [0.46, 0.64]} & \textbf{+0.40} & {\scriptsize [0.34, 0.47]} \\
QDE & 4 & \textbf{+0.53} & {\scriptsize [0.46, 0.60]} & +0.51 & {\scriptsize [0.45, 0.57]} & +0.39 & {\scriptsize [0.33, 0.46]} \\
\midrule
\multicolumn{8}{l}{\emph{Qwen3.5-9B, baselines}} \\
QP & 1 & +0.43 & {\scriptsize [0.25, 0.64]} & \textbf{+0.65} & {\scriptsize [0.46, 0.86]} & +0.03 & {\scriptsize [-0.02, 0.09]} \\
QP & 2 & +0.41 & {\scriptsize [0.28, 0.55]} & +0.58 & {\scriptsize [0.43, 0.75]} & +0.03 & {\scriptsize [-0.05, 0.11]} \\
QP & 4 & +0.43 & {\scriptsize [0.33, 0.53]} & +0.50 & {\scriptsize [0.38, 0.62]} & -0.03 & {\scriptsize [-0.11, 0.04]} \\
QC\textsubscript{0} & 1 & 0 & & +0.06 & {\scriptsize [-0.03, 0.15]} & -0.02 & {\scriptsize [-0.10, 0.05]} \\
QC & 1 & +0.31 & {\scriptsize [0.19, 0.45]} & +0.37 & {\scriptsize [0.26, 0.51]} & -0.01 & {\scriptsize [-0.05, 0.03]} \\
QC & 2 & +0.27 & {\scriptsize [0.19, 0.36]} & +0.30 & {\scriptsize [0.19, 0.41]} & 0.00 & {\scriptsize [-0.08, 0.07]} \\
QC & 4 & +0.28 & {\scriptsize [0.22, 0.35]} & +0.31 & {\scriptsize [0.19, 0.43]} & -0.02 & {\scriptsize [-0.09, 0.05]} \\
Q\textsubscript{0} & 1 & 0 & & 0.00 & {\scriptsize [-0.05, 0.07]} & +0.03 & {\scriptsize [-0.03, 0.08]} \\
QE\textsubscript{0} & 1 & 0 & & -0.06 & {\scriptsize [-0.14, 0.01]} & 0.00 & {\scriptsize [-0.05, 0.06]} \\
QP\textsubscript{0} & 1 & 0 & & +0.03 & {\scriptsize [-0.07, 0.12]} & +0.01 & {\scriptsize [-0.06, 0.07]} \\
\bottomrule
\end{tabular}

\caption{DAgger on Qwen3.5-9B, and its baselines, at each number of users they served, read as in Table~\ref{tab:tests}: the share learned on the stream and the share saved on each frozen test, against the method's wiped twin at the same number of users, and on the held-out test against the reset's plays at one user as well. At two and four users DAgger plays each test \testPlays{} times, and its baselines once. Bold marks each column's best among the values whose 95\% interval lies above zero.}
\label{tab:tests-users}
\end{table}

\paragraph{Why four users saved no more than two.}
On the held-out test DAgger saved 0.40 [0.34, 0.47] at two users and 0.39 [0.33, 0.46] at four (Table~\ref{tab:tests-users}), although four users gave it the games of twice as many tasks. We think the brake was the stream's training procedure, not the games. A phase measures its adapter every five steps and stops after six measures without a gain, so a phase that finds no gain within 30 steps stops at any number of users, while the games it trains on grow with the users: 30 steps are about a tenth of a pass over one user's final labels, but a fiftieth over four users'. Rolling back to each phase's best, the runs kept 760 of their 6,060 steps at one user, 1,375 of 7,010 at two and 2,147 of 7,852 at four, about 2.5, 1.9 and 1.4 passes over their final 19,243, 46,744 and 97,580 labels. To test this, we trained the student once more from the base model on all of each run's final labels, shuffled, with no phases, warm starts or rollbacks, at the run's learning rate, batch and held-out games. It measured about eight times a pass and stopped after eight measures without a gain, and its best adapter, reached after about two passes at every number of users, played both tests from the run's own end-of-stream memory, four times at four users and once at one, and at two users the held-out test four times and the seen test once. At one and two users the fit saved what its run did on the held-out test (the fit less the run, $-0.01$ [$-0.08$, 0.06] and $-0.02$ [$-0.06$, 0.03]). At four users, where the run had kept the fewest passes, the fit saved $S_{\mathrm{held}} = 0.46$ [0.40, 0.53], 0.07 [0.02, 0.13] more than its run and 0.06 [0.01, 0.11] more than two users' run, and $S_{\mathrm{seen}} = 0.59$ [0.53, 0.66], 0.08 [0.04, 0.12] more than its run. The held-out loss did not show this: each fit's best was about 3\% below its run's last at every number of users. So four users' games held more than the stream's procedure drew from them. They also held more than two users' games fit the same way, four users' fit saving 0.08 [0.03, 0.13] more on the held-out test, and whether generalization keeps growing with more users, up to all 400 of ARC-AGI-1's training tasks at 16 users, we leave to future work.

\begin{figure}[htbp]
\centering
\includegraphics[width=\linewidth]{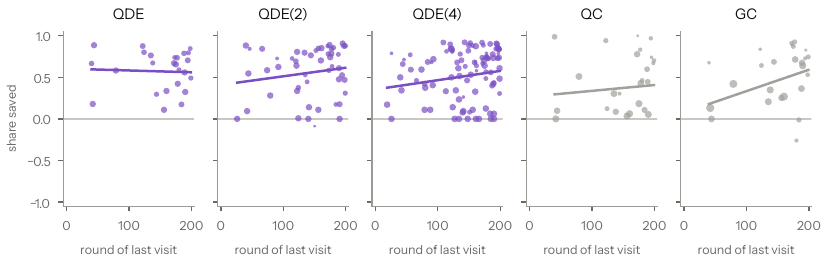}
\caption{Each task's share saved on the seen test against the round of its last visit in the stream, for DAgger at one, two and four users and for the two demo-cache floors. A dot's area is the reset's spend on the task, and each line is the fit weighted by it.}
\label{fig:forgetting}
\end{figure}

\paragraph{Forgetting.}
\label{app:forgetting}
Every task of the stream is visited eight times at any number of users, in a burst whose rounds differ from task to task: at one user the first task's visits end at round 44 and the second's at round 160. A task's last visit therefore says how long it went unpracticed before the seen test, with practice held equal. We read each task's share saved on the seen test against the reset on the same task, so that the task's difficulty cancels, and compare the tasks visited most recently with those left longest (Figure~\ref{fig:forgetting}). No method forgot measurably. DAgger shows no measurable gap at one, two or four users. At four users, with 100 tasks, the half of the tasks visited last saved 0.55 and the half left longest 0.47, a gap of 0.08 [$-0.05$, 0.21], as replaying every game in each phase should give. Over the ten methods whose $S_{\mathrm{seen}}$ is above zero, splitting the tasks into halves and quarters by their last and by their first visit, and taking the slope on each, 4 of 70 readings exclude zero, about what chance alone gives. The largest, GC's, cannot be forgetting, since the demo cache keeps each task's pairs and never drops one: it comes from two tasks of its oldest quarter that a cache of pairs barely helps with, on which QC saved as little (\texttt{scripts/forgetting.py}).

\FloatBarrier  %
\section{Results on GPT-6 Luna}
\label{app:results}

We run three harnesses on GPT-6 Luna, a commercial model, at low, medium and high reasoning effort: prime-agent \citep{karten2026primeagent}, hermes-agent \citep{nous2026hermesagent} and openclaw \citep{openclaw2026}, each with its persistence on, as it ships (H), and with its persistence switched off (H\textsubscript{0}), beside the two floors and the program library. GPT-6 Luna's weights are closed, so this grid has no trained method, and it checks what prime-agent does on the open models against two other harnesses.

\begin{figure}[htbp]
\centering
\includegraphics[width=\linewidth]{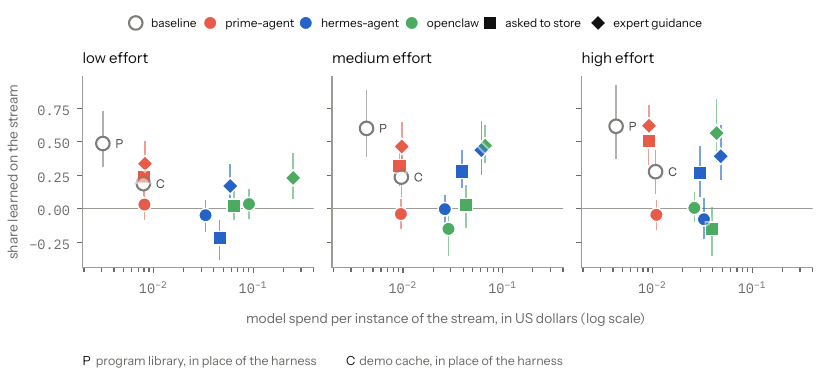}
\caption{The share learned on the stream (Equation~\ref{eq:learned}, higher is better) of every GPT-6 Luna system that keeps something, with its 95\% interval, against what its model calls cost per instance of the stream (log scale), a panel per effort: each harness in its color, as it ships (circle), asked to store (square) and given the expert guidance (diamond), beside the program library (P) and the demo-cache floor (C). Each system is read against itself reset before every instance. The program library learns as much as any system for the least spend.}
\label{fig:luna-cost}
\end{figure}

Figure~\ref{fig:plane-luna} draws the grid at high effort, with each harness also asked to store (H-A) and given the expert guidance (H-E, Appendix~\ref{app:ablations}), and Table~\ref{tab:tests-luna} gives the frozen tests at every effort. Figure~\ref{fig:luna-cost} sets each system's share learned on the stream against what its model calls cost, and Table~\ref{tab:luna-budget} gives the budget each system used on the stream, which the share learned is computed from. The records of every cell come with the paper (the reproducibility statement).

\paragraph{The two models.}
GLM-5.3 Flash and GPT-6 Luna are strong in different ways. On Artificial Analysis's Intelligence Index (v4.3.2), GLM-5.3 Flash scores 42 against 33 for GPT-6 Luna at high effort, with its widest margin on Terminal-Bench, 33 against 5\% \citep{artificialanalysis2026index}. At first sight on ARC the order reverses: from three demonstrations, GLM-5.3 Flash solves 26.8\% of the track's public tasks against 62.5 for GPT-6 Luna at high effort (Table~\ref{tab:heldout}). In the harness the two solve about as often. Once every task has been met, GLM-5.3 Flash in prime-agent solves \sbGlmHzeroSolved{}\% of instances with its notes wiped (Table~\ref{tab:solved}), and GPT-6 Luna at high effort \sbLunaHzeroSolvedMin{} to \sbLunaHzeroSolvedMax{}\% in its three harnesses.

\paragraph{The share learned.}
\label{sec:luna}

\begin{table}[htbp]
\centering
\footnotesize\setlength{\tabcolsep}{3pt}
\begin{adjustbox}{max width=\linewidth}
\begin{tabular}{c|c|l|rr|rr|rr|rr|r}
\toprule
\multicolumn{3}{c|}{} & \multicolumn{2}{c|}{floors} & \multicolumn{2}{c|}{hermes-agent} & \multicolumn{2}{c|}{openclaw} & \multicolumn{2}{c|}{prime-agent} & \multicolumn{1}{c}{program} \\
\cmidrule(lr){4-5}\cmidrule(lr){6-7}\cmidrule(lr){8-9}\cmidrule(lr){10-11}
\multicolumn{3}{l|}{GPT-6 Luna} & stateless & demo-cache & stateless & persistent & stateless & persistent & stateless & persistent & \multicolumn{1}{c}{library} \\
\midrule
\multirow{9}{*}{\rotatebox[origin=c]{90}{effort}} & \multirow{3}{*}{\rotatebox[origin=c]{90}{low}} & \rule{0pt}{2.3ex}workstreams & \makebox[2.15em][r]{57.2}\makebox[2.3em][l]{\hspace{0.45em}\scriptsize$\pm$5.8} & \makebox[2.15em][r]{47.7}\makebox[2.3em][l]{\hspace{0.45em}\scriptsize$\pm$6.4} & \makebox[2.15em][r]{43.8}\makebox[2.3em][l]{\hspace{0.45em}\scriptsize$\pm$4.9} & \makebox[2.15em][r]{45.7}\makebox[2.3em][l]{\hspace{0.45em}\scriptsize$\pm$5.0} & \makebox[2.15em][r]{45.4}\makebox[2.3em][l]{\hspace{0.45em}\scriptsize$\pm$5.0} & \makebox[2.15em][r]{44.0}\makebox[2.3em][l]{\hspace{0.45em}\scriptsize$\pm$5.1} & \makebox[2.15em][r]{47.2}\makebox[2.3em][l]{\hspace{0.45em}\scriptsize$\pm$5.6} & \makebox[2.15em][r]{45.8}\makebox[2.3em][l]{\hspace{0.45em}\scriptsize$\pm$4.8} & \makebox[2.15em][r]{\textbf{21.7}}\makebox[2.3em][l]{\hspace{0.45em}\textbf{\scriptsize$\pm$6.0}} \\
\cline{3-12}
 &  & \rule{0pt}{2.3ex}sequential & \makebox[2.15em][r]{54.7}\makebox[2.3em][l]{\hspace{0.45em}\scriptsize$\pm$5.5} & \makebox[2.15em][r]{47.0}\makebox[2.3em][l]{\hspace{0.45em}\scriptsize$\pm$6.2} & \makebox[2.15em][r]{43.3}\makebox[2.3em][l]{\hspace{0.45em}\scriptsize$\pm$4.7} & \makebox[2.15em][r]{47.5}\makebox[2.3em][l]{\hspace{0.45em}\scriptsize$\pm$5.5} & \makebox[2.15em][r]{43.2}\makebox[2.3em][l]{\hspace{0.45em}\scriptsize$\pm$4.6} & \makebox[2.15em][r]{45.3}\makebox[2.3em][l]{\hspace{0.45em}\scriptsize$\pm$5.3} & \makebox[2.15em][r]{45.2}\makebox[2.3em][l]{\hspace{0.45em}\scriptsize$\pm$4.8} & \makebox[2.15em][r]{49.0}\makebox[2.3em][l]{\hspace{0.45em}\scriptsize$\pm$5.5} & \makebox[2.15em][r]{\textbf{26.1}}\makebox[2.3em][l]{\hspace{0.45em}\textbf{\scriptsize$\pm$7.0}} \\
\cline{3-12}
 &  & \rule{0pt}{2.3ex}random & \makebox[2.15em][r]{54.9}\makebox[2.3em][l]{\hspace{0.45em}\scriptsize$\pm$6.5} & \makebox[2.15em][r]{45.4}\makebox[2.3em][l]{\hspace{0.45em}\scriptsize$\pm$6.9} & \makebox[2.15em][r]{43.0}\makebox[2.3em][l]{\hspace{0.45em}\scriptsize$\pm$5.4} & \makebox[2.15em][r]{46.6}\makebox[2.3em][l]{\hspace{0.45em}\scriptsize$\pm$5.5} & \makebox[2.15em][r]{46.4}\makebox[2.3em][l]{\hspace{0.45em}\scriptsize$\pm$5.3} & \makebox[2.15em][r]{43.4}\makebox[2.3em][l]{\hspace{0.45em}\scriptsize$\pm$5.4} & \makebox[2.15em][r]{46.3}\makebox[2.3em][l]{\hspace{0.45em}\scriptsize$\pm$6.0} & \makebox[2.15em][r]{46.9}\makebox[2.3em][l]{\hspace{0.45em}\scriptsize$\pm$5.9} & \makebox[2.15em][r]{\textbf{23.4}}\makebox[2.3em][l]{\hspace{0.45em}\textbf{\scriptsize$\pm$6.6}} \\
\cline{2-12}
 & \multirow{3}{*}{\rotatebox[origin=c]{90}{medium}} & \rule{0pt}{2.3ex}workstreams & \makebox[2.15em][r]{35.1}\makebox[2.3em][l]{\hspace{0.45em}\scriptsize$\pm$4.5} & \makebox[2.15em][r]{27.4}\makebox[2.3em][l]{\hspace{0.45em}\scriptsize$\pm$5.0} & \makebox[2.15em][r]{28.0}\makebox[2.3em][l]{\hspace{0.45em}\scriptsize$\pm$4.1} & \makebox[2.15em][r]{28.1}\makebox[2.3em][l]{\hspace{0.45em}\scriptsize$\pm$3.7} & \makebox[2.15em][r]{27.8}\makebox[2.3em][l]{\hspace{0.45em}\scriptsize$\pm$3.8} & \makebox[2.15em][r]{31.5}\makebox[2.3em][l]{\hspace{0.45em}\scriptsize$\pm$4.3} & \makebox[2.15em][r]{25.7}\makebox[2.3em][l]{\hspace{0.45em}\scriptsize$\pm$3.2} & \makebox[2.15em][r]{26.6}\makebox[2.3em][l]{\hspace{0.45em}\scriptsize$\pm$3.5} & \makebox[2.15em][r]{\textbf{11.5}}\makebox[2.3em][l]{\hspace{0.45em}\textbf{\scriptsize$\pm$4.0}} \\
\cline{3-12}
 &  & \rule{0pt}{2.3ex}sequential & \makebox[2.15em][r]{34.0}\makebox[2.3em][l]{\hspace{0.45em}\scriptsize$\pm$4.1} & \makebox[2.15em][r]{24.4}\makebox[2.3em][l]{\hspace{0.45em}\scriptsize$\pm$4.7} & \makebox[2.15em][r]{28.1}\makebox[2.3em][l]{\hspace{0.45em}\scriptsize$\pm$4.0} & \makebox[2.15em][r]{25.0}\makebox[2.3em][l]{\hspace{0.45em}\scriptsize$\pm$3.8} & \makebox[2.15em][r]{27.4}\makebox[2.3em][l]{\hspace{0.45em}\scriptsize$\pm$3.7} & \makebox[2.15em][r]{27.9}\makebox[2.3em][l]{\hspace{0.45em}\scriptsize$\pm$3.9} & \makebox[2.15em][r]{26.1}\makebox[2.3em][l]{\hspace{0.45em}\scriptsize$\pm$3.5} & \makebox[2.15em][r]{25.7}\makebox[2.3em][l]{\hspace{0.45em}\scriptsize$\pm$3.9} & \makebox[2.15em][r]{\textbf{9.8}}\makebox[2.3em][l]{\hspace{0.45em}\textbf{\scriptsize$\pm$3.6}} \\
\cline{3-12}
 &  & \rule{0pt}{2.3ex}random & \makebox[2.15em][r]{34.6}\makebox[2.3em][l]{\hspace{0.45em}\scriptsize$\pm$4.9} & \makebox[2.15em][r]{29.6}\makebox[2.3em][l]{\hspace{0.45em}\scriptsize$\pm$5.7} & \makebox[2.15em][r]{29.3}\makebox[2.3em][l]{\hspace{0.45em}\scriptsize$\pm$4.4} & \makebox[2.15em][r]{27.2}\makebox[2.3em][l]{\hspace{0.45em}\scriptsize$\pm$4.3} & \makebox[2.15em][r]{32.7}\makebox[2.3em][l]{\hspace{0.45em}\scriptsize$\pm$3.7} & \makebox[2.15em][r]{27.1}\makebox[2.3em][l]{\hspace{0.45em}\scriptsize$\pm$3.1} & \makebox[2.15em][r]{23.4}\makebox[2.3em][l]{\hspace{0.45em}\scriptsize$\pm$2.8} & \makebox[2.15em][r]{23.6}\makebox[2.3em][l]{\hspace{0.45em}\scriptsize$\pm$3.0} & \makebox[2.15em][r]{\textbf{8.8}}\makebox[2.3em][l]{\hspace{0.45em}\textbf{\scriptsize$\pm$3.9}} \\
\cline{2-12}
 & \multirow{3}{*}{\rotatebox[origin=c]{90}{high}} & \rule{0pt}{2.3ex}workstreams & \makebox[2.15em][r]{29.1}\makebox[2.3em][l]{\hspace{0.45em}\scriptsize$\pm$3.8} & \makebox[2.15em][r]{21.7}\makebox[2.3em][l]{\hspace{0.45em}\scriptsize$\pm$3.6} & \makebox[2.15em][r]{22.8}\makebox[2.3em][l]{\hspace{0.45em}\scriptsize$\pm$2.8} & \makebox[2.15em][r]{24.4}\makebox[2.3em][l]{\hspace{0.45em}\scriptsize$\pm$3.5} & \makebox[2.15em][r]{23.9}\makebox[2.3em][l]{\hspace{0.45em}\scriptsize$\pm$3.0} & \makebox[2.15em][r]{23.7}\makebox[2.3em][l]{\hspace{0.45em}\scriptsize$\pm$3.0} & \makebox[2.15em][r]{23.6}\makebox[2.3em][l]{\hspace{0.45em}\scriptsize$\pm$2.8} & \makebox[2.15em][r]{24.5}\makebox[2.3em][l]{\hspace{0.45em}\scriptsize$\pm$3.1} & \makebox[2.15em][r]{\textbf{8.9}}\makebox[2.3em][l]{\hspace{0.45em}\textbf{\scriptsize$\pm$3.7}} \\
\cline{3-12}
 &  & \rule{0pt}{2.3ex}sequential & \makebox[2.15em][r]{29.6}\makebox[2.3em][l]{\hspace{0.45em}\scriptsize$\pm$4.0} & \makebox[2.15em][r]{17.9}\makebox[2.3em][l]{\hspace{0.45em}\scriptsize$\pm$4.0} & \makebox[2.15em][r]{24.4}\makebox[2.3em][l]{\hspace{0.45em}\scriptsize$\pm$3.7} & \makebox[2.15em][r]{28.8}\makebox[2.3em][l]{\hspace{0.45em}\scriptsize$\pm$4.5} & \makebox[2.15em][r]{23.9}\makebox[2.3em][l]{\hspace{0.45em}\scriptsize$\pm$3.4} & \makebox[2.15em][r]{23.2}\makebox[2.3em][l]{\hspace{0.45em}\scriptsize$\pm$3.4} & \makebox[2.15em][r]{20.7}\makebox[2.3em][l]{\hspace{0.45em}\scriptsize$\pm$2.6} & \makebox[2.15em][r]{20.1}\makebox[2.3em][l]{\hspace{0.45em}\scriptsize$\pm$2.6} & \makebox[2.15em][r]{\textbf{8.3}}\makebox[2.3em][l]{\hspace{0.45em}\textbf{\scriptsize$\pm$3.0}} \\
\cline{3-12}
 &  & \rule{0pt}{2.3ex}random & \makebox[2.15em][r]{27.1}\makebox[2.3em][l]{\hspace{0.45em}\scriptsize$\pm$3.6} & \makebox[2.15em][r]{19.6}\makebox[2.3em][l]{\hspace{0.45em}\scriptsize$\pm$4.1} & \makebox[2.15em][r]{23.1}\makebox[2.3em][l]{\hspace{0.45em}\scriptsize$\pm$3.3} & \makebox[2.15em][r]{22.1}\makebox[2.3em][l]{\hspace{0.45em}\scriptsize$\pm$3.4} & \makebox[2.15em][r]{23.4}\makebox[2.3em][l]{\hspace{0.45em}\scriptsize$\pm$3.2} & \makebox[2.15em][r]{23.9}\makebox[2.3em][l]{\hspace{0.45em}\scriptsize$\pm$2.9} & \makebox[2.15em][r]{20.0}\makebox[2.3em][l]{\hspace{0.45em}\scriptsize$\pm$2.6} & \makebox[2.15em][r]{20.6}\makebox[2.3em][l]{\hspace{0.45em}\scriptsize$\pm$2.5} & \makebox[2.15em][r]{\textbf{6.2}}\makebox[2.3em][l]{\hspace{0.45em}\textbf{\scriptsize$\pm$2.7}} \\
\bottomrule
\end{tabular}

\end{adjustbox}
\caption{GPT-6 Luna on the ARC-AGI-1 stream: the budget used $U$ (percent, lower is better) by each system at each reasoning effort, the measure that $L$ is computed from. Workstreams is the benchmark's stream, and sequential and random are the same visits one task at a time and fully shuffled (Appendix~\ref{app:orders}). A harness's persistent column is method H and its stateless column H\textsubscript{0}, and the stateless floor is the demo-cache floor's reset. Small type is the standard error over tasks, and the best in each row is bold, even within error bars.}
\label{tab:luna-budget}
\end{table}

On GPT-6 Luna, no harness as shipped saves a measurable share of the learnable budget at any reasoning effort (Figure~\ref{fig:luna-cost}). Against their own methods H\textsubscript{0}, the persistent methods lie between $-0.15$ and 0.04, and every 95\% interval includes zero, while the program library saves 0.49 [0.31, 0.73] of its learnable budget at low effort, 0.60 [0.38, 0.88] at medium and 0.62 [0.37, 0.93] at high, and the demo-cache floor, which only keeps the pairs it bought, saves 0.19 [0.10, 0.29], 0.24 [0.09, 0.40] and 0.28 [0.11, 0.44] of its own. The learnable budget of the model alone falls as the effort rises, from 50.8 points at low effort to 26.5 at high, because a stronger model pays less on every instance, returns included. A model that leaves little of it has outgrown the track (Section~\ref{sec:metrics}).

The state that each harness stored shows why the harnesses saved nothing (Appendix~\ref{app:persist}). Left to decide, none stored what answered a returning task. The three are built for lessons that transfer across unrelated work, none ties its store to the task id, and each breaks at a different step between an instance and its return. prime-agent kept nothing beyond a session in any of its three persistent cells, so the digest of its store that it shows the model was empty on every return. hermes-agent always shows the model what it stored, but it stores general advice on how to approach a task, with files named for a task only at low effort, where the re-exposures on which one came back cost more than the rest. openclaw seldom searches its history, and its consolidation never writes a memory, so no earlier correct answer came back on any re-exposure its replay log holds. Told what to keep, they keep it: with the expert guidance, the program library's procedure, added to their rules (method \mbox{H-E}, Appendix~\ref{app:ablations}), every harness saves a measurable share of its learnable budget at every effort, 0.17 to 0.62 against the same harness given the guidance and wiped (method \mbox{H-E\textsubscript{0}}), and prime-agent's matches the program library within noise at medium and high effort, and openclaw's at high. The harnesses on GPT-6 Luna therefore memorize when they are told what to keep, as GLM-5.3 Flash's harness does, while Qwen3.5-9B's, told the same, saved nothing measurable (Section~\ref{sec:results}).

\paragraph{The model and its presets.}
\label{sec:models}
We choose the models from the ARC Prize leaderboard at the time of submission. A preset (a model at one reasoning effort) qualifies when its verified semi-private zero-shot score lies between about 15 and 60\%, the band in which we expect a harness to have room to learn. One model family, GPT-6 Luna, covers the track from within the band at the lowest cost on the board, and we use its reasoning effort as the ability axis. We run low, medium and high effort on ARC-AGI-1, the last above the band. The board is measured on the evaluation set, but the ARC-AGI-1 stream serves easier, generated training instances. For this reason, we also check each preset against the stateless floor's first-exposure solve rate on the stream itself. On ARC-AGI-1, pooled with the side experiment's two orders, the stateless floor solved 33\% of its 75 first exposures at first answer at low effort, 61 at medium and 67 at high, against board scores of 37.7, 61.0 and 70.3\%. Low and medium effort leave room to learn. On every preset we run, the model alone leaves a learnable budget of 26.5 to 50.8 points. On ARC-AGI-1, every method runs at every effort, which lets us compare the harnesses on identical weights at each effort.

\paragraph{The charge these runs were played under.}
\label{sec:charge}
Every GPT-6 Luna cell, the side experiment's on the order among them, and its frozen tests, was played under rules that state the charge of Equation~\ref{eq:score} (Appendix~\ref{app:prompts}). The rules were first written for an earlier charge, under which an instance that ended unsolved cost the attempt cap, 8, plus its demonstrations. Under the current charge a learner that will fail anyway loses nothing by buying every demonstration before its last attempt, so the rules it reads can change its play. We therefore played GPT-6 Luna's cells again under the current rules rather than charge their records again, and no cell played under the earlier rules enters a number we report.

\paragraph{The frozen tests.}
After the stream, every system plays the two tests of Section~\ref{sec:tests} from its final state, and the reset plays them from an empty state, apart from it (Table~\ref{tab:tests-luna}). A system that keeps nothing between instances, a floor or a wiped method, reads half its plays against the other half, so its spread, $-0.06$ to 0.13 on the seen test and $-0.17$ to 0.07 on the held-out test, is the noise of playing a test. Judged with the other learners of its effort (Appendix~\ref{app:bench}), every method that kept a program per task saved more than chance on the seen test, as did the demo-cache floor and prime-agent asked to store, while the harnesses as shipped did not. On the held-out test, none of the 30 readings of the methods that keep something saved more than chance once each effort's eleven learners are judged together, and four cost more, three of them hermes-agent's. The tests detect memorization, and no positive held-out saving survives the correction.

\begin{table}[htbp]
\centering
\footnotesize\setlength{\tabcolsep}{3pt}
\begin{adjustbox}{max width=\linewidth}
\begin{tabular}{lr@{\hspace{0.4em}}lr@{\hspace{0.4em}}lr@{\hspace{0.4em}}lr@{\hspace{0.4em}}lr@{\hspace{0.4em}}lr@{\hspace{0.4em}}l}
\toprule
method & \multicolumn{6}{c}{share saved on the seen test} & \multicolumn{6}{c}{share saved on the held-out test} \\
\cmidrule(lr){2-7}\cmidrule(lr){8-13}
 GPT-6 Luna, effort & \multicolumn{2}{c}{low} & \multicolumn{2}{c}{medium} & \multicolumn{2}{c}{high} & \multicolumn{2}{c}{low} & \multicolumn{2}{c}{medium} & \multicolumn{2}{c}{high} \\
\midrule
\multicolumn{13}{l}{\emph{floors}} \\
stateless & 0.00 & {\scriptsize [-0.07, +0.06]} & +0.12 & {\scriptsize [+0.01, +0.24]} & -0.04 & {\scriptsize [-0.14, +0.05]} & +0.02 & {\scriptsize [-0.03, +0.07]} & -0.04 & {\scriptsize [-0.15, +0.08]} & -0.12 & {\scriptsize [-0.23, -0.02]} \\
demo-cache & +0.21 & {\scriptsize [+0.08, +0.36]} & +0.34 & {\scriptsize [+0.22, +0.48]} & +0.51 & {\scriptsize [+0.38, +0.67]} & +0.03 & {\scriptsize [0.00, +0.07]} & 0.00 & {\scriptsize [-0.07, +0.07]} & -0.01 & {\scriptsize [-0.10, +0.07]} \\
\midrule
\multicolumn{13}{l}{\emph{program library}} \\
PL\textsubscript{0} & +0.02 & {\scriptsize [-0.06, +0.09]} & -0.04 & {\scriptsize [-0.14, +0.05]} & +0.05 & {\scriptsize [-0.05, +0.13]} & +0.07 & {\scriptsize [-0.03, +0.15]} & +0.05 & {\scriptsize [-0.08, +0.16]} & -0.02 & {\scriptsize [-0.11, +0.05]} \\
PL & \textbf{+0.54} & {\scriptsize [+0.35, +0.78]} & \textbf{+0.77} & {\scriptsize [+0.54, +0.97]} & \textbf{+0.73} & {\scriptsize [+0.47, +0.98]} & -0.06 & {\scriptsize [-0.12, 0.00]} & -0.06 & {\scriptsize [-0.15, +0.04]} & -0.02 & {\scriptsize [-0.08, +0.05]} \\
\midrule
\multicolumn{13}{l}{\emph{hermes-agent}} \\
H\textsubscript{0} & +0.04 & {\scriptsize [-0.05, +0.14]} & +0.01 & {\scriptsize [-0.07, +0.09]} & +0.01 & {\scriptsize [-0.12, +0.12]} & 0.00 & {\scriptsize [-0.08, +0.08]} & -0.17 & {\scriptsize [-0.35, -0.03]} & -0.07 & {\scriptsize [-0.20, +0.06]} \\
H-E\textsubscript{0} & +0.03 & {\scriptsize [-0.08, +0.13]} & -0.03 & {\scriptsize [-0.12, +0.06]} & +0.03 & {\scriptsize [-0.09, +0.13]} & +0.03 & {\scriptsize [-0.06, +0.12]} & +0.02 & {\scriptsize [-0.10, +0.13]} & 0.00 & {\scriptsize [-0.16, +0.13]} \\
H & +0.03 & {\scriptsize [-0.09, +0.15]} & -0.07 & {\scriptsize [-0.22, +0.05]} & +0.03 & {\scriptsize [-0.05, +0.10]} & -0.02 & {\scriptsize [-0.09, +0.04]} & -0.21 & {\scriptsize [-0.32, -0.10]} & -0.01 & {\scriptsize [-0.09, +0.06]} \\
H-A & -0.04 & {\scriptsize [-0.22, +0.13]} & +0.18 & {\scriptsize [+0.05, +0.35]} & +0.18 & {\scriptsize [-0.01, +0.39]} & -0.25 & {\scriptsize [-0.34, -0.17]} & -0.09 & {\scriptsize [-0.18, 0.00]} & -0.01 & {\scriptsize [-0.11, +0.07]} \\
H-E & +0.30 & {\scriptsize [+0.14, +0.49]} & +0.52 & {\scriptsize [+0.36, +0.71]} & +0.62 & {\scriptsize [+0.45, +0.81]} & -0.09 & {\scriptsize [-0.16, -0.03]} & \textbf{+0.11} & {\scriptsize [+0.01, +0.19]} & \textbf{+0.13} & {\scriptsize [+0.04, +0.21]} \\
\midrule
\multicolumn{13}{l}{\emph{openclaw}} \\
H\textsubscript{0} & +0.01 & {\scriptsize [-0.08, +0.09]} & -0.06 & {\scriptsize [-0.15, +0.01]} & 0.00 & {\scriptsize [-0.10, +0.10]} & -0.04 & {\scriptsize [-0.12, +0.04]} & -0.11 & {\scriptsize [-0.27, +0.02]} & +0.02 & {\scriptsize [-0.11, +0.14]} \\
H-E\textsubscript{0} & +0.06 & {\scriptsize [-0.02, +0.14]} & +0.01 & {\scriptsize [-0.12, +0.13]} & 0.00 & {\scriptsize [-0.08, +0.08]} & -0.02 & {\scriptsize [-0.11, +0.07]} & -0.02 & {\scriptsize [-0.14, +0.08]} & +0.06 & {\scriptsize [-0.05, +0.15]} \\
H & -0.04 & {\scriptsize [-0.13, +0.03]} & +0.02 & {\scriptsize [-0.09, +0.12]} & -0.04 & {\scriptsize [-0.12, +0.03]} & +0.01 & {\scriptsize [-0.05, +0.07]} & -0.06 & {\scriptsize [-0.17, +0.04]} & -0.05 & {\scriptsize [-0.13, +0.03]} \\
H-A & -0.11 & {\scriptsize [-0.21, -0.02]} & +0.04 & {\scriptsize [-0.08, +0.18]} & +0.07 & {\scriptsize [-0.02, +0.17]} & -0.07 & {\scriptsize [-0.14, 0.00]} & 0.00 & {\scriptsize [-0.10, +0.10]} & -0.14 & {\scriptsize [-0.23, -0.04]} \\
H-E & +0.28 & {\scriptsize [+0.12, +0.48]} & +0.58 & {\scriptsize [+0.38, +0.78]} & +0.64 & {\scriptsize [+0.44, +0.82]} & -0.06 & {\scriptsize [-0.14, 0.00]} & +0.02 & {\scriptsize [-0.07, +0.11]} & -0.02 & {\scriptsize [-0.11, +0.06]} \\
\midrule
\multicolumn{13}{l}{\emph{prime-agent}} \\
H\textsubscript{0} & -0.05 & {\scriptsize [-0.17, +0.05]} & +0.02 & {\scriptsize [-0.11, +0.14]} & +0.02 & {\scriptsize [-0.14, +0.14]} & +0.03 & {\scriptsize [-0.07, +0.12]} & -0.05 & {\scriptsize [-0.22, +0.09]} & -0.08 & {\scriptsize [-0.21, +0.05]} \\
H-E\textsubscript{0} & -0.03 & {\scriptsize [-0.13, +0.06]} & +0.13 & {\scriptsize [0.00, +0.24]} & +0.07 & {\scriptsize [-0.01, +0.15]} & +0.04 & {\scriptsize [-0.04, +0.12]} & -0.01 & {\scriptsize [-0.15, +0.11]} & -0.05 & {\scriptsize [-0.23, +0.10]} \\
H & -0.02 & {\scriptsize [-0.09, +0.05]} & +0.06 & {\scriptsize [-0.03, +0.14]} & -0.05 & {\scriptsize [-0.17, +0.04]} & 0.00 & {\scriptsize [-0.05, +0.06]} & -0.03 & {\scriptsize [-0.15, +0.07]} & +0.06 & {\scriptsize [-0.05, +0.14]} \\
H-A & +0.19 & {\scriptsize [+0.08, +0.35]} & +0.37 & {\scriptsize [+0.27, +0.49]} & +0.42 & {\scriptsize [+0.29, +0.55]} & +0.01 & {\scriptsize [-0.05, +0.07]} & +0.05 & {\scriptsize [-0.06, +0.15]} & +0.12 & {\scriptsize [+0.03, +0.21]} \\
H-E & +0.35 & {\scriptsize [+0.17, +0.58]} & +0.57 & {\scriptsize [+0.34, +0.80]} & +0.72 & {\scriptsize [+0.63, +0.82]} & +0.02 & {\scriptsize [-0.05, +0.08]} & 0.00 & {\scriptsize [-0.10, +0.08]} & +0.05 & {\scriptsize [-0.03, +0.12]} \\
\bottomrule
\end{tabular}

\end{adjustbox}
\caption{GPT-6 Luna's frozen tests: the share of the reset's spend on each test that a system no longer pays, against itself reset (Equation~\ref{eq:transfer}, higher is better), with its 95\% interval over task resamples in small type. A harness is read against its method H\textsubscript{0}, and \mbox{H-E} against \mbox{H-E\textsubscript{0}}; the demo-cache floor against the stateless floor, and the program library against PL\textsubscript{0}. A floor or a wiped method reads half its plays against the other half, so these methods show the noise of playing a test. Bold marks each column's best among the values whose 95\% interval lies above zero.}
\label{tab:tests-luna}
\end{table}

\section{The private set}
\label{app:heldout}

Every task of the track has a public answer, and every task of its stream a public solver, so the statement of what a learner may use, which comes with the code, forbids the lookup of Appendix~\ref{app:bench} but cannot make it impossible. A new seed changes the instances and not the rules. The private set holds new rules that only we have seen, on which we rerun a learner's code to verify a result. We keep it as ARC-AGI keeps its private evaluation set \citep{chollet2025arcagi2}: the code comes to the tasks, and the tasks never go to a submitter's machine, which could keep whatever it was shown. We name no rule here and show no instance, since anything shown of a rule publishes it. The construction below took the first track's set from 98 candidates to the 50 kept; the second and third tracks' sets, built by the same method, follow it.

\paragraph{The rules.} Six agents on Claude Opus 5.5 wrote the 98 candidates, one agent for each family of rules: objects, geometry, lines and paths, patterns and sequences, panels and logic, and physics and topology, with 16 or 17 rules each. A rule is a module with a generator, which draws an instance from a seed, and a solver, which computes the output from the input alone. The benchmark serves the rules as one more task source, with a fresh pair for every test input and every demonstration, so any learner that runs on the public track runs on the private set unchanged. Its held-out rules are generated as well, and a request there returns a fresh pair up to the cap of eight, where the public held-out test serves each task's own two to seven pairs. The agents were asked for one idea that a person could infer from three examples, with every color role, count or direction that the answer depends on visible in the grid, one answer for every instance, varied sizes, positions and colors, and no public task or plain ARC trope, such as gravity or symmetry completion, without a twist that changes the rule.

\paragraph{Filters and screens.} A candidate passes the mechanical filters when, over 300 draws, its generator and solver always finish, every grid is valid, the solver reproduces every generated output, its inputs and outputs vary, and the output is not always the input under one fixed rotation or reflection; all 98 passed, and the thresholds are in the set's code. Two screens then look for a rule that already exists. Re-ARC's 400 verifiers, the solvers of the ARC-AGI-1 training tasks, reproduced none of the candidates, and each candidate's own solver, given every pair of all 1{,}920 public ARC-AGI-1 and ARC-AGI-2 tasks, dropped six candidates, each of which matched one to three public tasks. No candidate contained another.

\paragraph{Calibration.} We measure how often a model solves a rule at first sight. An item is one fresh instance with three demonstrations, drawn as the demo-cache floor first meets the task, in a fixed prompt, and the model gives one answer, which must match exactly, with no feedback and no second attempt. Claude Opus 5.5 at max effort answered four instances of each of the 92 remaining candidates and solved 360 of the 368. GPT-6 Luna answered eight of each at low, medium and high effort. Both answered the ARC-AGI-1 track's 50 public tasks in the same way: fresh instances of its 25 stream tasks from the Re-ARC generators, as many as for a rule, and each test pair of its 25 held-out evaluation tasks with up to three of the task's own pairs. A candidate that Opus solved fewer than three times in four was set aside, which two were, and one that Luna solved every time at every effort would have been too easy, which none was. The calibration cost about 60 US dollars on Opus and 6 on Luna. These rates are not the floors' first-exposure rates of Appendix~\ref{sec:models}, since a floor chooses its own demonstration requests and often answers before it asks for any.

\paragraph{Selection.} From the 90 eligible candidates, we removed one at a time, each time the one whose removal brought the rest nearest the public tasks' first-sight rates over the four presets of Claude Opus 5.5 and GPT-6 Luna, keeping the families near equal shares, until 50 were left, 8 or 9 per family, and split them into 25 for the stream and 25 held out, each family dealt to both halves (Table~\ref{tab:heldout}). The 92 calibrated candidates were easier for GPT-6 Luna than the public tasks, by 3.7 to 5.6 points, and the selection removes the difference. The other 40 eligible candidates are a reserve, for a kept rule that has to be retired. The set is frozen, since a run on it is comparable with another only while its rules stay as they are. It serves one user, since every further user brings 25 rules of their own, and we will add rules to verify learners in the collective regimes.

\paragraph{Two models the selection never saw.} After the set was frozen, GLM-5.3 Flash and Qwen3.5-9B at medium effort answered eight instances of each of the 92 candidates and the ARC-AGI-1 track's public tasks in the same way, as a check on models that the selection did not match. GLM-5.3 Flash solves 26.5\% of the kept rules and 26.8\% of the public tasks, though its halves part in opposite directions, the streamed rules 6.5 points under the public stream and the held-out rules 6.0 points over the public held-out tasks. Qwen3.5-9B solves 5.5\% of the kept rules and 8.5\% of the public tasks, its streamed rules about as often as the public stream and its held-out rules 6.5 points less often than the public held-out tasks. The public held-out rates rest on 26 answers, one to each test pair, of which GLM-5.3 Flash solved 6 and Qwen3.5-9B 3, so one answer moves them by up to four points. More than nine in ten of Qwen3.5-9B's replies, on the rules and on the public tasks alike, reached the cap of 32{,}000 tokens before an answer, so its rates mostly count replies cut off. GLM-5.3 Flash's reached it on 22\% of its replies on the rules and 25\% on the public tasks, almost none of them solved, so its rates also count some replies cut off, about equally on both sides. The two cost about 9 US dollars together.

\begin{table}[htbp]
\centering
\small\setlength{\tabcolsep}{5pt}
\begin{tabular}{lrrrrrrr}
\toprule
 & \multicolumn{2}{c}{stream} & \multicolumn{2}{c}{held out} & \multicolumn{2}{c}{all} & \\
\cmidrule(lr){2-3}\cmidrule(lr){4-5}\cmidrule(lr){6-7}
model, effort & private & public & private & public & private & public & calibrated \\
\midrule
GPT-6 Luna, low & 31.5 & 31.5 & 21.0 & 20.0 & 26.3 & 25.8 & 31.4 \\
GPT-6 Luna, medium & 54.0 & 54.0 & 56.0 & 56.0 & 55.0 & 55.0 & 58.7 \\
GPT-6 Luna, high & 59.0 & 59.0 & 66.5 & 66.0 & 62.8 & 62.5 & 67.7 \\
Claude Opus 5.5, max & 97.0 & 97.0 & 100.0 & 100.0 & 98.5 & 98.5 & 97.8 \\
GLM-5.3 Flash, medium & 23.0 & 29.5 & 30.0 & 24.0 & 26.5 & 26.8 & 29.2 \\
Qwen3.5-9B, medium & 5.5 & 5.0 & 5.5 & 12.0 & 5.5 & 8.5 & 4.9 \\
\bottomrule
\end{tabular}

\caption{First-sight solve rates, in percent, on the private set and on the ARC-AGI-1 track's public tasks, measured the same way: three demonstrations, one answer and an exact match. Private columns are the 25 streamed rules, the 25 held-out rules and the 50 kept; public columns are the track's 25 stream tasks, its 25 held-out ARC-AGI-1 evaluation tasks and all 50. Calibrated is all 92 candidates that passed the filters and screens. Each rate is a mean over rules or tasks, of eight instances per rule or stream task for GPT-6 Luna, GLM-5.3 Flash and Qwen3.5-9B and four for Claude Opus 5.5. The selection matched the first four rows; the last two were calibrated after it.}
\label{tab:heldout}
\end{table}

\paragraph{The second track's set.} Seven agents on Claude Opus 5.5 wrote 116 candidate rules for the season, one agent for each of seven families new to the track: legends and keys, context, compositions, programs and procedures, relations and graphs, assembly and shape matching, and counting and reading out. Each was asked for ARC-AGI-2's character: symbols whose meaning the grid sets, rules that compose or apply by context, and grids up to 30 by 30. All 116 passed the filters and both screens, against the 1{,}920 public tasks and fresh draws of the track's ARC-GEN generators, and none contains another or one of the first track's rules. Claude Opus 5.5 at max effort solved 450 of its 464 answers, and the two rules it solved only twice in four were set aside. The selection matched Kimi K3.1 Pro at medium effort, the track's model, with 24 answers per rule, and kept 50 rules, 7 per family and 8 of counting, 25 for the stream and 25 held out, with 64 in reserve. Kimi solves 32\% of the kept rules and of the public tasks, the stream 45 against 45 and the held-out rules 18 against 18, and Opus 95 against 91. Two models the selection never saw agree: Kimi-K3 at medium effort solves 38\% against 40 and GLM-5.3 Flash 12 against 12, though half or more of GLM-5.3 Flash's replies reach the cap. These readings are of OpenRouter's routing on 7 November 2026, which served Kimi from many providers; pinned to Moonshot AI's own endpoint, which served all our Kimi runs, Kimi solves 30\% of the kept rules against 27 of the public tasks, the stream 46 against 41 and the held-out rules 15 against 14. On eight fresh instances of each kept rule, which the selection never read, Kimi still solves 32\%, but its halves part less, the stream 38 and the held-out rules 26, so some of the spread the selection matched was chance, and the held-out rules may read easier than the public held-out tasks. The calibration cost about 660 US dollars.

\paragraph{The third track's set.} ARC-AGI-3 keeps no hidden games that anyone outside ARC Prize can use, so agents on Claude Opus 5.5 wrote 28 candidate games for the season on ARC Prize's engine, each with sprite types and mechanics of its own, a first level, a generator and a solver, and each designed away from the 25 public games. Several are versions of well-known puzzles, and none is an ARC-AGI-3 game. All 28 pass the public track's checks at 50 seeds, draw the same levels in another process, and draw without failing at 500 of the seeds a run derives. A first-sight reading plays fresh levels as Appendix~\ref{app:tracks} does, the model wiped before each level and free to ask for three demonstrations. Claude Opus 5.5 at max effort solved all 112 levels it played, and all 100 of the public games'. Kimi K3.1 Pro at medium effort, the track's model, first solved fewer levels of the candidates than of the public games, 19 against 28\% on the first 22, and the eight of which it solved none, puzzles that need planning ahead, were made easier with their rules unchanged, with smaller boards, fewer pieces or shorter first levels, before the selection. The selection kept 25 games, 15 for the stream and 10 held out, with 3 in reserve, on which Kimi solves 28\% of the levels against 28 on the public games, the stream 31 against 31 and the held-out games 24 against 24, on OpenRouter's routing of 7 November. That match is selected on the readings that measure it. Pinned to Moonshot AI's own endpoint, which served all our Kimi runs and serves every verified run, the same levels read the kept games somewhat harder, 27\% against 33, the stream 29 against 35 and the held-out games 24 against 29, a gap of about one standard error over games. GLM-5.3 Flash at medium effort, which the selection never saw, solves 43\% against 40 on the routing, the held-out games 50 against 42, within the noise of 40 levels. The kept games are mostly played by clicking, 15 by clicks, 8 by keys and 2 by both, where the public games are 8, 4 and 13. A learner no better than its reset reads a held-out share saved within about $[-0.14, 0.12]$ on the 10 held-out games, against $[-0.11, 0.10]$ on the public track's. No person has played the games. The calibration cost about 580 US dollars.

\paragraph{The verified run.} A submitter opens an issue on the benchmark's public repository with a public commit, the learner, the published models it uses and a budget for their calls, which it pays ahead. We run the learner, its reset and each of its models bare on the private set at a seed drawn for the run and not disclosed, inside the sandbox of Appendix~\ref{app:harness} on a machine of its own that has no public network and holds neither the rules nor the key, and the learner reaches nothing but the tracking proxy, which makes every model call on a key of ours, each model served by one endpoint for the whole run, capped at that budget. The issue then receives each test's share saved with its 95\% interval, the share learned, the budgets and shares solved of the learner and the reset, and what the model calls cost, with the flags below. Only these numbers come back. The records, and anything the learner kept, stay with us, since they show the rules.

\paragraph{Checks against gaming.} A submission names every model its learner uses, each being a published model that must start as released. Each share is read beside its budgets (Table~\ref{tab:tests-absolute}), and a test on which the reset plays below any of the learner's models played bare at no greater cost, or below the cheapest of them at any cost, is flagged, so that a deliberately weakened reset cannot inflate the learner's relative gain, and so is one on which the learner plays no better than such a model, so that switching to a stronger model cannot read as learning, and one on which the learner sends a larger share of its calls than its reset to any of its models but the cheapest, so that switching to a stronger model for only some calls cannot either.

\paragraph{What remains open.} No person has tried the rules. Claude Opus 5.5 at max effort, solving every kept rule at least three times in four from three demonstrations, stands in for the human testers who solved each ARC-AGI-2 task \citep{chollet2025arcagi2}. The same model wrote the rules, and since it solves 98.5\% of both the kept rules and the public tasks, at the ceiling, the calibration cannot show whether rules written by a model are easier for that model. An Opus learner is read against itself reset on the same instances, which shares any such advantage. The match of Table~\ref{tab:heldout} is at first sight and says nothing of how much a rule can be learned on its return, so the private set's learnable budget need not equal the public stream's, and a verified run measures it with the reset. We read a share learned on the private set below the one on the public stream, beyond both intervals, as a sign of tuning to the public stream. A learner chooses its own scores, so the numbers a verified run publishes carry a little of what it was served. They are rounded, a submitter's runs in a season are capped, and the set can be retired for new rules, but only rules drawn fresh for every run would remove this. Learners that train their weights are verified on machines we rent, since a submitter's own machine would keep the instances it trained on.

\section{The second and third tracks}
\label{app:tracks}

Both tracks keep the first track's protocol. Tasks return with fresh instances after gaps while new tasks keep arriving, the learner may ask for a demonstration or act, a failure costs the whole budget, and after the stream the learner is frozen and plays a seen test and a held-out test, each read against the learner wiped before every instance. Each is its edition's configuration in the released code (\texttt{v0.6.0}), at seed 0. Since users bring tasks of their own, the second track's pool serves at most four users and the third's one. A verified run on either track plays a private set of its own, 50 new rules on ARC-AGI-2 and 25 new games on ARC-AGI-3 (Appendix~\ref{app:heldout}).

\paragraph{ARC-AGI-2.} Of ARC-AGI-2's 1,000 public training tasks, 767 are ARC-AGI-1 tasks, as are 6 of its 120 evaluation tasks \citep{chollet2025arcagi2}, so the track draws only on tasks new to the second edition. ARC-GEN \citep{moffitt2025arcgen} has generators for 172 of the 233 new training tasks, and a task enters the pool when its generator passes a screen over 1,000 draws, the private set's mechanical filters (Appendix~\ref{app:heldout}) and two checks of fidelity. ARC-TGI \citep{lehmann2026arctgi} also offers generators for part of ARC-AGI-2, but the track uses ARC-GEN alone. ARC-GEN's own validation must reproduce every published pair, and every draw must keep what all the published pairs share, such as the output's shape, its scale and colors drawn from the input; the thresholds are in the track's code. Of the 172, 117 pass. The track streams 25 of them, eight fresh instances each, in the first track's order, and its held-out test is 100 of the 114 evaluation tasks new to ARC-AGI-2, each served with its own pairs. A demonstration is a fresh generated pair, as on the first track. The stream's tasks are training tasks, which, unlike the evaluation tasks, were not calibrated on people \citep{chollet2025arcagi2}, so the stream may be the easier half.

\paragraph{ARC-AGI-3.} Each of ARC-AGI-3's 25 public games is a 64 by 64 grid that responds to seven actions and a reset over a fixed sequence of levels \citep{arcprize2026arcagi3}. In 22 of the games the later levels add sprite types, and in several they change a rule's parameters, so a game's levels are not instances of one task. A task of the track is therefore a game's rules at its first level, and an instance a fresh level of those rules, with the same sprite types and data in a new layout. A generator we wrote for each game draws the level, and the game's solver, a search over the game's state, solves it and checks the answer by replaying it in ARC Prize's engine. A game enters the track only if its generator passes a gate over 50 seeds, which checks that every level is drawn quickly, solved and won on replay, close in length to the game's own first level, distinct from the others and from the public first level, and made of the first level's sprite types alone, with the thresholds in the track's code. All 25 games pass, and they are split once, at seed 0, into 15 that the stream serves and 10 held out. The track serves eight fresh levels of each stream game in the first track's order, its seen test four fresh levels of each, and its held-out test four of each held-out game.

An instance's baseline is the human count of actions for the game's first level, scaled by the instance's solution length over the first level's. Its budget of actions is five times its baseline, at most 1,000, and every action and every reset of the level counts toward it. A solved level costs its actions and resets over its baseline, at most 5, and one more for each demonstration, at most three; an unsolved level costs the whole budget of 8, so that failing is never cheaper than trying, as on the first track. A demonstration is the solver's play of another fresh level of the same game. Since no instance is a level anyone has played before, a memory of the public levels, with which a harness has solved that set \citep{bedi2026learningmachines}, as have three concurrent ones \citep{wu2026kepler,zhao2026memento3,anonymous2026comate}, does not help. Compiled Worlds \citep{anonymous2026compiledworlds}, also concurrent, edits the public games' programs into variants, resampled layouts and mutated rules among them, and prices what each change adds to a harness's cost on the original level. A game's later levels, with their new rules, are left for a mode of the benchmark in which the rules drift.

\paragraph{First sight.} Table~\ref{tab:first-sight} gives how often three models solve each track's instances at first sight. On ARC-AGI-2 a model sees three demonstrations and answers once, as the private set's calibration does (Appendix~\ref{app:heldout}). On ARC-AGI-3 it plays the held-out test, four fresh levels of each held-out game, wiped before each and free to ask for up to three demonstrations per level. Qwen3.5-9B's replies reached their cap before an answer on 61 of its 100 ARC-AGI-2 instances, and a provider's error ended 7 more, so its rate mostly counts replies cut off, as on the first track. On ARC-AGI-3, where a level takes tens to hundreds of moves, a quarter of its replies reasoned to the cap, some for over 20 minutes, and after two and a half hours it had made 30 of its first level's 110 moves, so we report no result there.

\paragraph{Kimi K3.1 Pro on ARC-AGI-2.} Kimi K3.1 Pro has no published ARC-AGI-2 score, but its 15\% or so on our held-out test, which serves 100 of ARC-AGI-2's own evaluation tasks with their own pairs, is in line with ARC Prize's results for Kimi K3 at low reasoning effort on the same tasks, 16 solved against 45 at high effort and 55 at max \citep{arcprize2026leaderboard}. Kimi offers only low, high and max effort, and at the medium we requested through OpenRouter its first-sight replies ran about 1,000 tokens at the median, a fifth of Kimi-K3's at the same setting, with none reaching the cap; played again on 50 of those instances on Moonshot AI's own endpoint, which served all our Kimi runs, medium solved 19 and 24 at replies as long as low's, which solved 23, against 40 and 44 at high, so our runs, which request medium, play Kimi at what is in effect its low effort. Its higher rates elsewhere come from easier tasks: ARC-AGI-2's training tasks in the stream and the seen test, which were not calibrated on people, and each ARC-AGI-3 game's first level, which ARC Prize designed to be easy \citep{arcprize2026arcagi3}, and nearly every level Kimi solved there it solved after watching the solver play another level of the same game.

\begin{table}[h]
\centering
\small
\caption{First-sight solve rates on the second and third tracks, in percent, with the number of instances, requested at medium reasoning effort through OpenRouter, a level Kimi does not offer. ARC-AGI-2: four fresh instances of each of the track's 25 stream tasks (two of each for Kimi-K3), three demonstrations and one answer, an exact match, a reply capped at 32{,}000 tokens, which no reply of the Kimi models reached, and at 65{,}536 for Qwen3.5-9B. ARC-AGI-3: the track's 40 held-out levels, the bare model wiped before each, a reply capped at 64{,}000 tokens, and the mean charge out of 8. A dash marks a track a model is too slow to play.}
\label{tab:first-sight}
\begin{tabular}{lrrrr}
\toprule
& \multicolumn{2}{c}{ARC-AGI-2} & \multicolumn{2}{c}{ARC-AGI-3} \\
\cmidrule(lr){2-3} \cmidrule(lr){4-5}
model & solved & instances & solved & mean charge \\
\midrule
Kimi K3.1 Pro & 29 & 100 & 25 & 6.8 \\
Kimi-K3 & 56 & 50 & 35 & 6.0 \\
Qwen3.5-9B & 22 & 100 & -- & -- \\
\bottomrule
\end{tabular}
\end{table}

\begin{table}[h]
\centering
\footnotesize\setlength{\tabcolsep}{3pt}
\caption{Initial results on the second and third tracks: Kimi K3.1 Pro, requested at medium reasoning effort, at one user, the two floors through OpenRouter and every other method pinned to Moonshot AI's own endpoint. $P$ is the share of the stream's instances solved, in percent. $L$, $S_{\mathrm{seen}}$ and $S_{\mathrm{held}}$ are read as in Table~\ref{tab:tests}, the demo-cache floor against the stateless floor, KA against K\textsubscript{0} and every other method against itself wiped, from four plays of each test by each floor and two by every other method, a method that keeps nothing reading half its plays against the other half. A dash marks a method that does not play ARC-AGI-3 beyond the pilot of Table~\ref{tab:arc3-pilot}. Small type is the 95\% interval over task resamples. Bold marks each column's best, for $L$ and $S$ among the values whose 95\% interval lies above zero.}
\label{tab:tracks-light}
\begin{adjustbox}{max width=\linewidth}
\begin{tabular}{lrr@{\hspace{0.3em}}lr@{\hspace{0.3em}}lr@{\hspace{0.3em}}l@{\hspace{1em}}rr@{\hspace{0.3em}}lr@{\hspace{0.3em}}lr@{\hspace{0.3em}}l}
\toprule
 & \multicolumn{7}{c}{ARC-AGI-2} & \multicolumn{7}{c}{ARC-AGI-3} \\
\cmidrule(lr){2-8}\cmidrule(lr){9-15}
method & $P$ & \multicolumn{2}{c}{$L$} & \multicolumn{2}{c}{$S_{\mathrm{seen}}$} & \multicolumn{2}{c}{$S_{\mathrm{held}}$} & $P$ & \multicolumn{2}{c}{$L$} & \multicolumn{2}{c}{$S_{\mathrm{seen}}$} & \multicolumn{2}{c}{$S_{\mathrm{held}}$} \\
\midrule
KC\textsubscript{0} & 61.0 & 0 &  & -0.01 & {\tiny [-0.13, 0.10]} & -0.01 & {\tiny [-0.05, 0.03]} & 25.8 & 0 &  & -0.06 & {\tiny [-0.12, 0.00]} & +0.01 & {\tiny [-0.05, 0.06]} \\
KC & 71.0 & +0.35 & {\tiny [0.22, 0.48]} & +0.37 & {\tiny [0.26, 0.48]} & +0.01 & {\tiny [-0.02, 0.04]} & 32.5 & \textbf{+0.11} & {\tiny [0.01, 0.23]} & \textbf{+0.13} & {\tiny [0.04, 0.24]} & -0.02 & {\tiny [-0.10, 0.04]} \\
KP & \textbf{91.0} & \textbf{+0.82} & {\tiny [0.74, 0.90]} & \textbf{+0.92} & {\tiny [0.86, 0.98]} & -0.02 & {\tiny [-0.09, 0.04]} & -- & -- & & -- & & -- & \\
KP\textsubscript{0} & 70.5 & 0 &  & +0.01 & {\tiny [-0.18, 0.18]} & -0.08 & {\tiny [-0.18, 0.01]} & -- & -- & & -- & & -- & \\
K & 74.0 & +0.04 & {\tiny [-0.09, 0.16]} & -0.04 & {\tiny [-0.15, 0.07]} & +0.02 & {\tiny [-0.03, 0.08]} & \textbf{65.8} & -0.02 & {\tiny [-0.13, 0.07]} & +0.03 & {\tiny [-0.06, 0.12]} & \textbf{+0.10} & {\tiny [0.01, 0.18]} \\
K\textsubscript{0} & 71.0 & 0 &  & +0.11 & {\tiny [-0.02, 0.23]} & -0.05 & {\tiny [-0.15, 0.04]} & \textbf{65.8} & 0 &  & +0.01 & {\tiny [-0.21, 0.18]} & +0.02 & {\tiny [-0.21, 0.25]} \\
KA & 72.0 & +0.05 & {\tiny [-0.07, 0.18]} & +0.11 & {\tiny [-0.04, 0.25]} & -0.02 & {\tiny [-0.08, 0.03]} & 63.3 & +0.02 & {\tiny [-0.14, 0.17]} & +0.09 & {\tiny [-0.02, 0.21]} & +0.02 & {\tiny [-0.09, 0.10]} \\
KE & 87.5 & +0.66 & {\tiny [0.47, 0.83]} & +0.66 & {\tiny [0.49, 0.82]} & -0.01 & {\tiny [-0.07, 0.04]} & -- & -- & & -- & & -- & \\
KE\textsubscript{0} & 79.5 & 0 &  & +0.10 & {\tiny [-0.06, 0.23]} & +0.05 & {\tiny [-0.03, 0.13]} & -- & -- & & -- & & -- & \\
\bottomrule
\end{tabular}
\end{adjustbox}
\end{table}

\begin{figure}[htbp]
\centering
\includegraphics[width=\linewidth]{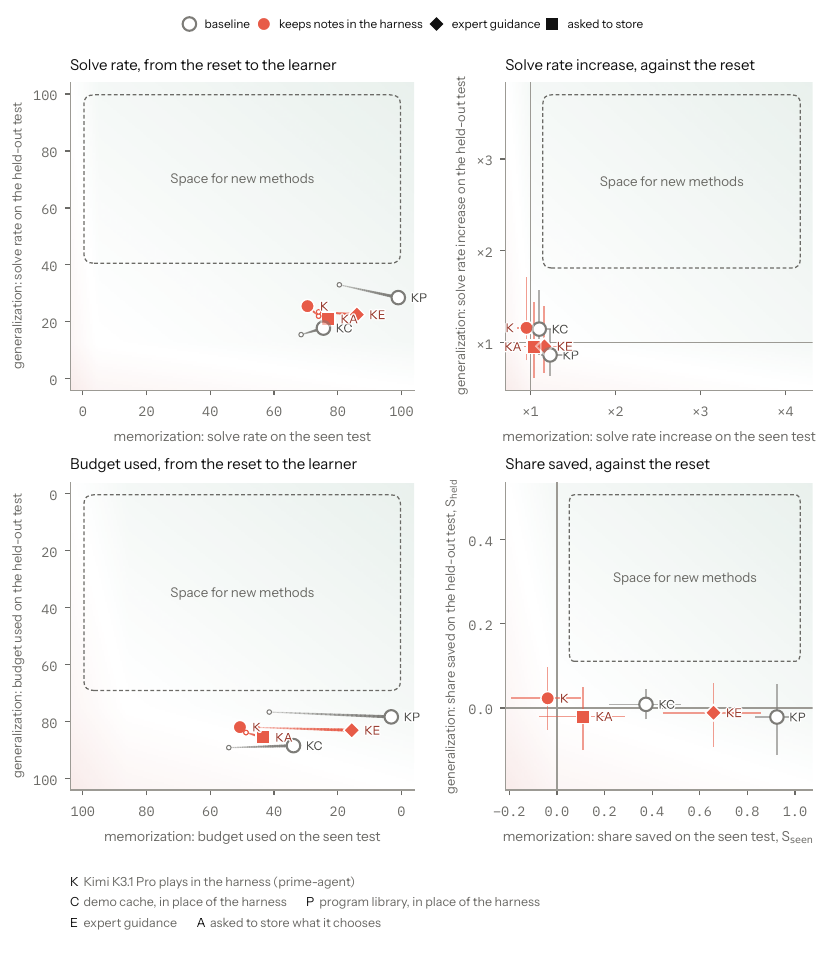}
\caption{Figure~\ref{fig:plane}'s four panels for ARC-AGI-2, with broad axis ranges: every learner Kimi K3.1 Pro has played on this track so far, requested at medium reasoning effort, at one user, each against its reset. The dashed boxes mark the space no learner here reaches yet, which we leave open for new methods and submissions: on the left panels everything above the points drawn, where a learner better on the held-out test than any here would stand, and on the right an open region above and to the right of no gain.}
\label{fig:plane-arc2}
\end{figure}

\begin{figure}[htbp]
\centering
\includegraphics[width=\linewidth]{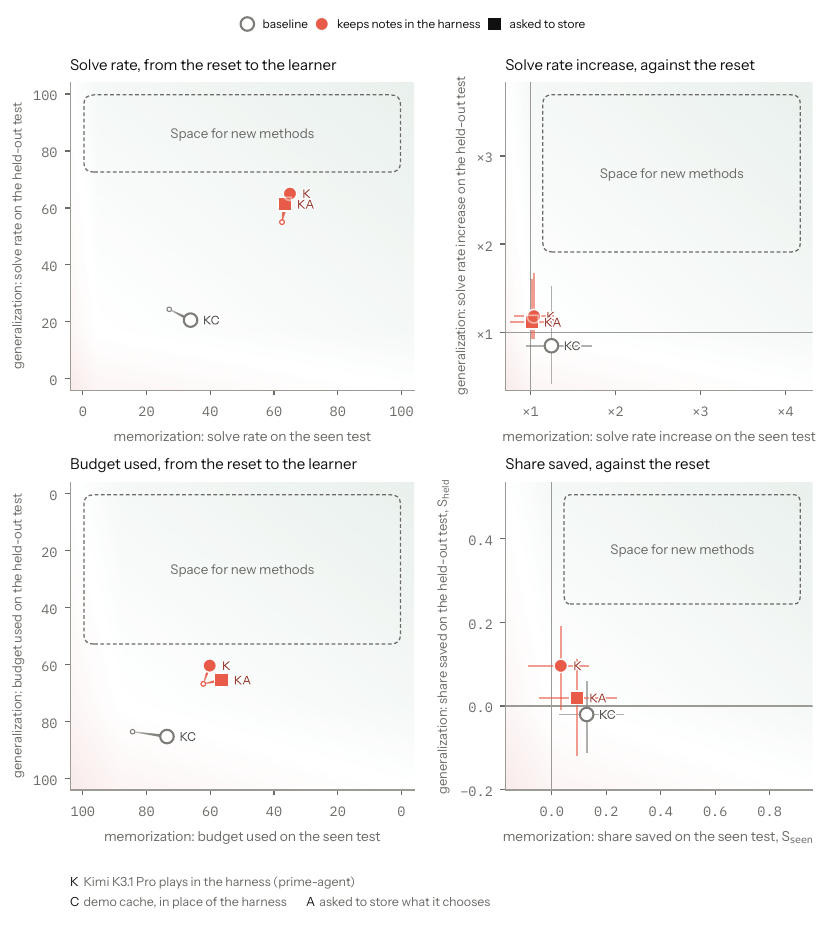}
\caption{Figure~\ref{fig:plane}'s four panels for ARC-AGI-3, as Figure~\ref{fig:plane-arc2} draws them for ARC-AGI-2, with the space left open boxed the same way. The program library and the expert guidance are procedures for grids, so they do not play this track, beyond the pilot of a version of each for games in Appendix~\ref{app:tracks-pilot}, which played no test.}
\label{fig:plane-arc3}
\end{figure}

\paragraph{Initial results.} Kimi K3.1 Pro also plays nine of Section~\ref{sec:baselines}'s methods on ARC-AGI-2 and five on ARC-AGI-3 at one user: the two floors (KC\textsubscript{0} and KC), prime-agent as shipped (K), wiped (K\textsubscript{0}) and asked to store (KA), and on ARC-AGI-2 also the program library (KP) and the expert guidance (KE) with their wiped twins. Each plays its stream once and each test four times for the floors and twice for the rest, which we pinned to Moonshot AI's own endpoint (Table~\ref{tab:tracks-light}). On ARC-AGI-2 the program library and the expert guidance learn their stream, saving 0.82 and 0.66 of the learnable budget, and save most of the seen test, 0.92 and 0.66 of the reset's spend, the program library solving 99\% of it on 3\% of the budget, but nothing on the held-out test. Besides them, on both tracks only the demo-cache floor gains more than chance, and only on returning tasks. It saves 0.35 of the learnable budget on ARC-AGI-2's stream and 0.37 of the reset's spend on the seen test, and 0.11 and 0.13 on ARC-AGI-3, but nothing on either held-out test. What prime-agent kept, as shipped or asked to store, saved nothing measurable against the same harness wiped, on either track's stream or tests: its 0.10 on ARC-AGI-3's held-out test lies above zero pointwise, but not once corrected over the track's three learners (Appendix~\ref{app:bench}). On ARC-AGI-3 the harness solves about two and a half times as many of the stream's levels as the model alone, 65.8\% as shipped and wiped alike against the stateless floor's 25.8, so what it brings there is its way of playing, not what it keeps. Figures~\ref{fig:plane-arc2} and~\ref{fig:plane-arc3} draw Figure~\ref{fig:plane}'s four panels for these methods with broad axis ranges, and they are sparse on purpose. We studied ARC-AGI-1 in depth and tested only lightweight methods on the harder tracks: playing every method of Section~\ref{sec:baselines} on them, with each model and number of users, would cost tens of thousands of dollars, and the student we train cannot yet play ARC-AGI-3, as the first-sight results above show. We expect Figure~\ref{fig:plane}'s trends to recur on both tracks, and leave the rest of each plane open for new methods and submissions, including from labs with the compute to play every method. Both tracks leave models far more to learn than ARC-AGI-1, so they should stay relevant as models improve.

\begin{table}[htbp]
\centering
\small
\caption{The pilot of the program library (KP) and the expert guidance (KE) on ARC-AGI-3, beside Table~\ref{tab:tracks-light}'s methods on the same 120 instances of its stream, a first visit and seven returns of each of the 15 games, in the same order. Kimi K3.1 Pro is requested at medium reasoning effort, at one user, the two floors through OpenRouter and every other method pinned to Moonshot AI's own endpoint. Each method played the stream once. Solved is the share of instances solved, in percent, and the number solved of the 15 first visits; charge is the mean charge out of 8. Bold marks each column's best.}
\label{tab:arc3-pilot}
\begin{tabular}{lrrrrrr}
\toprule
& \multicolumn{2}{c}{stream} & \multicolumn{2}{c}{first visits} & \multicolumn{2}{c}{returns} \\
\cmidrule(lr){2-3}\cmidrule(lr){4-5}\cmidrule(lr){6-7}
method & solved $\uparrow$ & charge $\downarrow$ & solved $\uparrow$ & charge $\downarrow$ & solved $\uparrow$ & charge $\downarrow$ \\
\midrule
KC\textsubscript{0} & 25.8 & 6.69 & 5 & 6.29 & 24.8 & 6.75 \\
KC & 32.5 & 6.04 & 6 & 5.99 & 31.4 & 6.05 \\
KP & 25.8 & 6.32 & 1 & 7.82 & 28.6 & 6.11 \\
K & 65.8 & 4.88 & 12 & 4.13 & 63.8 & 4.98 \\
K\textsubscript{0} & 65.8 & 4.79 & 9 & 5.11 & 66.7 & 4.75 \\
KA & 63.3 & 4.70 & 10 & 4.41 & 62.9 & 4.75 \\
KE & \textbf{77.5} & \textbf{3.25} & \textbf{13} & \textbf{3.45} & \textbf{76.2} & \textbf{3.22} \\
\bottomrule
\end{tabular}
\end{table}

\paragraph{A pilot of KP and KE on ARC-AGI-3.}
\label{app:tracks-pilot}
On ARC-AGI-3 an answer is a sequence of actions rather than a grid, so we built a version of each procedure for games. The program library keeps a world model per game, a program that gives the board after an action and whether a board is won. It is checked at no cost against every transition the game has shown, the demonstrations' and the learner's own, and kept once it gets none wrong. When the game returns, a fixed search plans the level with the kept program, and the plan is played before the model is asked, stopping at the first board the program did not predict. The expert guidance tells prime-agent to request a demonstration of a new game, try each action it did not show and keep notes on the controls and the win condition. It also tells it to write, check and plan with a world model through the check and the search, given as tools. Each played the stream once, with no test and no wiped twin, and Table~\ref{tab:arc3-pilot} reads them beside the methods of Table~\ref{tab:tracks-light} on the same instances. KE solved the most and spent the least, and on returns it solved 12.4 points more than K [7.6, 18.1] at 1.77 less charge [$-2.44$, $-1.09$], with 95\% intervals over the 15 games. Its first visits gained as much, and its returns solved no more than its first visits, so the gain lies in how it plays a game it meets. What it learned across visits is for its wiped twin to show. It kept notes and trajectories for every game, and no program. KP did no better than the floors. It kept programs for three games, and their plans won sp80's last six visits and ft09's last four with no model call, each for 0.09 to 0.10 of the budget of 8. It lost its first visits, solving 1 of 15 against the demo-cache floor's 6, for a reason we have not shown (\texttt{scripts/read\_arc3\_pilot.py}).

\clearpage
\section{Costs}
\label{app:costs}

\begin{figure}[htbp]
\centering
\includegraphics[width=\linewidth]{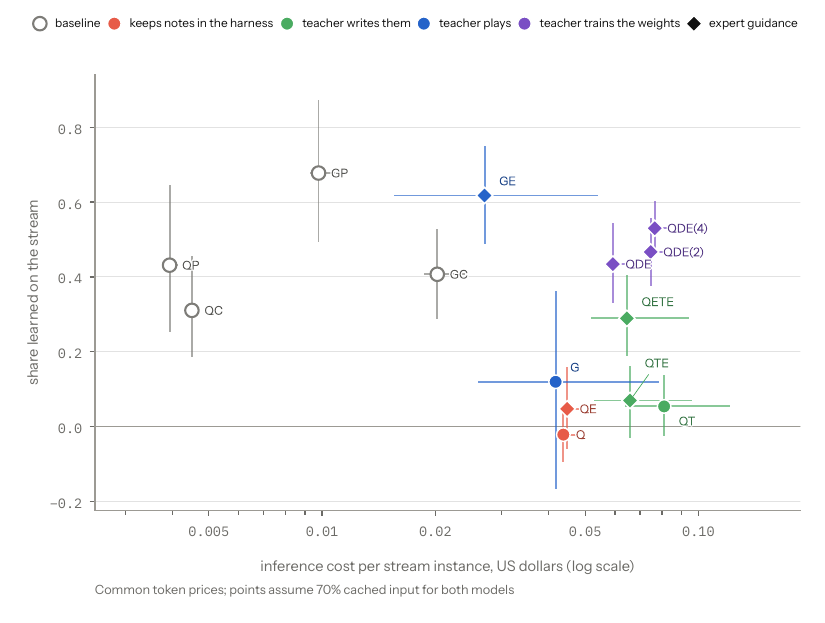}
\caption{Inference cost and share learned on the stream, with labels as in Figure~\ref{fig:plane}. Vertical bars are pointwise 95\% task-bootstrap intervals; horizontal bars span 0 to 100\% cached input, with points at 70\%. QDE(2) and QDE(4) serve two and four users; every other learner serves one.}
\label{fig:inference-cost}
\end{figure}

\paragraph{Inference at common token prices.}
Figure~\ref{fig:inference-cost} reprices the stream tokens of Figure~\ref{fig:plane}'s learners at the same tariff per model (Table~\ref{tab:inference-pricing}). For QDE this is an API-equivalent serving cost, not its GPU bill or a claim that the provider hosts its adapters. QT, QTE and QETE include the teacher's online play; QDE excludes training compute and teacher labels. Costs are divided by all users' valid stream instances and exclude the reset runs and final tests.

Per-run resource records do not retain cache-hit counts. We therefore assume 70\% cached input for both models and show the full sensitivity to that fraction, from 0 to 100\%. These are serving scenarios, not measured bills. Output counts already include reasoning tokens.

\begin{table}[htbp]
\centering
\footnotesize
\begin{tabular}{llrrr}
\toprule
Model & Provider & Input & Cached input & Output \\
\midrule
Qwen3.5-9B & PrimeIntellect & 0.020 & 0.018 & 0.030 \\
GLM-5.3-Flash & Z.ai & 0.150 & 0.030 & 0.500 \\
\bottomrule
\end{tabular}

\caption{US dollars per million tokens, retrieved on 7 November 2026. Qwen uses PrimeIntellect's BF16 endpoint on \href{https://openrouter.ai/api/v1/models/qwen/qwen3.5-9b/endpoints}{OpenRouter}; GLM uses \href{https://docs.z.ai/guides/overview/pricing}{Z.ai}.}
\label{tab:inference-pricing}
\end{table}

\clearpage
\paragraph{Recorded experiment spending.}
Table~\ref{tab:experiment-costs} sums the recorded API charges for the final run families, their controls and test repeats, the Luna order comparisons, and the reported calibrations. Shared controls and teacher streams are counted once. The training estimate uses completed phase durations and GPU allocations, with corrections when a resumed phase changed allocation, at \$4.50 per H100-hour.

The API subtotal is incomplete: some provider charges and failed teacher-label calls were not recorded. GPU serving, idle allocation, earlier configuration searches and failed methods are also excluded. These figures therefore give a documented portion of the experiment costs, not the project's total bill. The released cost manifest lists every included source and the exclusions.

\begin{table}[htbp]
\centering
\footnotesize
\setlength{\tabcolsep}{4pt}
\begin{adjustbox}{max width=\linewidth}
\begin{tabular}{lrrr}
\toprule
Experiment group & Recorded API (\$) & H100 hours & Est. training (\$) \\
\midrule
ARC-AGI-1, Qwen and GLM & 5,735.66 & -- & -- \\
ARC-AGI-1, Luna (three efforts and orders) & 4,406.10 & -- & -- \\
ARC-AGI-2, Kimi methods & 1,549.09 & -- & -- \\
ARC-AGI-3, Kimi methods & 683.04 & -- & -- \\
Private-set and ARC-AGI-2 calibration & 83.18 & -- & -- \\
QDE, 1 user: labels and training & 70.81 & 262.8 & 1,182 \\
QDE, 2 users: labels and training & 170.33 & 293.3 & 1,320 \\
QDE, 4 users: labels and training & 344.00 & 368.4 & 1,658 \\
\midrule
Subtotal & 13,042.22 & 924.5 & 4,160 \\
\bottomrule
\end{tabular}

\end{adjustbox}
\caption{Recorded API charges and estimated H100 training costs, in US dollars. QDE's rows give its teacher-label charges and training over all completed phases of its one-, two- and four-user runs.}
\label{tab:experiment-costs}
\end{table}

\end{document}